\documentclass{article}

\usepackage[main, final]{neurips_2026}

\usepackage[utf8]{inputenc} %
\usepackage[T1]{fontenc}    %
\usepackage{hyperref}       %
\usepackage{url}            %
\usepackage{booktabs}       %
\usepackage{amsfonts}       %
\usepackage{nicefrac}       %
\usepackage{microtype}      %
\usepackage{xcolor}         %
\usepackage{amsmath}
\usepackage{bm}
\usepackage{cleveref}
\usepackage{subcaption}
\usepackage{graphicx}
\usepackage{siunitx}
\usepackage{rotating}
\usepackage[percent]{overpic}
\usepackage{xspace}
\usepackage{makecell}
\usepackage{multirow}
\usepackage{tikz}
\hypersetup{
  pdftitle={RPC-GS: Gaussian Splatting with native RPC Rendering for Satellite Imagery},
  pdfauthor={Valentin Wagner, Sebastian Bullinger, Christoph Bodensteiner, Michael Arens}
}

\renewcommand{\vec}[1]{\ensuremath{\mathbf{#1}}}
\newcommand{\vecs}[1]{\ensuremath{\bm{#1}}}
\newcommand{\mtrx}[1]{\ensuremath{\bm{#1}}}

\crefname{section}{Section}{Sections}
\Crefname{section}{Section}{Sections}

\crefname{figure}{Figure}{Figures}
\Crefname{figure}{Figure}{Figures}

\crefname{table}{Table}{Tables}
\Crefname{table}{Table}{Tables}

\crefname{equation}{Equation}{Equations}
\Crefname{equation}{Equation}{Equations}

\crefname{appendix}{Appendix}{Appendices}
\Crefname{appendix}{Appendix}{Appendices}

\makeatletter
\DeclareRobustCommand\onedot{\futurelet\@let@token\@onedot}
\def\@onedot{\ifx\@let@token.\else.\null\fi\xspace}

\def\eg{\emph{e.g}\onedot} 
\def\ie{\emph{i.e}\onedot} 
 
\def\cf{\emph{c.f}\onedot} 
 
\def\wrt{w.r.t\onedot} 

\makeatother

\title{RPC-GS: Gaussian Splatting with native RPC Rendering for Satellite Imagery}

\author{%
  Valentin Wagner \quad Sebastian Bullinger \quad Christoph Bodensteiner \quad Michael Arens \\
  Fraunhofer IOSB\\
  Ettlingen, Germany \\
  \texttt{\{firstname.lastname\}@iosb.fraunhofer.de} \\
}

\begin{document}

\maketitle

\begin{abstract}

We present RPC-GS, the first Gaussian Splatting framework for satellite imagery that operates natively with Rational Polynomial Camera (RPC) models. The RPC model is the de facto standard for representing the complex imaging geometry of modern pushbroom satellite sensors. To simplify rendering, prior satellite Gaussian Splatting methods replace the RPC model with perspective or affine camera approximations, leading to geometric errors during reconstruction. RPC-GS avoids these approximations by projecting Gaussian means and covariances directly through the RPC model during the splatting process. We embed the RPC model in a chain of carefully selected geo-coordinate transformations representing a mapping from splatting-suitable scene coordinates to image coordinates. To map the Gaussian covariance matrices, we derive a numerically robust Jacobian-based covariance projection for the (partially nonlinear) coordinate transformations. Since RPCs lack an explicit notion of camera depth, we integrate a metric ray-based depth formulation. We benchmark RPC, perspective, and affine camera models in a unified framework, with our native RPC renderer achieving the lowest reconstruction error on leading satellite benchmark datasets, improving mean altitude error over perspective and affine approximations by 34.3\% and 13.5\% on DFC2019, and by 5.0\% and 1.2\% on IARPA2016. Our implementation is publicly available at \url{https://github.com/wagnva/RPC-GS}.

\end{abstract}

\section{Introduction}

Novel methods for view synthesis such as Gaussian Splatting have recently achieved impressive results for 3D reconstruction from multi-view imagery~\citep{eogs,eogs++,shadowgs,skyfallgs}. 
Since Gaussian Splatting is natively designed for use with perspective camera models, its applicability to satellite imagery is limited.
Satellite imaging typically relies on the non-linear Rational Polynomial Camera (RPC) \citep{rpc} to capture complex imaging effects 
caused by push-broom camera systems, while also accounting for effects such as atmospheric distortion and Earth curvature.

Prior Gaussian Splatting methods in the satellite domain address this mismatch by approximating the individual RPC models with either perspective ~\citep{skyfallgs,shadowgs,satgs} or affine~\citep{eogs,eogs++} camera models. 
This approximation is convenient, but the substitution of complex RPC models with simple perspective or affine camera models inherently introduces systematic errors in the projection and covariance splatting process. 
Moreover, perspective approximations often require skew correction by warping the input images, introducing an additional error source~\citep{vissat}. Consequently, replacing complex RPC models with simple approximations potentially introduces errors that degrade the geometry reconstructed by Gaussian Splatting.

\subsection{Contributions}

We eliminate this error source by introducing the first Gaussian Splatting renderer for satellite imagery that operates natively with RPC models.
Specifically, our main contributions are:

(A)~Integrate the RPC model into a carefully chosen geo-coordinate transformation chain, enabling the mapping of splatting-compatible normalized local scene coordinates to global geodetic coordinates and their projection into image coordinates via the RPC model.

(B)~Derive a numerically robust, RPC-native covariance projection using Jacobians. In particular, for the nonlinear transformation from global Cartesian coordinates to ellipsoidal geodetic coordinates, we obtain a solver-independent Jacobian by inverting the local differential of the forward mapping.

(C)~Introduce a metric ray-based depth formulation for RPC cameras, enabling
RPC-consistent front-to-back alpha compositing despite the absence of an inherent
depth coordinate in RPC models.

(D)~Provide the first unified Gaussian Splatting benchmark of RPC, perspective, and affine camera models for satellite imagery, enabling a direct comparison of camera-model choices. Our RPC-native renderer achieves the lowest reconstruction error, reducing mean altitude error over perspective and affine approximations by 34.3\% and 13.5\% on DFC2019~\citep{dfc2019}, and by 5.0\% and 1.2\% on IARPA2016~\citep{IARPA2016}, respectively.

\section{Related Work}

Existing satellite Gaussian Splatting~\citep{gaussian_splatting} methods address the discrepancy between the standard Gaussian Splatting renderer and satellite RPC imagery by replacing RPC projections with simpler surrogate camera models. We categorize these approaches based on the type of camera model used for approximation; the camera models themselves and their introduced approximation errors are analyzed in detail in \cref{sec:camera_models}.

Recent works approximate the RPC model using perspective cameras, enabling the use of the original 3DGS-renderer~\citep{gaussian_splatting}.
Shadow-GS~\citep{shadowgs} models geometric shadows through a ray-casting procedure optimized for Gaussian Splatting.
Skyfall-GS~\citep{skyfallgs} reduces artifacts in off-nadir views using diffusion-guided iterative refinement.
Sat-GS~\citep{satgs} models view-dependent appearance variations and transient objects in satellite imagery using uncertainty masks.

Alternatively, EO-GS~\citep{eogs} adapts Gaussian Splatting to satellite data by approximating the RPC model with an affine camera model, adapting the 3DGS-renderer accordingly. The authors additionally model shadows by detecting geometric occlusions in a secondary sun-view rendering pass.
EO-GS++~\citep{eogs++} extends this approach by using only panchromatic images, evaluating the impact of color information on reconstruction quality.

\section{Camera models for satellite Gaussian Splatting}
\label{sec:camera_models}
In this section, we examine camera models for Gaussian splatting of satellite images. We begin with the Rational Polynomial Camera (RPC) model, the de facto standard for describing the projection geometry of Earth observation satellites~\citep{rpc_mvs}, and then consider two common approximations based on perspective and affine camera models. By quantifying the projection errors induced by these approximations, we motivate a native RPC formulation for Gaussian Splatting.

\subsection{Rational Polynomial Camera}
\label{subsec:rpc}
The RPC (Rational Polynomial Camera) model is the de facto standard representation used by earth observation satellite image vendors~\citep{rpc_mvs}, enabling complex imaging systems to be modeled without satellite-specific camera models.
Notably, the RPC model does not correspond to a physically grounded imaging process but instead serves as a purely mathematical approximation of the
mapping 
$\mathcal{P}: (\lambda, \phi, h) \mapsto (r, c)$
from geodetic coordinates 
(longitude $\lambda$, latitude $\phi$, ellipsoidal altitude $h$)
to image coordinates $(r, c)$.
Specifically, the normalized geodetic coordinates $(\tilde{\lambda}, \tilde{\phi}, \tilde{h})$ are mapped to normalized image coordinates $(\tilde{r}, \tilde{c})$ using four polynomials $p_m$ \citep{rpc}:
\begin{equation}
	\tilde{r}=p_0(\tilde{\lambda},\tilde{\phi},\tilde{h}) / p_1(\tilde{\lambda},\tilde{\phi},\tilde{h}),
	\qquad
	\tilde{c}=p_2(\tilde{\lambda},\tilde{\phi},\tilde{h}) / p_3(\tilde{\lambda},\tilde{\phi},\tilde{h}).
	\label{eq:rpc}
\end{equation}
For numerical stability, each geodetic input and image output coordinate is normalized
componentwise using the corresponding offset and scale parameters specified by
the RPC model. The image coordinates $(r,c)$ are then recovered by applying the
corresponding inverse normalization.
Each polynomial $p_m$ is a cubic polynomial with $20$ coefficients:
\begin{equation}
	p_m(\tilde{\lambda}, \tilde{\phi}, \tilde{h}) = \sum_{\substack{i,j,k \geq 0, \\ i+j+k \leq 3}} c_{ijk}^{(m)} \, \tilde{\lambda}^i \tilde{\phi}^j \tilde{h}^k.
\end{equation}
Here, $i,j,k$ index the powers of the normalized coordinates, and
$c_{ijk}^{(m)}$ is the corresponding RPC coefficient for polynomial $p_m$.
This rational form provides a flexible sensor model for satellite imagery without requiring an explicit physical camera model,  while capturing complex imaging effects such as optical distortion, Earth curvature, atmospheric refraction, and camera motion. \citep{rpc}

\subsection{Approximation as pinhole cameras}

\label{subsec:perspective_cams}
Previous satellite-adapted Gaussian Splatting pipelines \citep{skyfallgs,shadowgs,satgs} locally approximate the RPC camera model using a perspective pinhole camera, \ie a central projection model in which
all rays pass through a single camera center. The main motivation is of practical nature, since standard Gaussian Splatting pipelines are designed for conventional imagery captured by perspective cameras. In practice, this approximation is commonly obtained via SatelliteSfM \citep{vissat}.

For each view, a set of 3D–2D correspondences $\{(\mathbf{x}_i^{\mathrm{enu}},\mathbf{x}_i^{\mathrm{img}})\}_{i=1}^M$ is constructed by sampling a finite grid of $M$ points in a local East–North–Up (ENU) coordinate system and projecting the points to image coordinates using the RPC model. Using a standard direct linear transform allows to estimate a projection matrix $\mtrx{P} \in \mathbb{R}^{3\times4}$ satisfying $\vecs{x}^{\mathrm{img}} \sim \mtrx{P}\vecs{x}^\mathrm{enu}$ for the sampled correspondences.
This is followed by a decomposition into intrinsics $\mtrx{K}$, rotation $\mtrx{R}$ and translation $\vecs{t}$ using a QR decomposition with sign and scale normalization.

Due to the large distances involved, the camera centers are positioned far from the scene, leading to large absolute depth values with limited relative variation.
Additionally, the resulting solution yields a full intrinsic matrix $\mtrx{K}$ that includes a non-zero skew parameter. Since Gaussian Splatting does not natively support pinhole cameras with skew, SatelliteSfM \citep{vissat} applies a skew correction by warping the input images, effectively converting the camera to one with zero skew. This step is not lossless, as it shifts visible scene content outside the image bounds and introduces interpolation artifacts.

\subsection{Approximation as affine cameras}
\label{subsec:affine}
EOGS \cite{eogs} instead approximate the RPC camera using an affine transformation $\mathcal{A}: \mathbf{x} \in \mathbb{R}^3 \mapsto A\mathbf{x} + \mathbf{a} \in \mathbb{R}^2$, where $\vecs{x} \in \mathbb{R}^3$ is a 3D point, $\vecs{A} \in \mathbb{R}^{2\times3}$ is the affine projection, and $\vecs{a} \in \mathbb{R}^2$ is a translation offset.
The authors argue that this parallel projection is well suited to satellite imagery because of the large camera-to-scene distances, while also offering improved computational efficiency. 
Since the affine camera model does not encode depth, EOGS instead uses inverse altitude. While this is reasonable for nadir views, it can lead to incorrect depth ordering in off-nadir scenes, especially around tall buildings.

Building on the affine formulation, the authors additionally propose an altitude-dependent shadow computation implemented via an additional sun-view rendering pass.
Although the affine formulation is simpler than the perspective approximation in \cref{subsec:perspective_cams}, practical use requires modifying the Gaussian mean and covariance projection inside of the Gaussian Splatting renderer.

\subsection{Limitations of RPC approximation}

Perspective and affine cameras provide two principled approximations to the RPC model, corresponding respectively to central projection and parallel projection. 
Although both models approximate the RPC mapping locally, this approximation introduces a fundamental model mismatch, since neither a pinhole nor an affine camera can exactly represent the nonlinear image formation encoded by the RPC model. We evaluate the resulting projection error explicitly in \cref{tab:rpc_approx_error}, reporting the mean projection error for varying scene extents over all 50 images of the IARPA2016 dataset \citep{IARPA2016}.
The error is small for limited image regions but grows rapidly with scene size. At larger scales, it becomes substantial: $3.5$ pixels at a ground sampling distance of $0.31\,\mathrm{m}$ corresponds to about $1\,\mathrm{m}$ horizontal reconstruction error from camera model approximation alone.

Motivated by these approximation errors, we propose a native RPC rendering formulation for Gaussian Splatting that uses the sensor model directly. This removes the need for camera conversion and image warping, while avoiding a source of systematic geometric error that existing approaches inherit from surrogate camera models.

\begin{table*}
	\caption{Mean approximation error in pixels when replacing the RPC camera model with either a perspective or an affine camera model. The error increases rapidly with scene size. Results are averaged over all 50 images of the IARPA2016 dataset~\citep{IARPA2016}.}
	\label{tab:rpc_approx_error}
	\begin{center}
		\begin{tabular}{c|ccccccc}
			\toprule
			Image size (pixels) & $256^2$ & $512^2$ & $1024^2$ & $2048^2$ & $4096^2$ & $8192^2$ & $16384^2$ \\
			\midrule
			Perspective & 0.002 & 0.005 & 0.015 & 0.056 & 0.218 & 0.865 & 3.456 \\
			Affine & 0.003 & 0.007 & 0.022 & 0.085 & 0.337 & 1.346 & 5.378 \\
			\bottomrule
		\end{tabular}
	\end{center}
\end{table*}

\section{RPC-GS: Native RPC Gaussian Splatting}
Gaussian Splatting~\citep{gaussian_splatting} represents a 3D scene with a
set of Gaussian primitives, each parameterized by a mean
$\vecs{\mu}^{\mathrm{3D}}$, covariance $\mtrx{\Sigma}^{\mathrm{3D}}$,
opacity $\alpha$, and color $\vec{c}$. The primitives are optimized from
multi-view imagery via differentiable rendering.

The Gaussian Splatting rendering process is divided into two separate stages.
The first stage projects the 3D Gaussians (with mean $\vecs{\mu}^{\mathrm{3D}}$ and covariance  $\vecs{\Sigma}^{\mathrm{3D}}$) into image coordinates as $\vecs{\mu}^{\mathrm{2D}}$ and $\mtrx{\Sigma}^{\mathrm{2D}}$. 
The depth $d$ is defined by the position of the Gaussian mean
$\vecs{\mu}^{\mathrm{3D}}$ along the camera viewing direction.
In the second stage (rasterization), the Gaussians are sorted front-to-back by the depth $d$ and aggregated using traditional alpha compositing to render the image. 

\begin{figure}[t]
	\includegraphics[width=\textwidth]{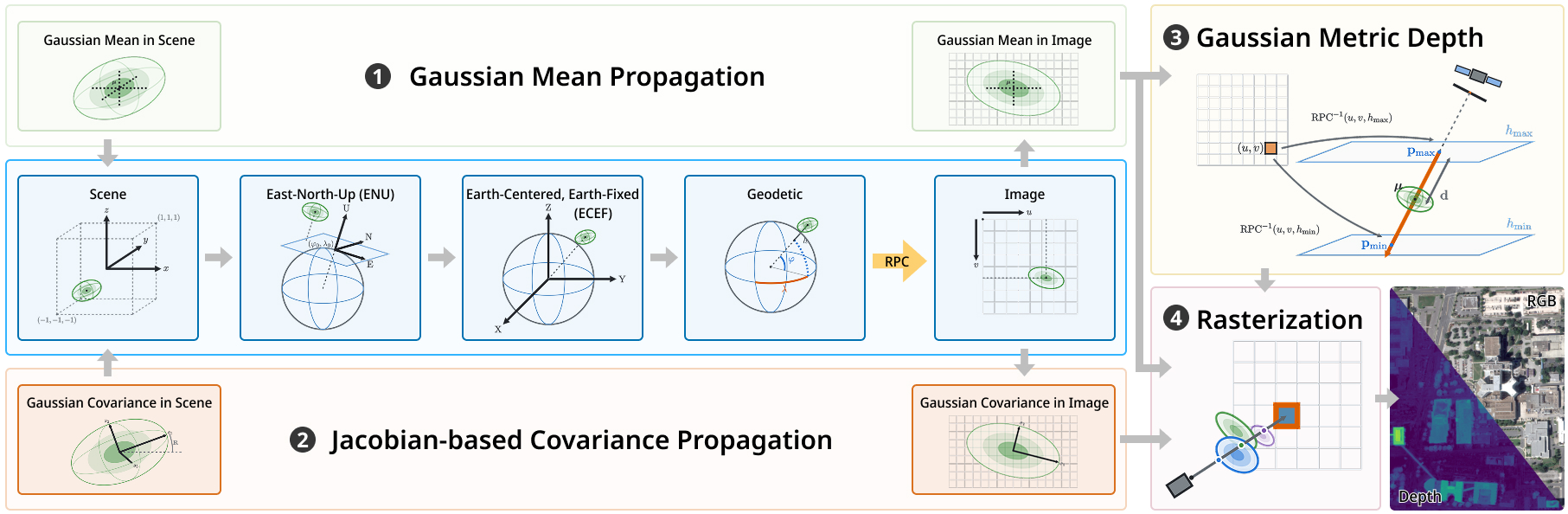}
	
	\caption{
	\textbf{Overview of RPC-GS.}
	We integrate satellite-native RPC camera models into Gaussian Splatting.
	(1) Gaussian means are mapped from the normalized local scene frame through
	metric ENU, ECEF, and geodetic coordinates before RPC projection into image space.
	(2) Covariances are propagated through the same transformation chain using
	Jacobians, with a numerically robust treatment of the ECEF $\rightarrow$ geodetic transformation.
	(3) A height-bounded inverse-RPC ray defines metric depth as the scalar position
	of each Gaussian mean along the ray.
	(4) Gaussians are depth-sorted and alpha-composited to render RGB and depth.
	}
	\label{fig:pipeline}
\end{figure}

RPC-GS retains this two-stage rendering structure, but replaces the standard perspective projection of Gaussian means and covariances with an RPC-native projection embedded in a geo-coordinate transformation chain. We further introduce a metric, ray-based depth for RPC-consistent front-to-back ordering during rasterization. An overview of the proposed RPC-GS pipeline is shown in \cref{fig:pipeline}.

\subsection{Coordinate System Transformations for Gaussian Splatting of Satellite Data}

RPC models are typically defined in geodetic coordinates (\ie longitude $\lambda$, latitude $\phi$, and ellipsoidal altitude $h$).
Due to the approximately ellipsoidal shape of the Earth, geodetic coordinates
are defined with respect to a reference ellipsoid such as WGS84~\citep{wgs84}. This results in a nonlinear representation with non-uniform scaling. Equal changes in longitude, latitude, and altitude therefore do not correspond to equal physical distances, making Gaussian means and covariances poorly conditioned and less geometrically interpretable.

A metric alternative to geodetic coordinates is the Earth-centered, Earth-Fixed (ECEF) coordinate system. It is defined as a global Cartesian frame, with its origin at the center of mass, rotating with the earth. 
ECEF coordinates are expressed at the scale of Earths radius (around $6.4 \cdot 10^6$ meters for WGS84), resulting in poor numerical conditions \citep{GPSS}.
Although this may be mitigated by subtracting a local
origin and applying an affine normalization, the resulting frame still inherits the globally fixed ECEF axis alignment.

Instead, we parameterize the Gaussians in a local East-North-Up (ENU) frame
anchored at an observer point $(\lambda_o, \phi_o, h_o)$. The East-North plane
is set parallel to the tangent plane of the reference ellipsoid at
$(\lambda_o, \phi_o, 0)$, yielding a local Cartesian frame adapted to the scene
geometry. Finally, we normalize the ENU coordinates into a normalized scene frame to improve
numerical conditioning and ensure stable, well-scaled gradients during
reconstruction.	
The chain of coordinate systems is therefore defined as:
\begin{equation}
	\text{Scene} \rightarrow \text{ENU} \rightarrow \text{ECEF} \rightarrow \text{Geodetic} \rightarrow \text{Image}.
\end{equation}

\subsection{Gaussian Mean Projection}
To project the Gaussian mean $\vecs{\mu}$ into image coordinates, we need to convert its normalized coordinates $\vecs{\mu}^{\mathrm{scene}}$ into the global geodetic frame $\vecs{\mu}^{\mathrm{geodetic}}$ and project the resulting values into image coordinates using the RPC projection function $\mathcal{P}$ from \cref{subsec:rpc}. The individual steps are:

\paragraph{Scene $\rightarrow$ ENU}
Reverse center shift $\vecs{c}$ and scale $s$ via $\vecs{\mu}^{\mathrm{enu}} = \frac{1}{s}\vecs{\mu}^\mathrm{scene} + \vecs{c}$.

\paragraph{ENU $\rightarrow$ ECEF}
The ENU observer point $(\lambda_o, \phi_o, h_o)$ is converted to the corresponding ECEF position $\vecs{x}_o^{\mathrm{ecef}}$ using the WGS84 ellipsoid.
Given a Gaussian center $\vecs{\mu}^{\mathrm{enu}}$ in the metric ENU frame, the corresponding ECEF coordinate is obtained via an affine transformation consisting of a rotation $R^{\mathrm{ecef}\leftarrow\mathrm{enu}}(\phi_o, \lambda_o)$ and a translation by $\vecs{x}_o^{\mathrm{ecef}}$ \citep{Geodesy}.
\begin{equation}
	\vecs{\mu}^{\mathrm{ecef}} =
	\vecs{x}^{\mathrm{ecef}}_o +
	R^{\mathrm{ecef}\leftarrow\mathrm{enu}}(\phi_o, \lambda_o)\,\vecs{\mu}^{\mathrm{enu}}
\end{equation}

\paragraph{ECEF $\rightarrow$ Geodetic}
\label{par:ecef_to_geodetic}
We first define the forward Geodetic $\rightarrow$ ECEF mapping before
discussing the inverse ECEF $\rightarrow$ Geodetic transformation.
The forward mapping between geodetic coordinates $(\lambda_\mu, \phi_\mu, h_\mu)$ and ECEF coordinates $(x_\mu, y_\mu, z_\mu)$ is given by
\begin{equation}
	\begin{bmatrix}
		x_\mu\vphantom{(N(\phi)+h)\cos\phi\cos\lambda}\\
		y_\mu\vphantom{(N(\phi)+h)\cos\phi\sin\lambda}\\
		z_\mu\vphantom{\left(N(\phi)(1-e^2)+h\right)\sin\phi}
	\end{bmatrix}
	=
	\begin{bmatrix}
		(N_{\phi_\mu}+h_\mu)\cos\phi_\mu\cos\lambda_\mu \\
		(N_{\phi_\mu}+h_\mu)\cos\phi_\mu\sin\lambda_\mu \\
		\left(N_{\phi_\mu}(1-e^2)+h_\mu\right)\sin\phi_\mu
	\end{bmatrix},
	\label{eq:geodetic_to_ecef}
\end{equation}
where 
$N_{\phi_\mu}=a/(1-e^2\sin^2\phi_\mu)^{1/2}$ 
denotes the radius of curvature in the prime vertical,
$a$ is the semi-major axis radius, and 
$e$
is the first eccentricity of the WGS84 ellipsoid, quantifying its deviation from a perfect sphere~\citep{navipedia_geodetic}. 
The inverse ECEF $\rightarrow$ Geodetic mapping is nonlinear because both the trigonometric terms and curvature radius $N_{\phi_\mu}$ depend on latitude $\phi_\mu$ in \cref{eq:geodetic_to_ecef}.
We therefore compute the geodetic Gaussian center $\vecs{\mu}_\mathrm{geodetic}$ using the non-iterative, closed-form method of \citet{You2000}.

\paragraph{Geodetic $\rightarrow$ Image}
Project $\vecs{\mu}_\mathrm{geodetic}$ onto the image plane using the RPC model $\mathcal{P}$ with $\mathcal{P}(\vecs{\mu}^\mathrm{geodetic}) = \vecs{\mu}^{\mathrm{image}} = (r, c)$.

\subsection{Gaussian Covariance Projection}
To splat the covariance $\mtrx{\Sigma}^{\mathrm{3D}}$ from scene coordinates into the 2D image, standard Gaussian Splatting \citep{gaussian_splatting} applies a linear world-to-camera view transform $\mtrx{W}$ and a local linearization of the projection using its Jacobian $\mtrx{J}_\mu$ at the Gaussian mean $\vecs{\mu}^{\mathrm{3D}}$~\citep{zwicker_ewa_splatting}.
\begin{equation}
	\mtrx{\Sigma}^{\mathrm{image}} = \mtrx{J}_\mu\mtrx{W}\mtrx{\Sigma}^{\mathrm{3D}}\mtrx{W}^T\mtrx{J}_\mu^T
\end{equation}
In contrast, in an RPC-based formulation, there is no explicit camera-coordinate frame.
We instead view the Scene $\rightarrow$ Image mapping as a composition of coordinate
transformations and propagate the covariance using the corresponding Jacobian $\mtrx{J}^{\mathrm{image} \leftarrow \mathrm{scene}}$.
\begin{equation}
	\mtrx{\Sigma}^{\mathrm{image}} =
	\mtrx{J}^{\mathrm{image} \leftarrow \mathrm{scene}}
	\mtrx{\Sigma}^{\mathrm{scene}}
	(\mtrx{J}^{\mathrm{image} \leftarrow \mathrm{scene}})^{T}.
\end{equation}
Here, $\mtrx{J}^{\mathrm{image} \leftarrow \mathrm{scene}}$ is the product of the Jacobians of the
Scene $\rightarrow$ ENU $\rightarrow$ ECEF $\rightarrow$ Geodetic $\rightarrow$ Image
transformation chain.
\begin{equation}
	\mtrx{J}^{\mathrm{image} \leftarrow \mathrm{scene}} =
	\mtrx{J}^{\mathrm{image}\leftarrow\mathrm{geodetic}}_{\mu}
	\mtrx{J}^{\mathrm{geodetic}\leftarrow\mathrm{ecef}}_{\mu}
	\mtrx{J}^{\mathrm{ecef}\leftarrow\mathrm{enu}}
	\mtrx{J}^{\mathrm{enu}\leftarrow\mathrm{scene}}
\end{equation}
To compute the Jacobians, we solve the affine transformations (\ie Scene $\rightarrow$ ENU and ENU $\rightarrow$ ECEF) exactly, and locally linearize the nonlinear transformations (\ie ECEF $\rightarrow$ Geodetic and Geodetic $\rightarrow$ Image) at the Gaussian mean $\vecs{\mu}$, denoted by the subscript $_{\mu}$.
Analog to standard Gaussian Splatting \cite{gaussian_splatting}, we drop the third row and column of $\mtrx{\Sigma}^{\mathrm{image}}$ to obtain the final $2\times2$ covariance $\mtrx{\Sigma}^{2D}$.

\paragraph{Scene $\rightarrow$ ENU}
The Jacobian $\mtrx{J}^{\mathrm{enu} \leftarrow \mathrm{scene}}$ is the inverse scaling given by
$\frac{1}{s} I_3$.

\paragraph{ENU $\rightarrow$ ECEF}
As the metric ENU$\rightarrow$ ECEF transformation is a rigid-body-transformation, the Jacobian $\mtrx{J}^{\mathrm{ecef} \leftarrow \mathrm{enu}}$ is given by the constant rotation matrix $\mtrx{R}^{\mathrm{ecef}\leftarrow\mathrm{enu}}(\lambda_o, \phi_o)$ based on the ENU observer location $(\lambda_o, \phi_o)$.

\paragraph{ECEF $\rightarrow$ Geodetic}
To avoid the differentiation of the nonlinear ECEF $\rightarrow$ Geodetic inverse
solver in \cref{par:ecef_to_geodetic}, we obtain the inverse Jacobian from the
local differential of the forward Geodetic $\rightarrow$ ECEF map
\citep{Geodesy}. This gives a solver-independent and numerically stable first-order derivative
at the Gaussian mean
$\boldsymbol{\mu}^{\mathrm{geodetic}}=(\lambda_\mu,\phi_\mu,h_\mu)$.

To allow for easier differentiation, we define a local orthonormal basis
$(\mathbf{e}_{\lambda_\mu},\mathbf{e}_{\phi_\mu},\mathbf{e}_{h_\mu})$,
corresponding to the east, north, and ellipsoidal-up axes expressed in ECEF coordinates
at the Gaussian mean $\boldsymbol{\mu}^{\mathrm{geodetic}}$~\citep{navipedia_ecef_enu}. 
\begin{equation}
	\vecs{e}_{\lambda_\mu} =
	\begin{bmatrix}
		-\sin\lambda_\mu \\
		\cos\lambda_\mu \\
		0
	\end{bmatrix},
	\qquad
	\vecs{e}_{\phi_\mu} =
	\begin{bmatrix}
		-\sin\phi_\mu\cos\lambda_\mu \\
		-\sin\phi_\mu\sin\lambda_\mu \\
		\cos\phi_\mu
	\end{bmatrix},
	\qquad
	\vecs{e}_{h_\mu} =
	\begin{bmatrix}
		\cos\phi_\mu\cos\lambda_\mu \\
		\cos\phi_\mu\sin\lambda_\mu \\
		\sin\phi_\mu
	\end{bmatrix}
	\label{eq:enu_base}
\end{equation}
Differentiating the forward Geodetic $\rightarrow$ ECEF map in
\cref{eq:geodetic_to_ecef} yields the local linear map $\mathbf{J}^{\mathrm{ecef} \leftarrow \mathrm{geodetic}}_{\mu}$. A
step-by-step derivation is provided in \cref{app:diff_geo_ecef}.
\begin{equation}
	\mathbf{J}^{\mathrm{ecef} \leftarrow \mathrm{geodetic}}_{\mu}
	=
	\underbrace{
		\begin{bmatrix}
			\mathbf{e}_{\lambda_\mu} &
			\mathbf{e}_{\phi_\mu} &
			\mathbf{e}_{h_\mu}
		\end{bmatrix}
	}_{\mathbf{R}_\mu}
	\underbrace{
		\operatorname{diag}\!\left(
		(N_{\phi_\mu}+h_\mu)\cos\phi_\mu,
		M_{\phi_\mu}+h_\mu,
		1
		\right)
	}_{\mathbf{H}_\mu}
	\label{eq:jac_geo_to_ecef}
\end{equation}
Here \(N_{\phi_\mu}\) and \(M_{\phi_\mu}\) denote the prime-vertical and
meridional radii of curvature of the WGS84 reference ellipsoid evaluated at
\(\phi_\mu\)~\citep{wgs84}. The matrix \(\mathbf{H}_\mu\) converts geodetic
coordinate perturbations, into local East-North-Up metric displacements, and
\(\mathbf{R}_\mu\) rotates these local components into ECEF coordinates.
Since $\mtrx{R}_\mu$ is orthonormal (\ie $\mtrx{R}_\mu^{-1}=\mtrx{R}_\mu^{T}$), the inverse Jacobian $\mtrx{J}_{\mu}^{\mathrm{geodetic} \leftarrow \mathrm{ecef}}$ is easily obtained through inversion:
\begin{equation}
	\mathbf{J}_{\mu}^{\mathrm{geodetic} \leftarrow \mathrm{ecef}}
	=
	\mathbf{H}_\mu^{-1}\mathbf{R}_\mu^\top
	=
	\operatorname{diag}\!\left(
	\frac{1}{(N_{\phi_\mu}+h_\mu)\cos\phi_\mu},
	\frac{1}{M_{\phi_\mu}+h_\mu},
	1
	\right)
	\begin{bmatrix}
		\mathbf{e}_{\lambda_\mu}^{\top}\\
		\mathbf{e}_{\phi_\mu}^{\top}\\
		\mathbf{e}_{h_\mu}^{\top}
	\end{bmatrix}
\end{equation}

\paragraph{Geodetic $\rightarrow$ Image}
Given the RPC definition in~\cref{eq:rpc}, the derivatives of the
normalized row and column coordinates are obtained by differentiating the
corresponding polynomials. For normalized row $\tilde{r}=p_0/p_1$ the quotient rule gives following derivation:
\begin{equation}
	\frac{\partial \tilde{r}}{\partial \tilde{\vecs{\mu}}^\mathrm{geodetic}}
	=
	\frac{\partial (p_0/p_1)}{\partial \tilde{\vecs{\mu}}^\mathrm{geodetic}}
	= \frac{p_{1}\,\partial p_{0} / \partial \tilde{\vecs{\mu}}^\mathrm{geodetic} - p_0\,\partial p_1 / \partial \tilde{\vecs{\mu}}^\mathrm{geodetic}}{p_1^2}
\end{equation}
The derivative of the normalized column coordinate $\tilde{c}=p_2/p_3$ is obtained analogously.
Stacking the derivatives of $\tilde{r}$ and $\tilde{c}$ yields the Jacobian of the RPC projection \wrt normalized geodetic and image coordinates
The Jacobian with respect to unnormalized geodetic and image coordinates follows
from the RPC normalization parameters (\ie  $r_{\mathrm{scale}},c_{\mathrm{scale}},\lambda_{\mathrm{scale}},\phi_{\mathrm{scale}},h_{\mathrm{scale}}$)
\begin{equation}
	\mtrx{J}^{\mathrm{image}\leftarrow \mathrm{geodetic}}_\mu
	=
	\operatorname{diag}\!\left(r_{\mathrm{scale}},c_{\mathrm{scale}}\right)
	\,
	\begin{bmatrix}
		\partial \tilde{r}/\partial \tilde{\lambda} & \partial \tilde{r}/\partial \tilde{\phi} & \partial \tilde{r}/\partial \tilde{h} \\
		\partial \tilde{c}/\partial \tilde{\lambda} & \partial \tilde{c}/\partial \tilde{\phi} & \partial \tilde{c}/\partial \tilde{h}
	\end{bmatrix}
	\,
	\operatorname{diag}\!\left(
	\frac{1}{\lambda_{\mathrm{scale}}},
	\frac{1}{\phi_{\mathrm{scale}}},
	\frac{1}{h_{\mathrm{scale}}}
	\right)
\end{equation}

\subsection{RPC Ray-based Depth}
Standard Gaussian Splatting \citep{gaussian_splatting} orders projected Gaussian means $\vecs{\mu}$ by their $z$-axis depth in camera coordinates for front-to-back alpha compositing. However, the RPC projection $\mathcal{P}$ (\cf \cref{subsec:rpc}) maps global geodetic coordinates directly to image coordinates and does not provide an explicit notion of depth.
We therefore introduce a ray-based depth parameterization inspired by SatNeRF \citep{satnerf}. For each pixel $(r, c)$, we determine two geodetic points $\mathbf{x}_{\min}^\mathrm{geodetic} = (\lambda_{\min}, \phi_{\min}, h_{\min})$ and $\mathbf{x}_{\max}^\mathrm{geodetic} = (\lambda_{\max}, \phi_{\max}, h_{\max})$ using the inverse RPC projection function such that 
\begin{equation}
	\mathcal{P}(\mathbf{x}_{\min}^\mathrm{geodetic}) = \mathcal{P}(\mathbf{x}_{\max}^\mathrm{geodetic}) = (r, c).
\end{equation}
These points define a viewing ray in metric ENU coordinates:
\begin{equation}
	\mathbf{r}(t) = \mathbf{x}_{\max}^{\mathrm{enu}} + t \mathbf{d},
	\qquad
	\mathbf{d} = 
	\left(\mathbf{x}_{\min}^{\mathrm{enu}} - \mathbf{x}_{\max}^{\mathrm{enu}}\right)
	\,/\,\,
	||\mathbf{x}_{\min}^{\mathrm{enu}} - \mathbf{x}_{\max}^{\mathrm{enu}}||_2
\end{equation}
Given a Gaussian mean $\boldsymbol{\mu}^{\mathrm{enu}}$, we define its depth $d$ as the scalar for which $\mathbf{r}(d) = \boldsymbol{\mu}^{\mathrm{enu}}$. This yields a metric depth along the ray.
The height bounds $[h_{\min}, h_{\max}]$ are fixed per scene and is obtained from external sources such as low-resolution elevation data (\eg SRTM \citep{srtm4}).
We precompute depth rays for each pixel and bilinearly interpolate the corresponding ray parameters based on the projected Gaussian center $\vecs{\mu}^{image}$.

\section{Implementation}
\label{sec:implementation}
We implement the native RPC mean projection, covariance splatting, and ray-based depth computation using custom CUDA kernels. 
The corresponding backward passes are implemented explicitly in CUDA, enabling efficient end-to-end optimization of the rendering pipeline. 
Alpha compositing is performed using the original 3DGS CUDA rasterizer~\citep{gaussian_splatting}.
To facilitate development and support extensibility, we additionally provide a PyTorch implementation of the mean projection, covariance splatting, and ray-based depth computation. Automatic differentiation is used to obtain the corresponding gradients, with alpha compositing remaining in the original 3DGS CUDA rasterizer.
Runtime and memory costs for both implementations are reported in \cref{tab:computational_efficiency}.

Following WildGaussians~\citep{wild_gaussians}, we employ trainable embeddings $\{\vecs{g}_i\}^M_{i=1}$ and $\{\vecs{e}_j\}^N_{j=1}$ for all Gaussians $M$ and images $N$. A lightweight MLP $f$ predicts an affine color transformation $(\beta_i, \gamma_i) = f(\vecs{e}_j, \vecs{g}_i, \vecs{a}_i)$ for each Gaussian $i$ based on its base color $\vecs{a}_i$, embedding $\vecs{g}_i$ and image-embedding $\vecs{e}_j$.
We assume a Lambertian surface analog to prior work~\citep{satnerf,eonerf,eogs,skyfallgs} and omit the use of spherical harmonics, predicting the view-independent color $\vecs{c}_i=\gamma_i \hat{\vecs{c}}_i + \beta_i$ for each Gaussian $i$. 

For all camera models, we optimize using the standard photometric objective $\mathcal{L} = (1-\lambda)\mathcal{L}_\mathrm{1}+\lambda\mathcal{L}_{\mathrm{SSIM}}$ with $\lambda=0.2$~\citep{gaussian_splatting}. Apart from the appearance embeddings described above, we introduce no additional learnable parameters, loss terms, or regularizers, and retain the original 3DGS densification and pruning strategy.

\section{Experiments and Evaluation}
\label{sec:experiments}
We evaluate our proposed RPC-renderer as well as the perspective and affine camera approximations in a shared framework with identical
training and evaluation settings; only the camera model and corresponding
renderer are substituted. 
We integrate the standard 3DGS~\citep{gaussian_splatting} renderer for perspective cameras and the EOGS~\citep{eogs} renderer for affine cameras. 
For our RPC renderer, we use the custom CUDA-kernel implementation throughout.
To ensure a camera-focused evaluation, we omit non-camera satellite-domain adaptations such as explicit shadow modeling or regularization, trading absolute performance for a focused camera model evaluation.
All experiments are run for 30,000 iterations on an NVIDIA RTX 4090.

We conduct experiments on four standard benchmark scenes from the 2019 IEEE GRSS Data Fusion Contest
(DFC2019) dataset~\citep{dfc2019} and three standard benchmark scenes from the 2016 IARPA
Multi-View Stereo 3D Mapping Challenge (IARPA2016) dataset~\citep{IARPA2016}, which have been widely adopted in prior work.
The DFC2019 dataset contains multi-date WorldView-3 true-color RGB images
stored as 8-bit unsigned integers. The selected scenes cover suburban and
urban areas of Jacksonville, USA, with 10--20 images per scene.
The IARPA2016 dataset contains 40--50 images of full-precision panchromatic and multispectral
imagery of Buenos Aires, Argentina. We pansharpen and tone-map to
8-bit RGB~\citep{vissat,ssr}. We bundle-adjust the RPC models~\citep{rpc_bundle_adjustment} and the derived perspective and affine camera models, and initialize Gaussians with Structure-from-Motion~\citep{vissat}.

\begin{table*}[t]
	
	\renewcommand{\r}[2]{%
		\makecell{#1\\[-1pt]{\scriptsize $\pm$#2}}%
	}
	\newcommand{\header}[2]{%
		\makecell[l]{#1\\[-1pt]\tiny{#2}}%
	}
	
	\centering
	\caption{
Quantitative comparison of native RPC Gaussian Splatting against perspective and affine approximations on DFC2019~\cite{dfc2019} and IARPA2016~\cite{IARPA2016} scenes. All methods use identical settings, substituting only the camera model and renderer. 
Native RPC
achieves the lowest mean altitude MAE on both datasets,
while maintaining competitive PSNR. We report the mean and standard deviation over n = 3 runs for
each model and scene, evaluated on the training views.
	}
	\label{tab:results_both}
	\setlength{\tabcolsep}{3pt}
	
	\textbf{DFC2019~\citep{dfc2019}}
	\vspace{0.25em}

	\begin{tabular}{@{}lccccc ccccc@{}}
		\toprule
		& \multicolumn{5}{c}{\textbf{MAE [m] $\downarrow$}}
		& \multicolumn{5}{c}{\textbf{PSNR [dB] $\uparrow$}} \\
		\cmidrule(lr){2-6}\cmidrule(lr){7-11}
		Camera Model
		& 004 & 068 & 214 & 260 & Mean
		& 004 & 068 & 214 & 260 & Mean \\
		\midrule

		\header{Perspective}{\citep{gaussian_splatting}}
		& \r{2.40}{0.07} & \r{2.72}{0.95} & \r{5.61}{0.36} & \r{3.29}{0.92} 
		& \r{3.50}{0.34}
		& \r{\textbf{34.55}}{0.35} & \r{\textbf{34.44}}{0.20} & \r{\textbf{32.14}}{0.11} & \r{\textbf{31.09}}{0.24}
		& \r{\textbf{33.05}}{0.12}
		\\

		\header{Affine}{\citep{eogs}}
		& \r{\textbf{2.15}}{0.04} & \r{2.50}{0.64} & \r{3.68}{0.13} & \r{2.32}{0.07}
		& \r{2.66}{0.16}
		& \r{33.85}{0.78} & \r{33.74}{0.15} & \r{31.71}{0.33} & \r{30.89}{0.39}
		& \r{32.55}{0.24}
		\\

		RPC (Ours)
		& \r{2.16}{0.06} & \r{\textbf{1.72}}{0.04} & \r{\textbf{3.07}}{0.16} & \r{\textbf{2.24}}{0.25}
		& \r{\textbf{2.30}}{0.08}
		& \r{34.36}{0.53} & \r{33.82}{0.56} & \r{31.72}{0.15} & \r{30.67}{0.53}
		& \r{32.64}{0.24}
		\\

		\bottomrule
	\end{tabular}
	
	\vspace{0.9em}
	
	\textbf{IARPA2016~\citep{IARPA2016}}
	\vspace{0.25em}

	\begin{tabular}{@{}lcccc cccc@{}}
		\toprule
		& \multicolumn{4}{c}{\textbf{MAE [m] $\downarrow$}}
		& \multicolumn{4}{c}{\textbf{PSNR [dB] $\uparrow$}} \\
		\cmidrule(lr){2-5}\cmidrule(lr){6-9}
		Camera Model
		& 001 & 002 & 003 & Mean
		& 001 & 002 & 003 & Mean \\
		\midrule

		\header{Perspective}{\citep{gaussian_splatting}}
		& \r{1.74}{0.02} & \r{1.36}{0.02} & \r{2.30}{0.02}
		& \r{1.80}{0.01}
		& \r{30.05}{0.46} & \r{\textbf{32.23}}{0.49} & \r{\textbf{32.44}}{0.08}
		& \r{\textbf{31.57}}{0.23}
		\\

		\header{Affine}{\citep{eogs}}
		& \r{1.73}{0.08} & \r{\textbf{1.25}}{0.01} & \r{2.22}{0.03}
		& \r{1.73}{0.03}
		& \r{\textbf{30.27}}{0.12} & \r{31.79}{0.28} & \r{\textbf{32.44}}{0.28}
		& \r{31.50}{0.14}
		\\

		RPC (Ours) 
		& \r{\textbf{1.69}}{0.06} & \r{\textbf{1.25}}{0.01} & \r{\textbf{2.20}}{0.02}
		& \r{\textbf{1.71}}{0.02}
		& \r{30.01}{0.11} & \r{31.80}{0.61} & \r{32.20}{0.28}
		& \r{31.33}{0.23}
		\\

		\bottomrule
	\end{tabular}
	
\end{table*}

We evaluate geometry by converting the view-dependent surface depth from each training image into a georegistered, view-agnostic digital surface model (DSM). 
For each pixel, to obtain the surface depth values, the depths of contributing Gaussians are averaged according to their alpha-compositing render weights~\citep{eogs}.
Geometry is therefore evaluated across the original off-nadir image observations, yielding a multi-view assessment of the reconstruction.
We measure geometric accuracy using altitude mean absolute error (MAE) against LiDAR-based DSM ground-truth. Following common practice~\citep{eonerf,eogs,shadowgs}, we exclude water regions in each scene using manually created masks.
For image reconstruction quality, we report peak signal-to-noise ratio (PSNR) between rendered and reference images.

As shown in \cref{tab:results_both}, 
our native RPC renderer outperforms the reconstruction accuracy of camera-approximation-based approaches used by previous work.
On DFC2019~\citep{dfc2019}, our method achieves a mean MAE of 2.30 m across four scenes, compared with 3.50 m and 2.66 m for perspective and affine approximations, corresponding to relative reductions of 34.3\% and 13.5\%, respectively.
On IARPA2016~\citep{IARPA2016}, the improvements are smaller, with an MAE of 1.71m compared with 1.80m (5.0\%) and 1.73m (1.2\%) for perspective and affine approximations, respectively.
These improvements in geometric accuracy are achieved while maintaining competitive image reconstruction quality, with mean PSNR within 0.41 dB of the best approximation-based model on both datasets.
Qualitative results are shown in \cref{fig:qualitative_main}, with additional examples in Appendix~\cref{fig:app_004,fig:app_214,fig:app_260,fig:app_site1,fig:app_site2,fig:app_site3}.

\subsection{Computational Efficiency}

We report the average training time, iteration time, peak memory usage, and final number of Gaussians for each camera model, computed over $n=12$ runs per method across the four DFC2019~\mbox{\citep{dfc2019}} scenes. Relative iteration time is normalized to the perspective-camera baseline, i.e., the standard 3DGS~\citep{gaussian_splatting} implementation.

Our CUDA-optimized implementation achieves a relative iteration time of $1.13\times$ compared with perspective 3DGS, closely matching the $1.12\times$ of the evaluated affine rendering implementation~\citep{eogs}. Thus, despite the additional complexity of RPC-native rendering, our CUDA renderer operates at near-baseline efficiency, introducing only modest computational overhead relative to the perspective and affine baselines.

For ease of development and comparison, we additionally provide a PyTorch implementation, which requires $2.25\times$ the baseline iteration time, primarily due to its autograd-based gradient computation. For both implementations, the increased peak memory usage results from precomputing and retaining per-pixel RPC viewing rays on the GPU to support depth sorting during rasterization.

\begin{table*}[t]
	\centering
	\renewcommand{\r}[2]{%
		\makecell{#1 {\scriptsize $\pm$#2}}%
	}
	\caption{%
	Computational cost comparison across camera models and implementations on the four DFC2019~\citep{dfc2019} scenes. Results are reported as the mean and standard deviation over $n=12$ runs per method.
	RPC-native rendering introduces limited additional computational costs, with
	our CUDA-optimized implementation requiring $1.13\times$ the iteration time
	of perspective 3DGS~\citep{gaussian_splatting} and remaining comparable to
	affine rendering~\citep{eogs}.
	}
	\label{tab:computational_efficiency}
	\setlength{\tabcolsep}{5pt}
	\begin{tabular}{@{}lccccc@{}}
		\toprule
		\textbf{Method}
			& \makecell{\textbf{Train time}\\[-1pt]\textbf{(h:mm:ss)}}
			& \makecell{\textbf{Iter. time}\\[-1pt]\textbf{(ms)}}
			& \makecell{\textbf{Rel. iter.}\\[-1pt]\textbf{time}}
			& \makecell{\textbf{Peak GPU}\\[-1pt]\textbf{memory (GB)}}
			& \makecell{\textbf{Number of}\\[-1pt]\textbf{Gaussians}} \\
		\midrule
		\makecell[l]{Perspective~\tiny{\citep{gaussian_splatting}}}
				& \r{0:35:15}{26s}
				& \r{66.6}{0.8}
				& 1.00$\times$
				& \r{4.05}{0.01}
				& \r{702,784}{3,656}\\
		\makecell[l]{Affine~\tiny{\citep{eogs}}}
				& \r{0:39:13}{14s}
				& \r{74.3}{0.4}
				& 1.12$\times$
				& \r{4.31}{0.02}
				& \r{788,987}{9,843}\\
		RPC (PyTorch)
				& \r{1:16:59}{34s}
				& \r{150.0}{1.1}
				& 2.25$\times$
				& \r{6.48}{0.02}
				& \r{750,266}{7,234}\\
		RPC (CUDA)
				& \r{0:39:39}{27s} 
				& \r{75.3}{0.9}
				& 1.13$\times$
				& \r{5.83}{0.02}
				& \r{749,863}{7,509}\\
		\bottomrule
	\end{tabular}
\end{table*}

\section{Limitations and Future Work}
\label{sec:limitations}
Gaussian splatting typically uses Jacobian-based covariance projection,
which locally approximates nonlinear mappings as affine transformations. In our case, this approximation is applied to both the ECEF $\rightarrow$ geodetic transform and the RPC projection, and may introduce errors when first-order propagation is insufficient. 
Recent alternatives such as the Unscented Transform \cite{3DGUT} could provide more accurate uncertainty propagation and are a promising direction for future work.

Our experiments are designed to isolate the effect of the camera model rather than to maximize reconstruction performance through satellite-specific extensions. Accordingly, apart from the appearance embeddings described in \cref{sec:implementation}, we retain the standard 3DGS optimization setup, introduce no additional loss terms or geometric regularizers, and do not explicitly model effects such as cast shadows.
Combining the proposed RPC-native renderer with shadow-aware appearance models or satellite-specific geometric regularization is therefore a natural direction for future work.

\section{Conclusion}
In this work, we introduce RPC-GS, the first native RPC Gaussian Splatting renderer for satellite imagery. RPC-GS embeds the RPC model in a suitable geo-coordinate transformation chain that maps splatting-compatible normalized scene coordinates to image coordinates. We derive a numerically robust Jacobian-based covariance projection for this RPC-native setting and introduce a metric ray-based depth formulation for RPC-consistent alpha compositing.
We provide the first unified Gaussian Splatting benchmark comparing RPC, perspective, and affine camera models for satellite imagery. RPC-GS achieves the lowest altitude reconstruction error, reducing mean altitude error over perspective and affine approximations by 34.3\% and 13.5\% on DFC2019, and by 5.0\% and 1.2\% on IARPA2016, respectively.

\newlength{\qualimg}
\setlength{\qualimg}{0.193\textheight}

\newcommand{\qpanel}[3]{%
	\begin{subfigure}[c]{\qualimg}
		\centering
		\begin{overpic}[
			width=\qualimg,
			height=\qualimg,
			keepaspectratio
			]{#1}
			#3
		\end{overpic}
		\phantomsubcaption
		\label{#2}
	\end{subfigure}%
}

\newcommand{\qmissing}[1]{%
	\begin{subfigure}[c]{\qualimg}
		\centering
		\raisebox{16.5ex}{%
			\begingroup
			\setlength{\fboxsep}{0pt}%
			\setlength{\fboxrule}{0.35pt}%
			\fbox{%
				\begin{minipage}[c][\dimexpr\qualimg-2\fboxrule\relax][c]
					{\dimexpr\qualimg-2\fboxrule\relax}
					\centering
					\footnotesize
					\textit{#1}
				\end{minipage}%
			}%
			\endgroup
		}%
		\phantomsubcaption
	\end{subfigure}%
}

\newcommand{\rowlabel}[1]{%
	\makebox[1.1em][c]{%
		\raisebox{-7.5ex}{%
			\rotatebox[origin=c]{90}{\small\bfseries #1}%
		}%
	}%
}

\newlength{\detailLineWidth}
\setlength{\detailLineWidth}{0.7pt}

\newcommand{\detailColor}{yellow}

\newcommand{\setdetailstyle}[2]{%
	\setlength{\detailLineWidth}{#1}%
	\renewcommand{\detailColor}{#2}%
}

\newcommand{\detailbox}[4]{%
	\linethickness{\detailLineWidth}%
	\put(#1,#2){\color{\detailColor}\framebox(#3,#4){}}%
}
\newcommand{\detailone}{%
	\detailbox{14}{45}{12}{12}%
}
\newcommand{\detailtwo}{%
	\detailbox{33}{53}{22}{15}%
}
\newcommand{\detailthree}{%
}

\newcommand{\detaildsmone}{%
	\detailbox{57}{60}{40}{30}%
}
\newcommand{\detaildsmtwo}{%
	\detailbox{6}{35}{20}{20}%
}

\begin{figure*}[ht]
	\centering
	\setlength{\tabcolsep}{4pt}
	\renewcommand{\arraystretch}{0.9}
	
	\begin{tabular}{
			@{}
			>{\centering\arraybackslash}m{1.1em}
			@{\hspace{2pt}}
			>{\centering\arraybackslash}m{\qualimg}
			@{\hspace{4pt}}
			>{\centering\arraybackslash}m{\qualimg}
			@{\hspace{4pt}}
			>{\centering\arraybackslash}m{\qualimg}
			@{}
		}
		& \textbf{RGB} & \textbf{Altitude} & \textbf{Evaluation DSM} \\[0.1em]
		
		\rowlabel{GT}
		& \qpanel{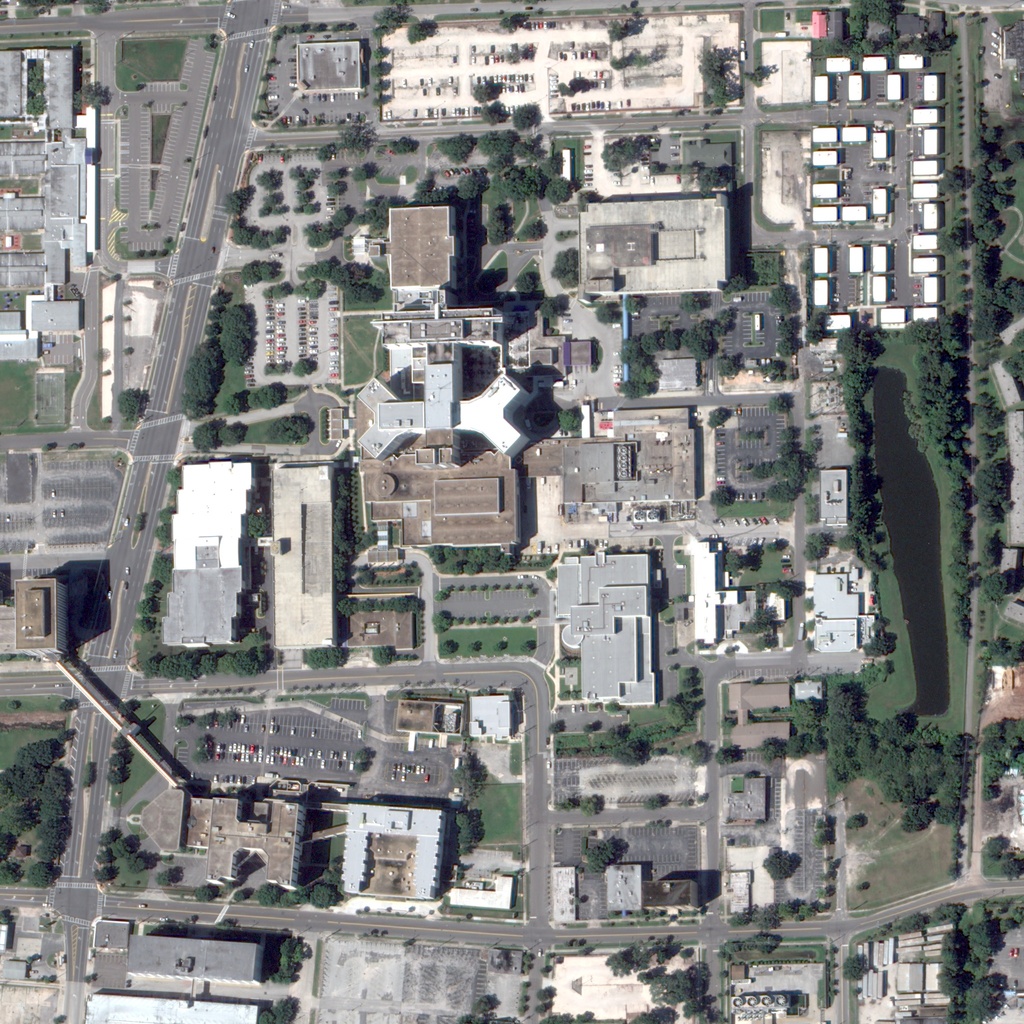}{fig:gt_rgb}{}
		& \qmissing{No view-specific GT available}{}
		& \qpanel{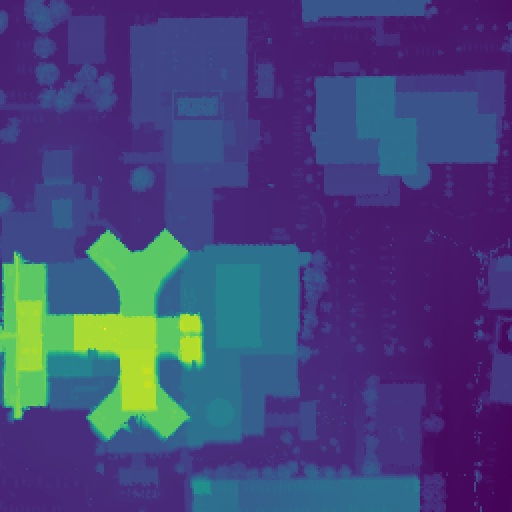}{fig:gt_dsm}{\detaildsmone\detaildsmtwo}
		\\[0.2em]

		\rowlabel{Affine Approx.}
		& \qpanel{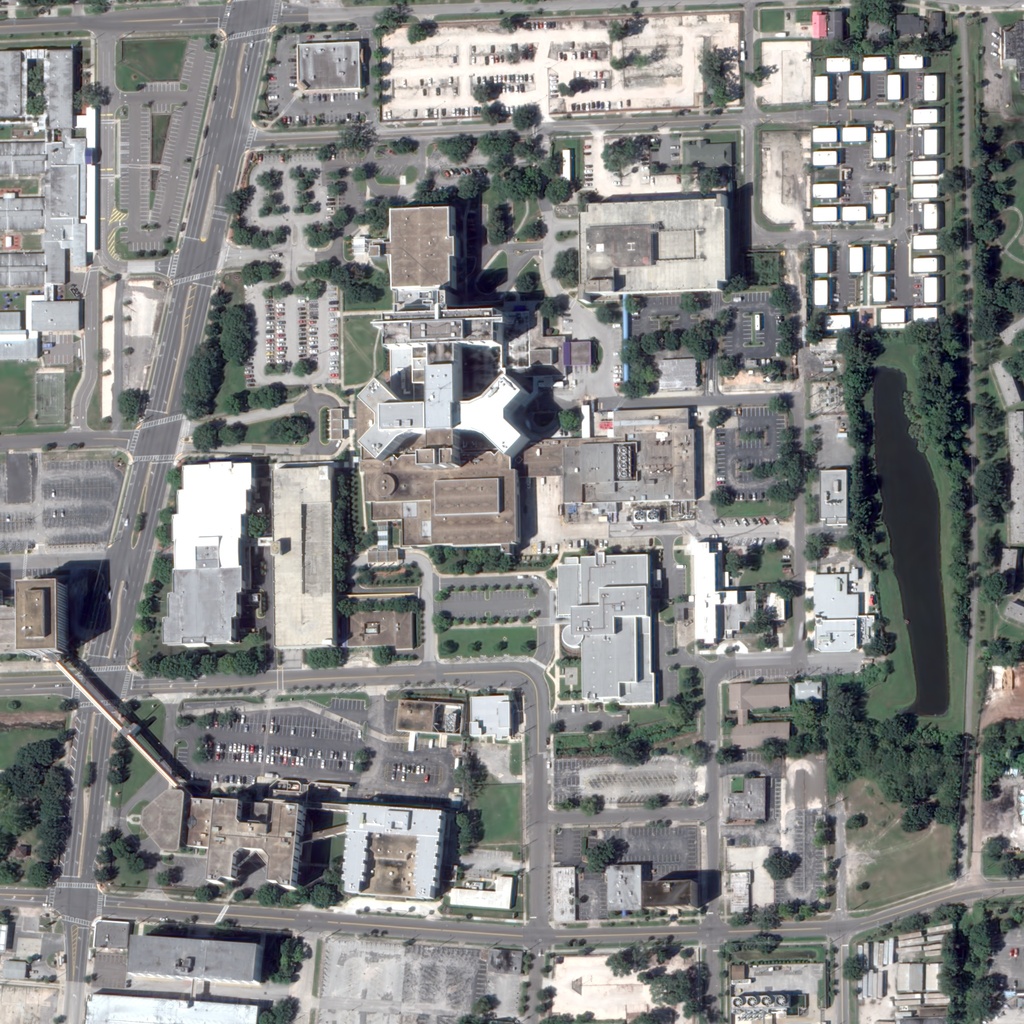}{fig:method_a_rgb}{}
		& \qpanel{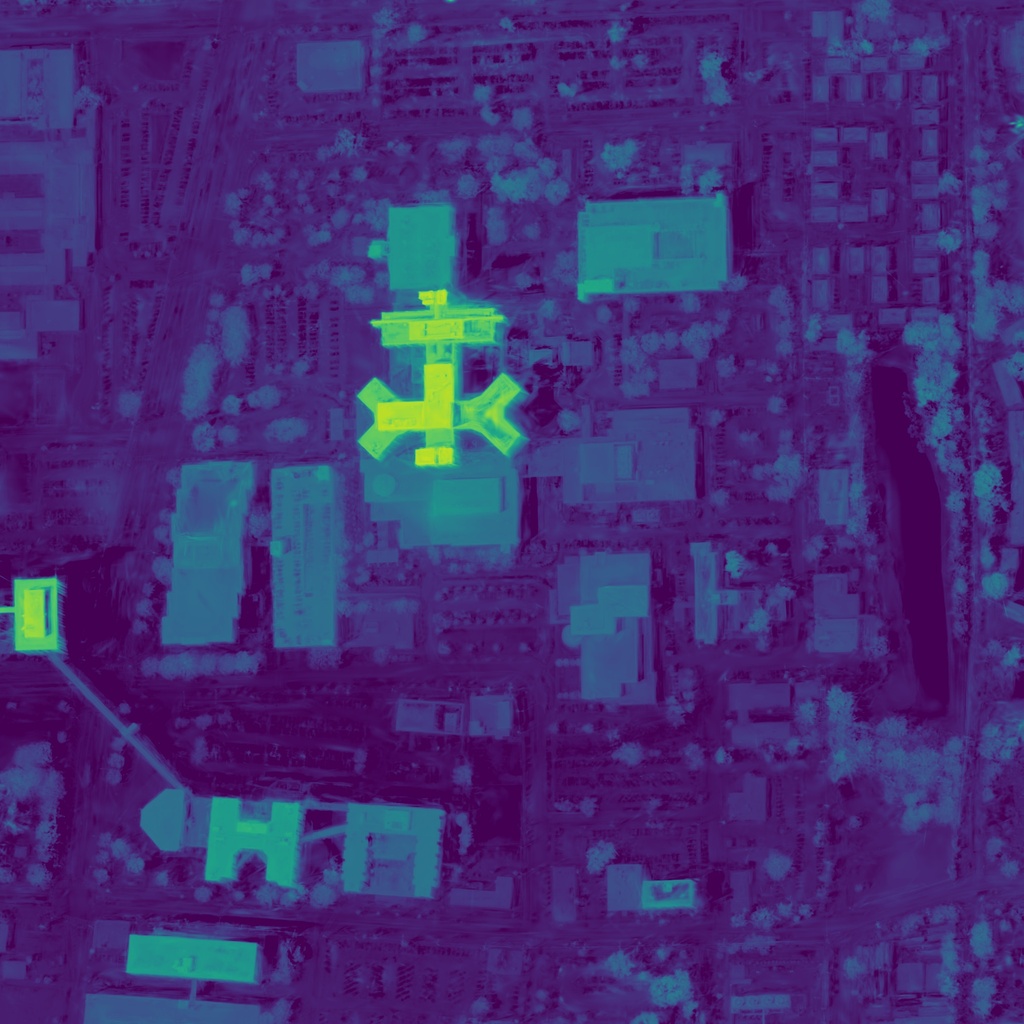}{fig:method_a_depth}{\detailone\detailtwo\detailthree}
		& \qpanel{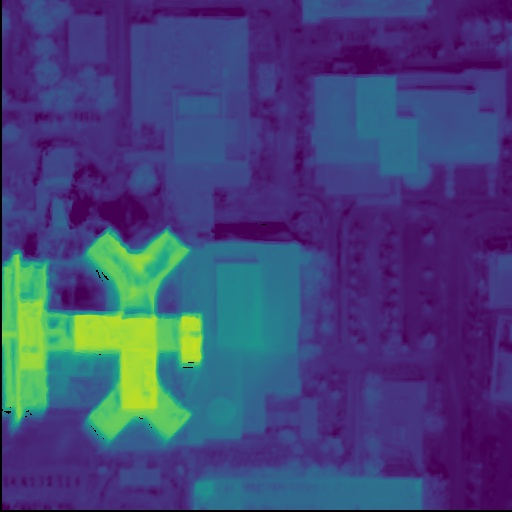}{fig:method_a_dsm}{\detaildsmone\detaildsmtwo}
		\\[0.2em]
		
		\rowlabel{Perspective Approx.}
		& \qpanel{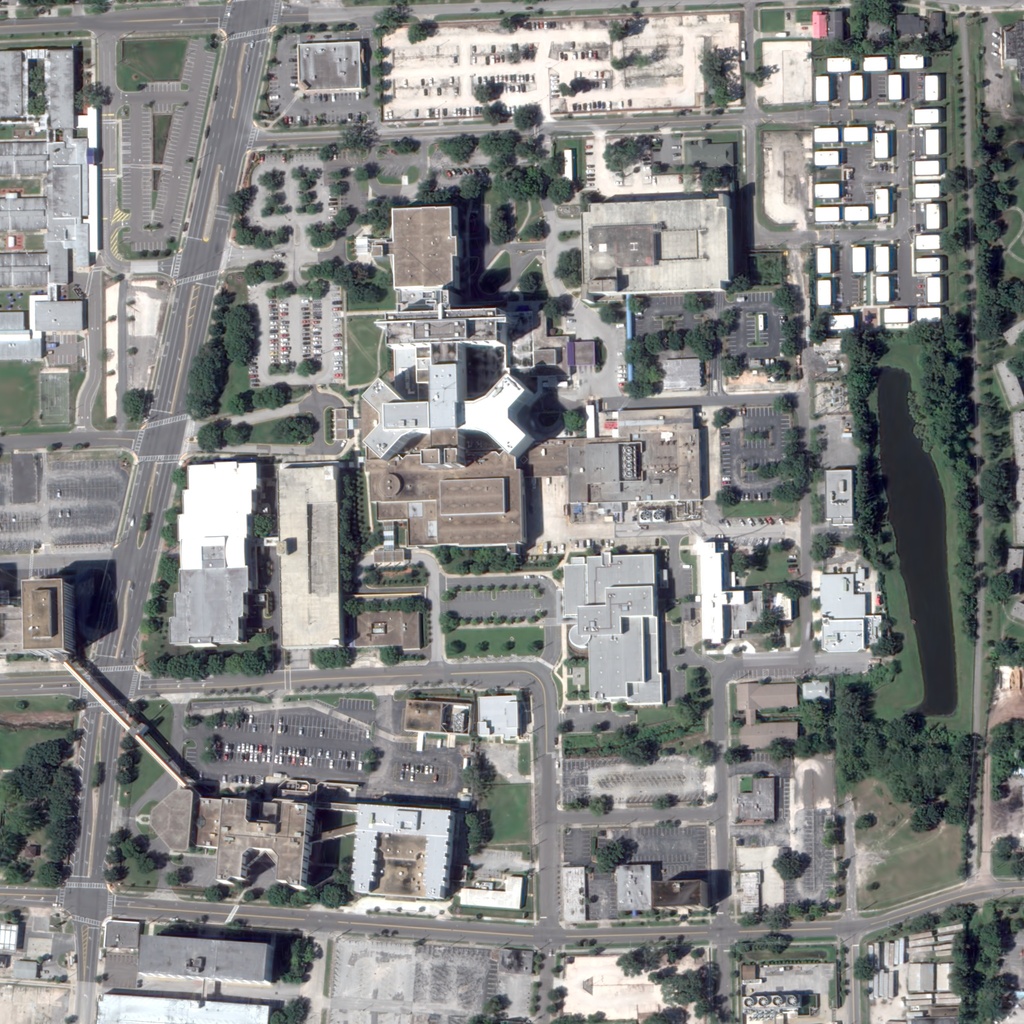}{fig:method_a_rgb}{}
		& \qpanel{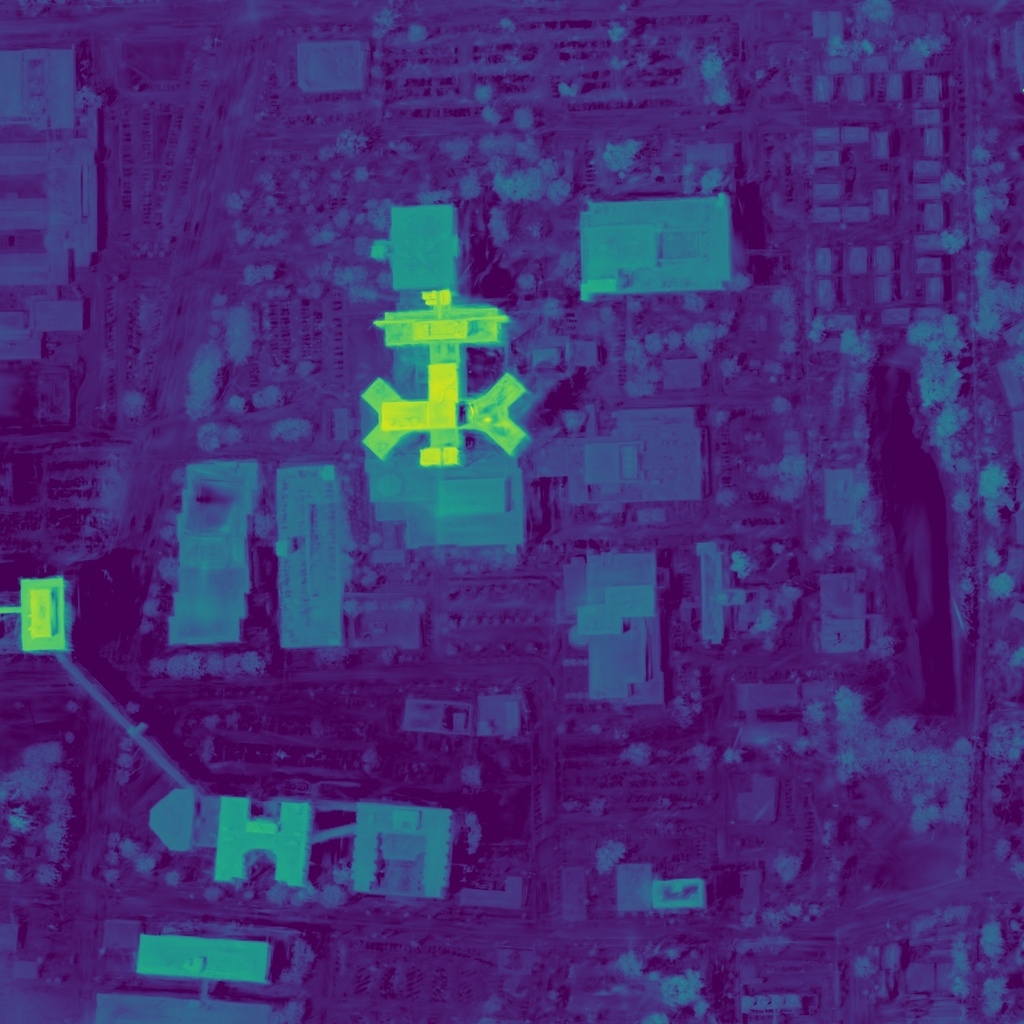}{fig:method_a_depth}{\detailone\detailtwo\detailthree}
		& \qpanel{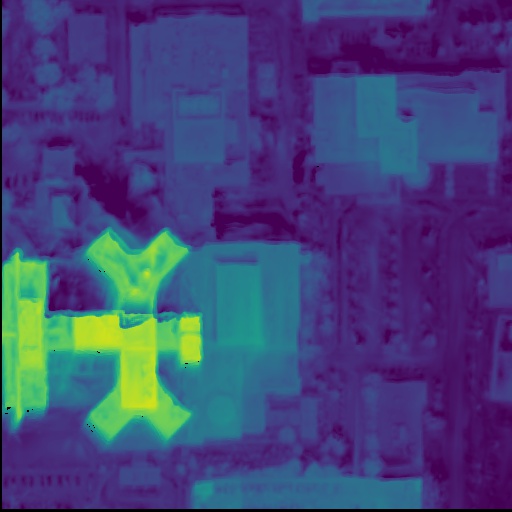}{fig:method_a_dsm}{\detaildsmone\detaildsmtwo}
		\\[0.2em]
		
		\rowlabel{Native RPC (Ours)}
		& \qpanel{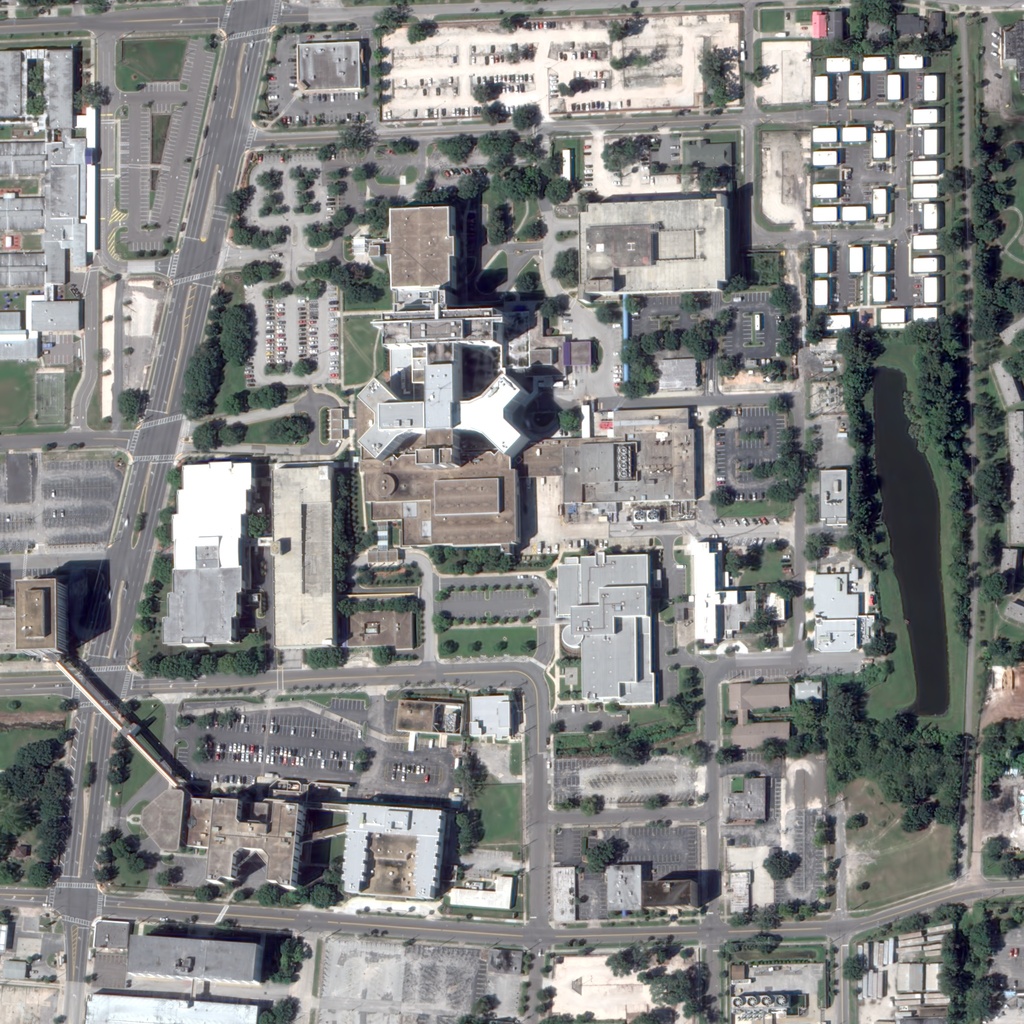}{fig:ours_rgb}{}
		& \qpanel{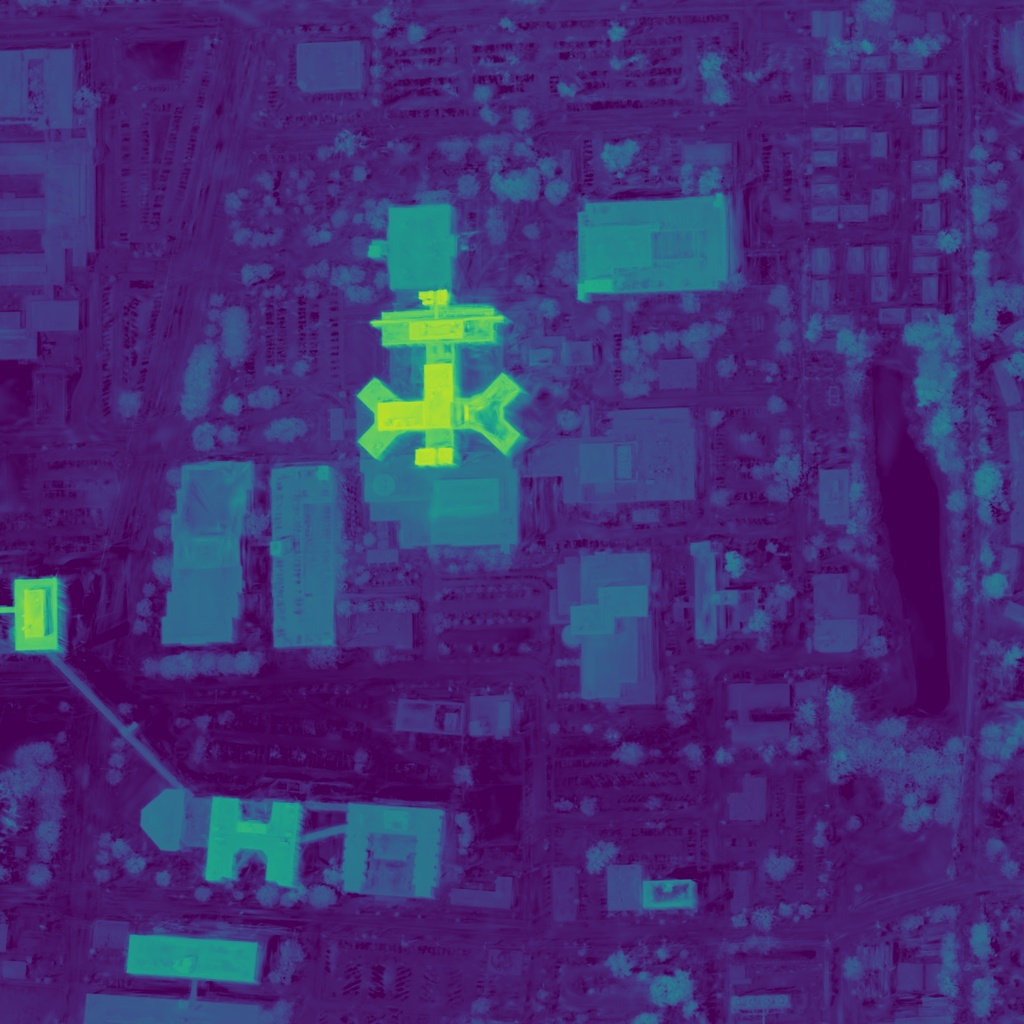}{fig:ours_depth}{\detailone\detailtwo\detailthree}
		& \qpanel{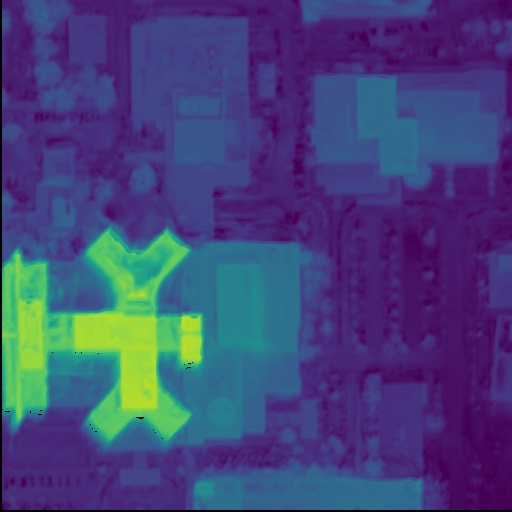}{fig:ours_dsm}{\detaildsmone\detaildsmtwo}
	\end{tabular}
	
	\caption{
		Qualitative comparison of scene JAX\_068 from the DFC2019 dataset~\cite{dfc2019}. 
		We convert rendered depth to altitude to enable cross-camera-model comparison. 
		The evaluation DSM indicates the LiDAR-covered region used for altitude error computation. 
		Our native RPC renderer preserves fine image structure and yields sharper, more consistent geometry than perspective and affine approximations. 
		Areas of interest are highlighted in yellow. Colors encode altitude, ranging from blue (low) to yellow (high).
	}
	\label{fig:qualitative_main}
\end{figure*}

\begin{ack}
  This research was partially funded by the German Federal Ministry of Defence. The authors declare no competing interests.
\end{ack}

\bibliographystyle{plainnat}
\bibliography{references}

\appendix

\section{Technical appendices and supplementary material}
\label{app:main}
\subsection{Differentiating Geodetic $\rightarrow$ ECEF}
\label{app:diff_geo_ecef}
In this section we provide the step-by-step differentiation of the forward Geodetic $\rightarrow$ ECEF map as seen in \cref{eq:jac_geo_to_ecef}, which is restated here for ease of reference.

The forward map as defines the conversion from geodetic coordinates $(\lambda, \phi, h)$ to ECEF coordinates $(x,y,z)$:
\begin{equation}
	f(\lambda,\phi,h)
	=
	\begin{bmatrix}
		(N_{\phi}+h)\cos\phi\cos\lambda \\
		(N_{\phi}+h)\cos\phi\sin\lambda \\
		\left(N_{\phi}(1-e^2)+h\right)\sin\phi
	\end{bmatrix}
	=
	\begin{bmatrix}
		x\vphantom{(N(\phi)+h)\cos\phi\cos\lambda}\\
		y\vphantom{(N(\phi)+h)\cos\phi\sin\lambda}\\
		z\vphantom{\left(N(\phi)(1-e^2)+h\right)\sin\phi}
	\end{bmatrix}
\end{equation}
We recall the definition of the local orthonormal basis
$(\mathbf{e}_{\lambda},\mathbf{e}_{\phi},\mathbf{e}_{h})$ of \cref{eq:enu_base},
corresponding to the east, north, and ellipsoidal-up axes expressed in ECEF coordinates~\citep{navipedia_ecef_enu}. 
\begin{equation}
	\vecs{e}_{\lambda} =
	\begin{bmatrix}
		-\sin\lambda \\
		\cos\lambda \\
		0
	\end{bmatrix},
	\qquad
	\vecs{e}_{\phi} =
	\begin{bmatrix}
		-\sin\phi\cos\lambda \\
		-\sin\phi\sin\lambda \\
		\cos\phi
	\end{bmatrix},
	\qquad
	\vecs{e}_{h} =
	\begin{bmatrix}
		\cos\phi\cos\lambda \\
		\cos\phi\sin\lambda \\
		\sin\phi
	\end{bmatrix}
\end{equation}
Since $\phi$ and $h$ are held fixed when differentiating with respect to
longitude $\lambda$, and $N_\phi$ depends only on latitude $\phi$, the partial derivative with
respect to longitude $\lambda$ is
\begin{equation}
	\frac{\partial f}{\partial \lambda}
	=
	\begin{bmatrix}
		-(N_\phi + h)\cos\phi\sin\lambda\\
		(N_\phi + h)\cos\phi\cos\lambda\\
		0
	\end{bmatrix}
	=
	\left(N_\phi + h\right)\cos\phi
	\begin{bmatrix}
		-\sin\lambda\\
		\cos\lambda\\
		0
	\end{bmatrix}
	=
	\left(N_\phi + h\right)\cos\phi \vecs{e}_\lambda
\end{equation}
The derivative with respect to latitude $\phi$ is less direct because the
prime-vertical radius of curvature $N_\phi$ also depends on latitude $\phi$. We define $N'_\phi=\frac{\mathrm{d}N_\phi}{\mathrm{d}\phi}$ for readability. Applying the product rule gives:
\begin{equation}
	\frac{\partial f}{\partial \phi}
	=
	\begin{bmatrix}
		\left(N'_\phi\cos\phi-\left(N_\phi+h\right)\sin\phi\right)\cos\lambda\\
		\left(N'_\phi\cos\phi-\left(N_\phi+h\right)\sin\phi\right)\sin\lambda\\
		N'_\phi \left(1-e^2\right)\sin\phi + \left(N_\phi\left(1-e^2\right)+h\right)\cos\phi
	\end{bmatrix}
\end{equation}
We recall the definitions of both the prime-vertical radius $N_\phi$ and the meridionial radius $M_\phi$ of the WGS84 reference ellipsoid \cite{wgs84}. 
\begin{equation}
	N_\phi = 
	\frac{a}{D_\phi^{1/2}},
	\qquad
	M_\phi =
	\frac{a\left(1 - e^2\right)}{D_\phi^{3/2}}
\end{equation}
$D_\phi$ refers to $D_\phi=1-e^2\sin^2\phi$ and $a$ is the semi-major axis radius of the WGS84 ellipsoid.
The derivative of the prime-vertical radius $N'_\phi = \frac{\mathrm{d} N_\phi}{\mathrm{d}\phi}$ is given by
\begin{equation}
	N'_\phi = 
	\frac{\mathrm{d} N_\phi}{\mathrm{d}\phi} =
	a e^2 \sin\phi \cos\phi D_\phi^{-3/2}
\end{equation}
Substituting these expressions and collecting terms gives the following two identities
as purely algebraic consequences of the definitions of
$N_\phi$ and $M_\phi$.
\begin{equation}
N'_\phi\cos\phi - N_\phi\sin\phi
=
-M_\phi\sin\phi,
\qquad
N'_\phi(1-e^2)\sin\phi
+
N_\phi(1-e^2)\cos\phi
=
M_\phi\cos\phi.
\end{equation}
Subsequently, we can reduce the derivation $\frac{\partial f}{\partial \phi}$ to:
\begin{equation}
	\frac{\partial f}{\partial \phi}
	=
	\begin{bmatrix}
		-\left(M_\phi + h\right)\sin\phi\cos\lambda\\
		-\left(M_\phi + h\right)\sin\phi\sin\lambda\\ 
		\left(M_\phi+h\right)\cos\phi
	\end{bmatrix}
	= 
	\left(M_\phi+h\right)
	\vecs{e}_\phi
\end{equation}
Finally, the partial derivative $\frac{\partial f}{\partial h}$ for the height $h$ is given by:
\begin{equation}
	\frac{\partial f}{\partial h}
	=
	\begin{bmatrix}
		\cos\phi\cos\lambda\\
		\cos\phi\sin\lambda\\
		\sin\phi
	\end{bmatrix}
	=
	\vecs{e}_h
\end{equation}
Conceptually, these derivatives show that small longitude increments correspond
to eastward displacements of magnitude $\left(N_\phi+h\right)\cos\phi$, small
latitude increments correspond to northward displacements of magnitude
$\left(M_\phi+h\right)$, and height increments translate directly to equal
displacements along the ellipsoidal-up direction $\vecs{e}_h$.

The full Jacobian of the Geodetic $\rightarrow$ ECEF map, with
columns ordered as $(\lambda,\phi,h)$, is
\begin{equation}
	\begin{aligned}
		\mathbf{J}^{\mathrm{ecef} \leftarrow \mathrm{geodetic}}
		=
		\begin{bmatrix}
			\mathbf{e}_{\lambda} &
			\mathbf{e}_{\phi} &
			\mathbf{e}_{h}
		\end{bmatrix}
		\operatorname{diag}\!\left(
		(N_{\phi}+h)\cos\phi,
		M_{\phi}+h,
		1
		\right)
		\\
		=
		\begin{bmatrix}
			-\left(N_\phi+h\right)\cos\phi\sin\lambda
			&
			-\left(M_\phi+h\right)\sin\phi\cos\lambda
			&
			\cos\phi\cos\lambda
			\\
			\left(N_\phi+h\right)\cos\phi\cos\lambda
			&
			-\left(M_\phi+h\right)\sin\phi\sin\lambda
			&
			\cos\phi\sin\lambda
			\\
			0
			&
			\left(M_\phi+h\right)\cos\phi
			&
			\sin\phi
		\end{bmatrix}
	\end{aligned}	
\end{equation}

\subsection{Additional qualitative evaluation results}
We provide additional qualitative comparisons for the three remaining scenes (\ie ¸JAX\_004, JAX\_214, and JAX\_260) of the DFC2019 dataset~\citep{dfc2019} in \cref{fig:app_004,fig:app_214,fig:app_260} and for all three scenes of the IARPA2016 dataset~\citep{IARPA2016} in \cref{fig:app_site1,fig:app_site2,fig:app_site3}. We select the training view with the lowest MAE. The altitude and corresponding DSM are clipped to a consistent value range across all three camera models to allow for visual comparison. Areas-of-Interest are marked in yellow.

\newcommand{\detailAA}{%
	\detailbox{45}{50}{25}{28}%
}
\newcommand{\detailDSMAA}{%
	\detailbox{64}{10}{15}{35}%
}

\begin{figure*}[ht]
	\centering
	\setlength{\tabcolsep}{4pt}
	\renewcommand{\arraystretch}{0.9}
	
	\begin{tabular}{
			@{}
			>{\centering\arraybackslash}m{1.1em}
			@{\hspace{2pt}}
			>{\centering\arraybackslash}m{\qualimg}
			@{\hspace{4pt}}
			>{\centering\arraybackslash}m{\qualimg}
			@{\hspace{4pt}}
			>{\centering\arraybackslash}m{\qualimg}
			@{}
		}
		& \textbf{RGB} & \textbf{Altitude} & \textbf{Evaluation DSM} \\[0.1em]
		
		\rowlabel{GT}
		& \qpanel{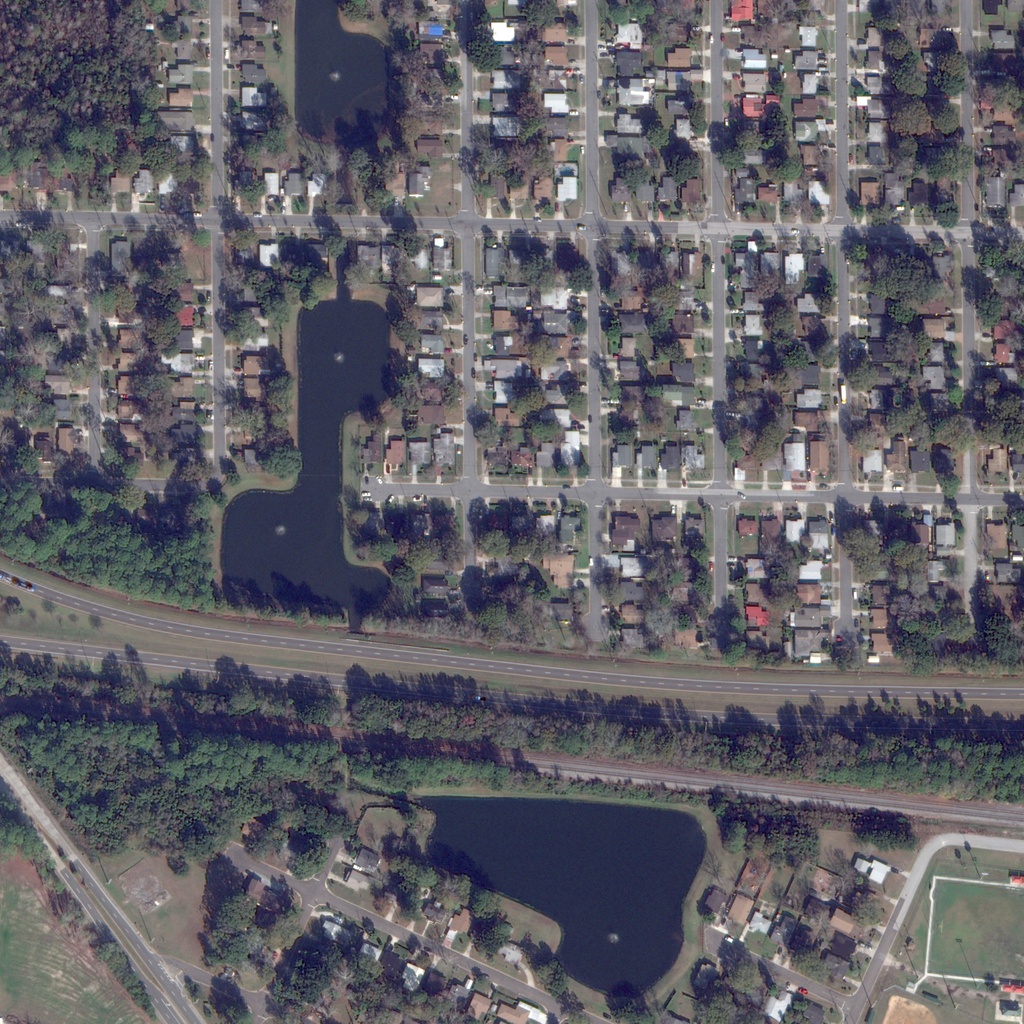}{fig:gt_rgb}{}
		& \qmissing{No view-specific GT available}{}
		& \qpanel{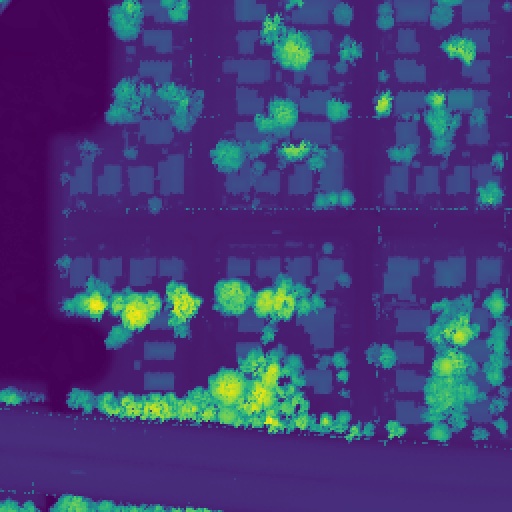}{fig:gt_dsm}{\detailDSMAA}
		\\[0.2em]

		\rowlabel{Affine Approx.}
		& \qpanel{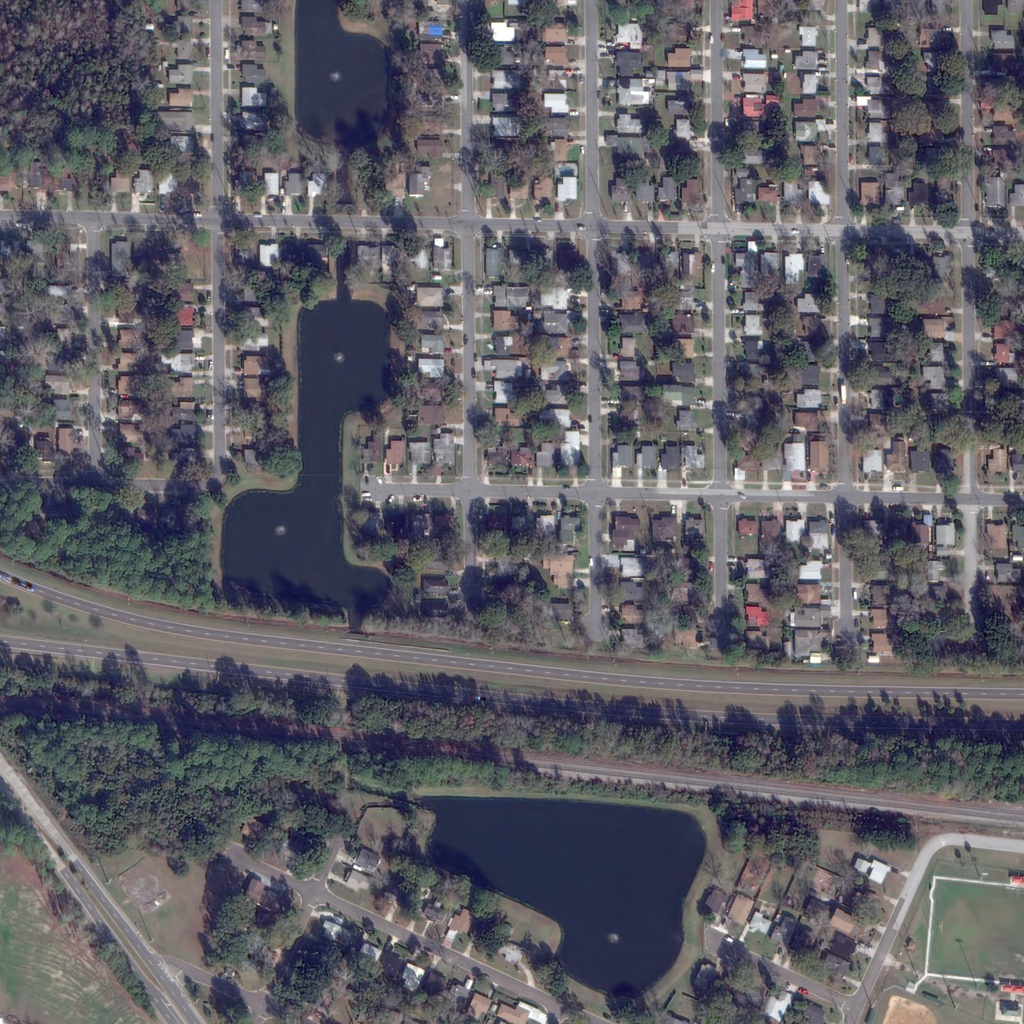}{fig:method_a_rgb}{}
		& \qpanel{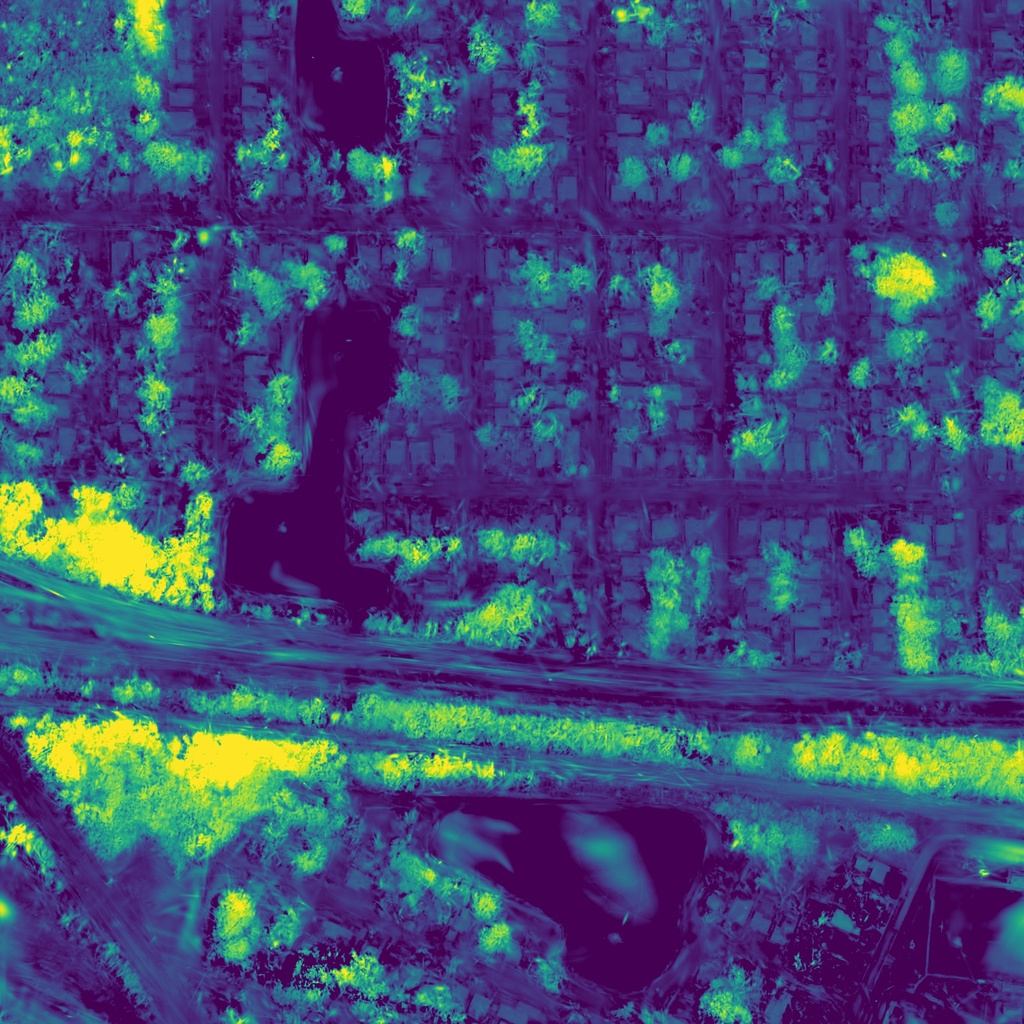}{fig:method_a_depth}{\detailAA}
		& \qpanel{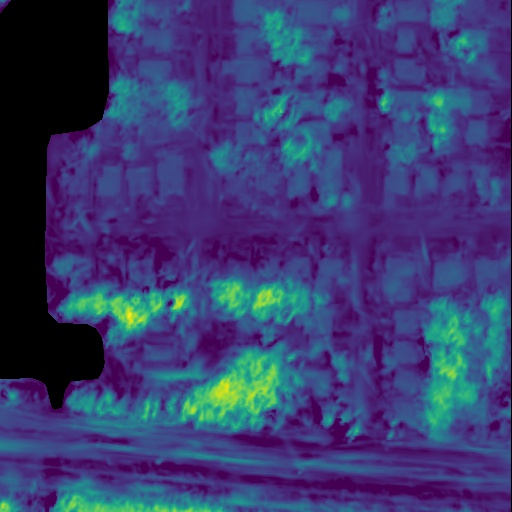}{fig:method_a_dsm}{\detailDSMAA}
		\\[0.2em]
		
		\rowlabel{Perspective Approx.}
		& \qpanel{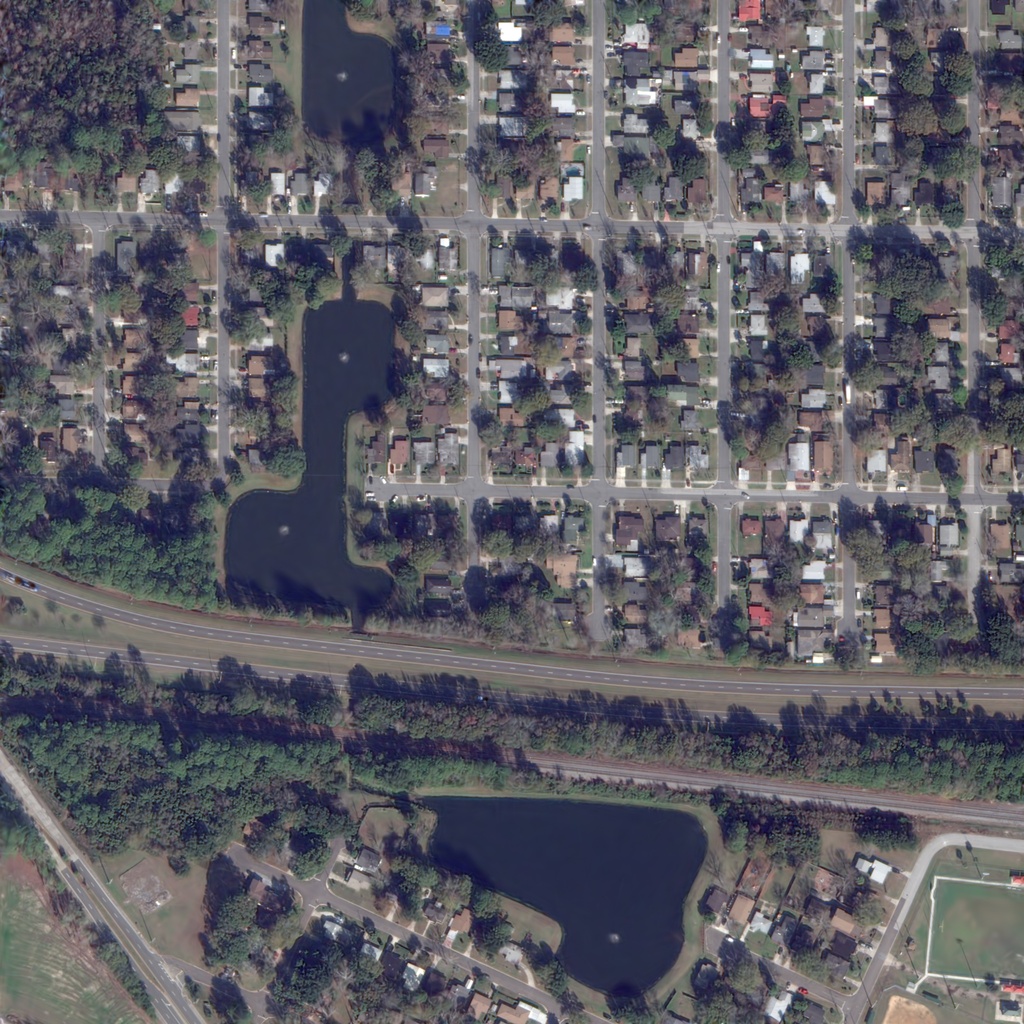}{fig:method_a_rgb}{}
		& \qpanel{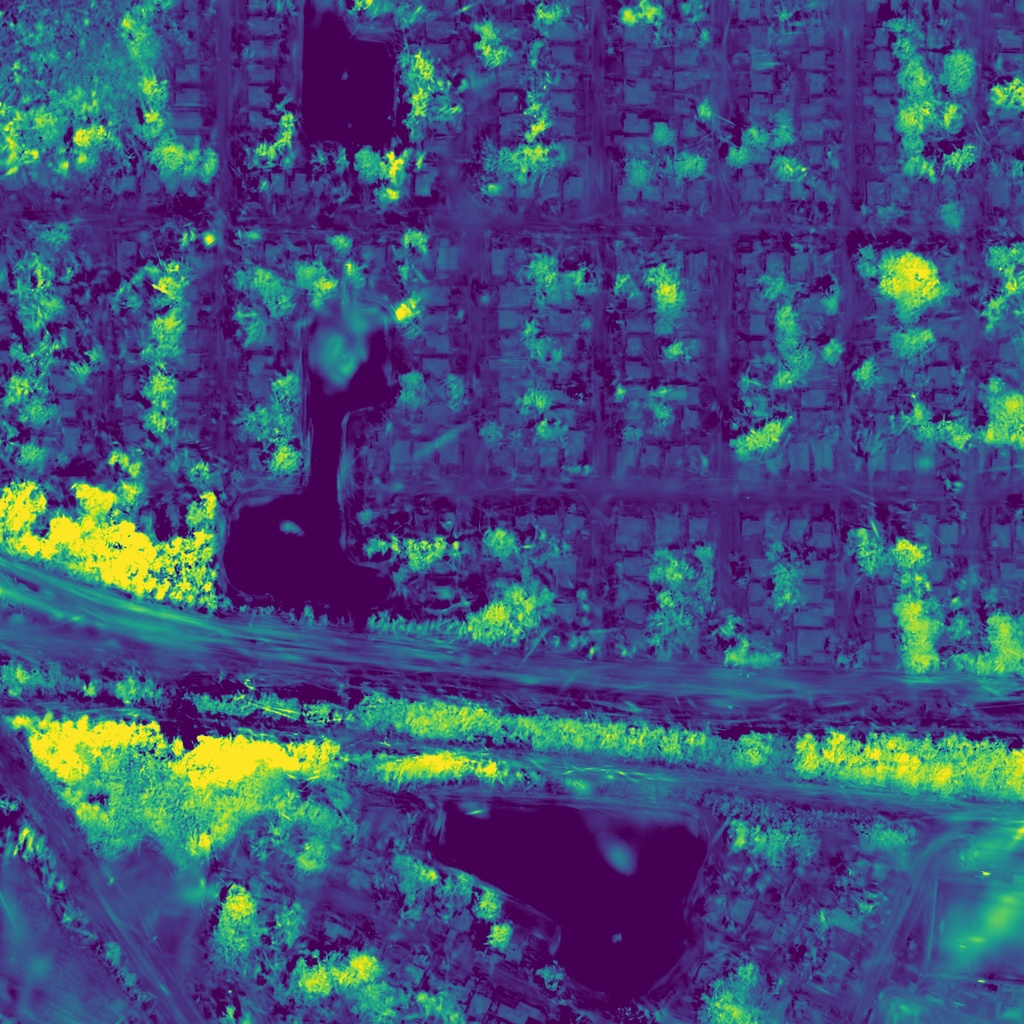}{fig:method_a_depth}{\detailAA}
		& \qpanel{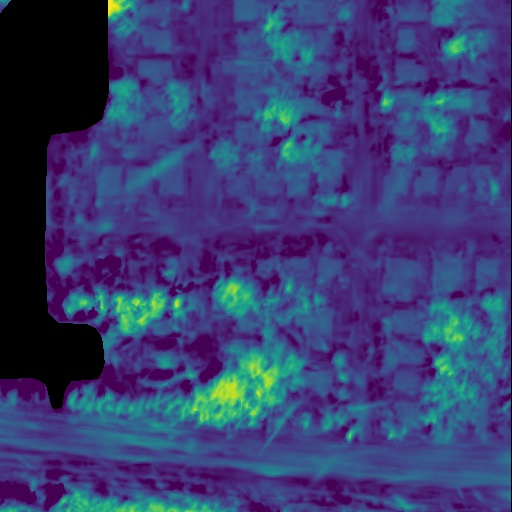}{fig:method_a_dsm}{\detailDSMAA}
		\\[0.2em]
		
		\rowlabel{Native RPC (Ours)}
		& \qpanel{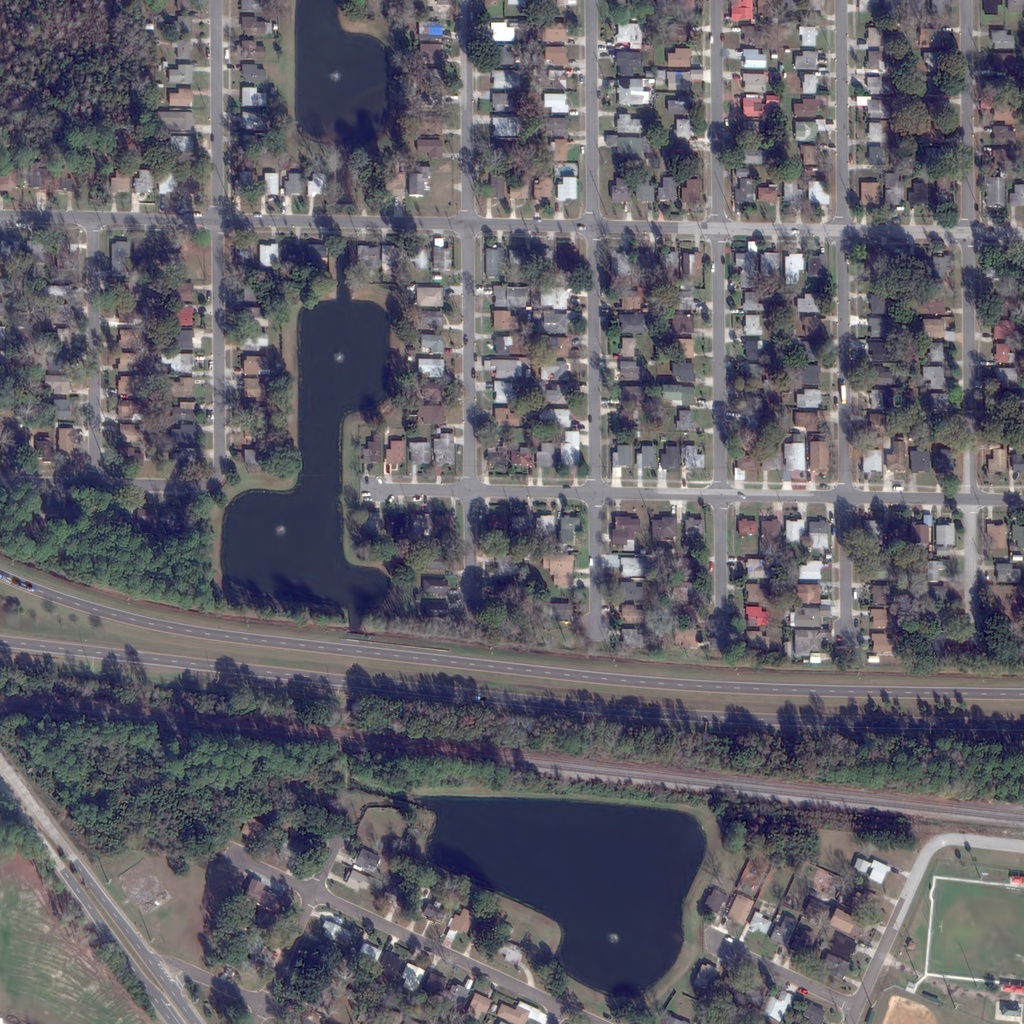}{fig:ours_rgb}{}
		& \qpanel{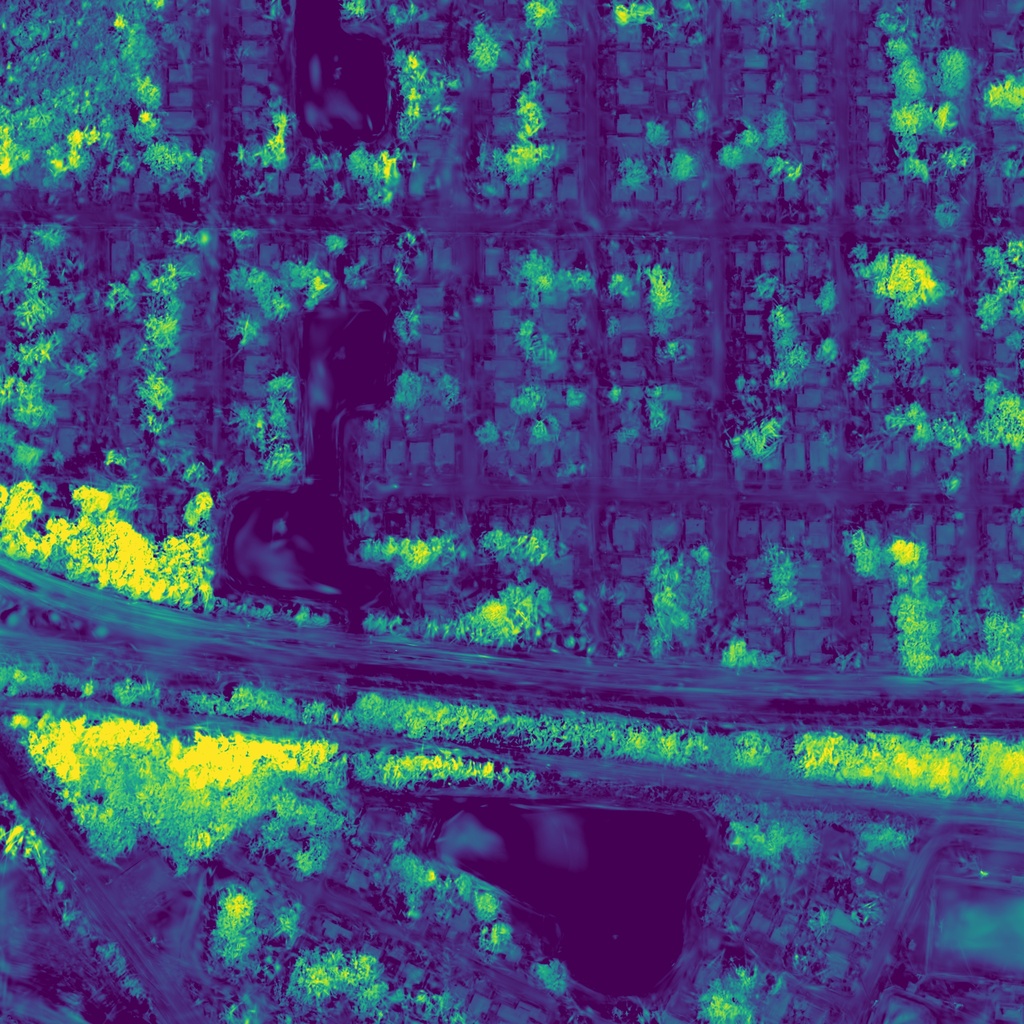}{fig:ours_depth}{\detailAA}
		& \qpanel{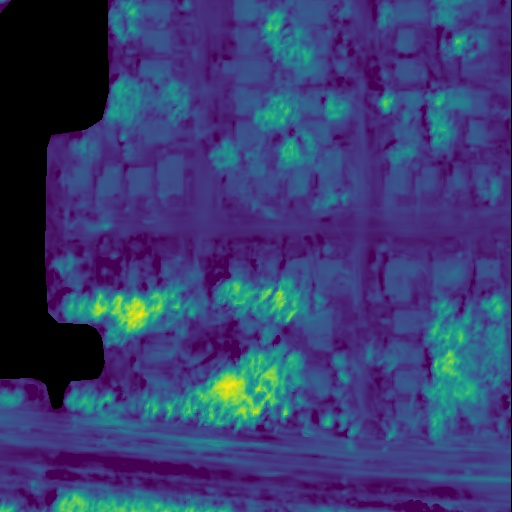}{fig:ours_dsm}{\detailDSMAA}
	\end{tabular}
	
	\caption{
		Qualitative comparison of scene JAX\_004 from the DFC2019 dataset~\cite{dfc2019}. 
		Depth is converted to altitude for cross-camera comparison; the evaluation DSM marks the LiDAR-covered region used for altitude error computation.
		Areas of interest are highlighted in yellow. Colors encode altitude, ranging from blue (low) to yellow (high).
	}
	\label{fig:app_004}
\end{figure*}

\newcommand{\detailCA}{%
}
\newcommand{\detailDSMCA}{%
	\detailbox{14}{0}{22}{18}%
}

\begin{figure*}[ht]
	\centering
	\setlength{\tabcolsep}{4pt}
	\renewcommand{\arraystretch}{0.9}
	
	\begin{tabular}{
			@{}
			>{\centering\arraybackslash}m{1.1em}
			@{\hspace{2pt}}
			>{\centering\arraybackslash}m{\qualimg}
			@{\hspace{4pt}}
			>{\centering\arraybackslash}m{\qualimg}
			@{\hspace{4pt}}
			>{\centering\arraybackslash}m{\qualimg}
			@{}
		}
		& \textbf{RGB} & \textbf{Altitude} & \textbf{Evaluation DSM} \\[0.1em]
		
		\rowlabel{GT}
		& \qpanel{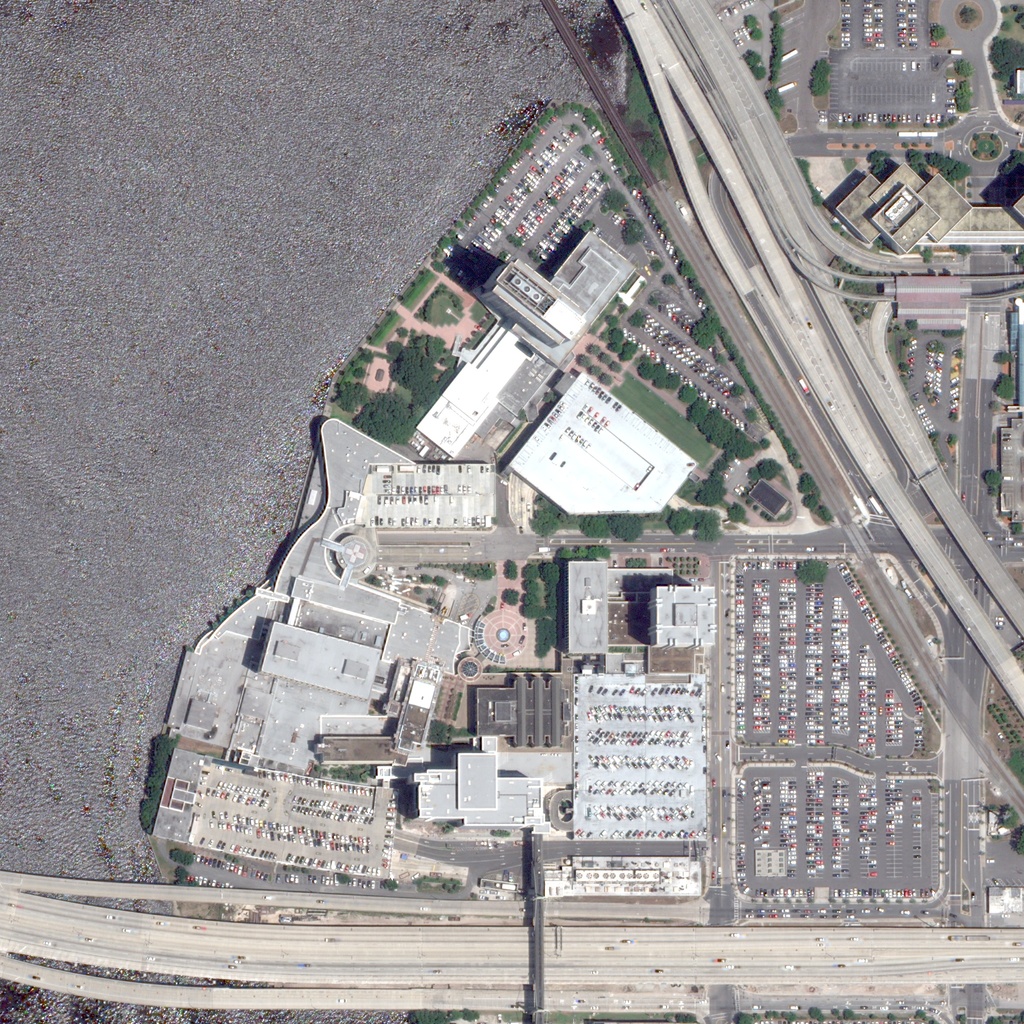}{fig:gt_rgb}{}
		& \qmissing{No view-specific GT available}{}
		& \qpanel{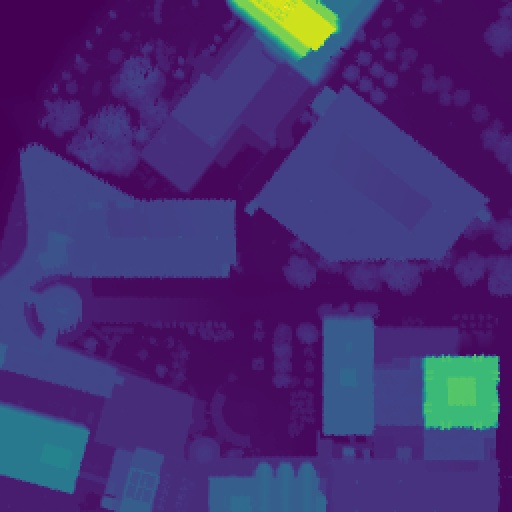}{fig:gt_dsm}{\detailDSMCA}
		\\[0.2em]

		\rowlabel{Affine Approx.}
		& \qpanel{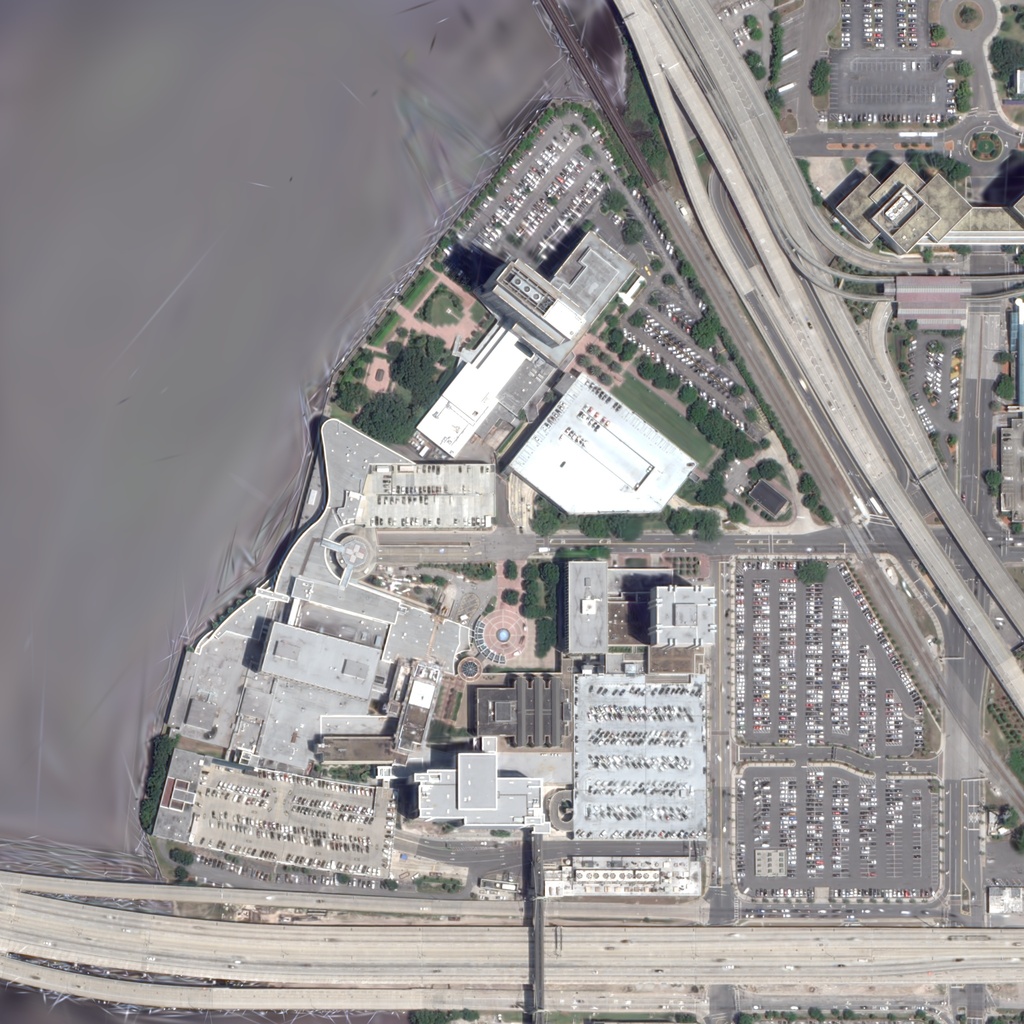}{fig:method_a_rgb}{}
		& \qpanel{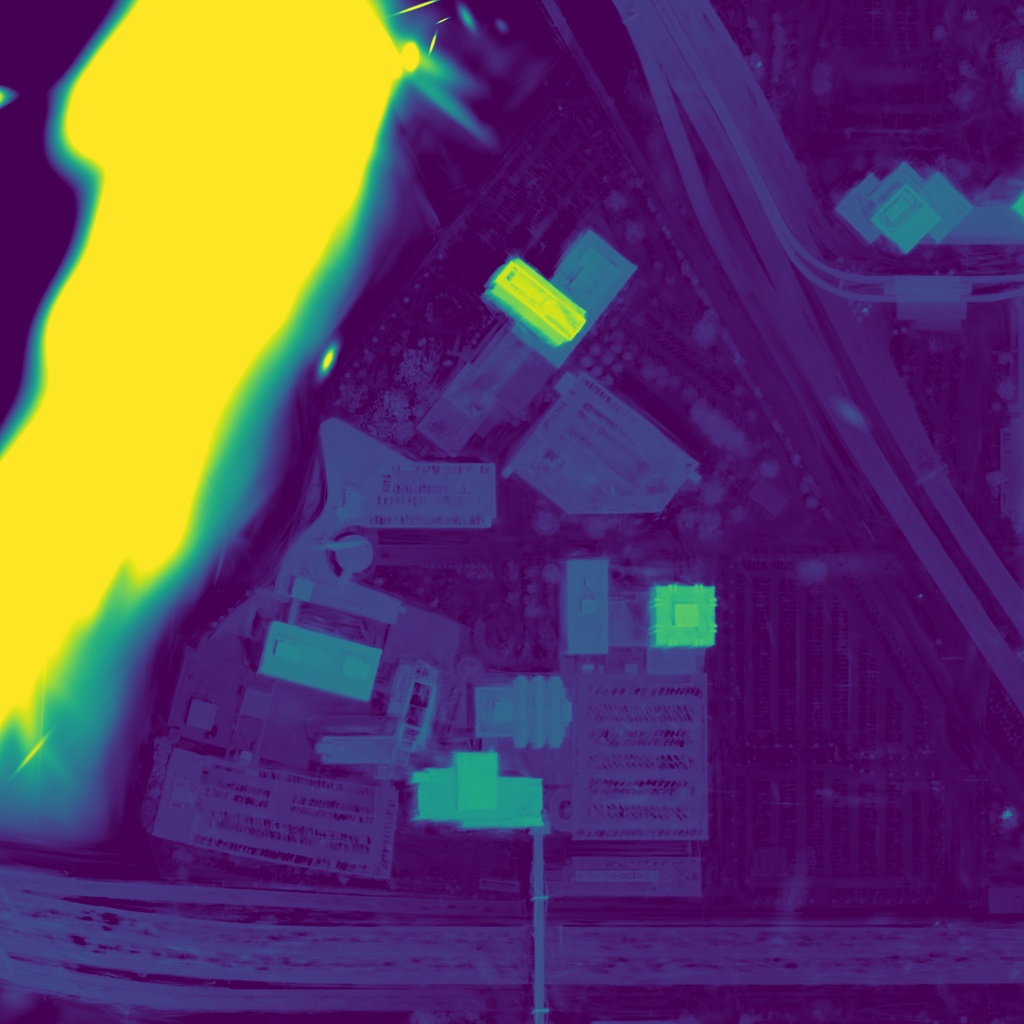}{fig:method_a_depth}{\detailCA}
		& \qpanel{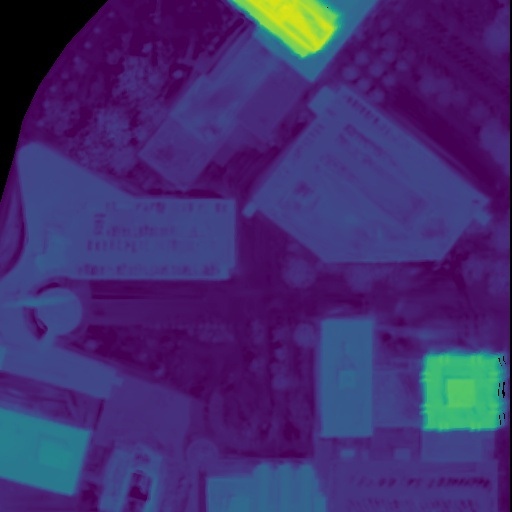}{fig:method_a_dsm}{\detailDSMCA}
		\\[0.2em]
		
		\rowlabel{Perspective Approx.}
		& \qpanel{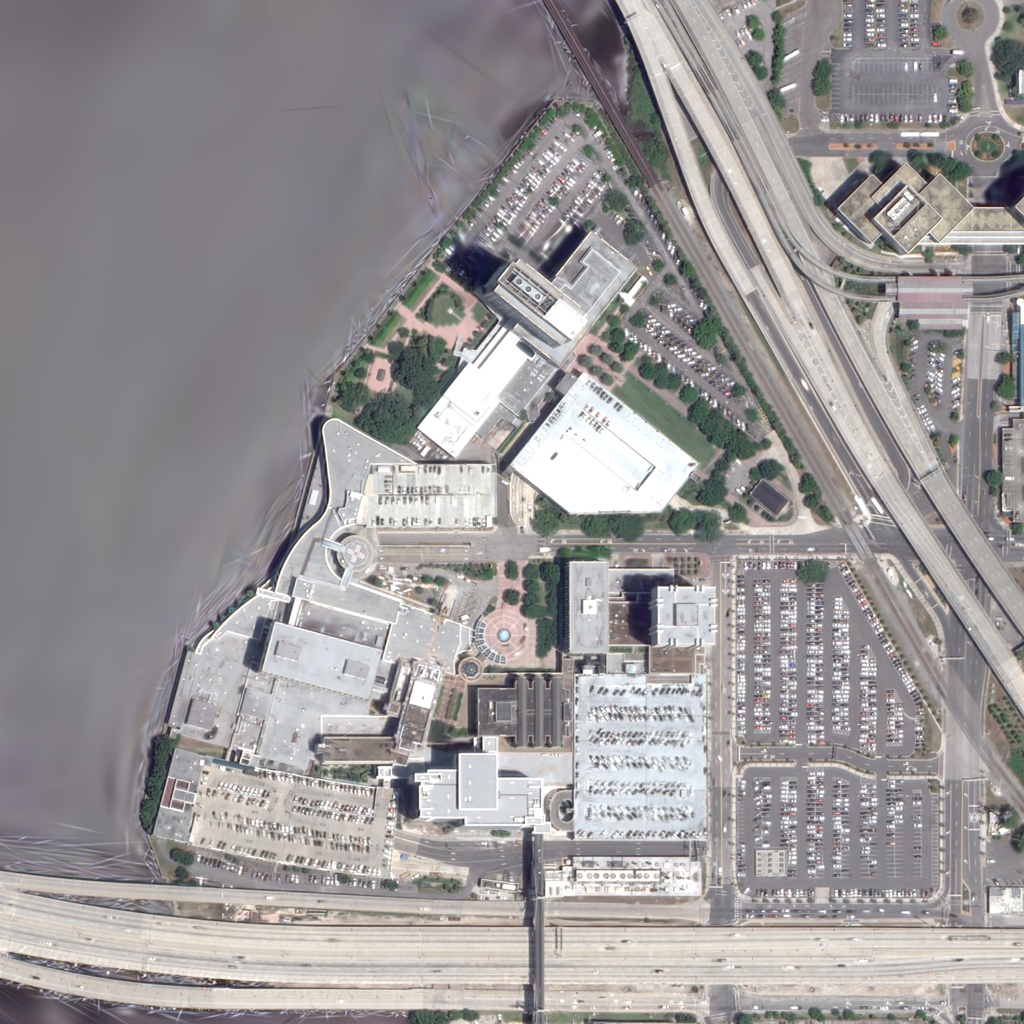}{fig:method_a_rgb}{}
		& \qpanel{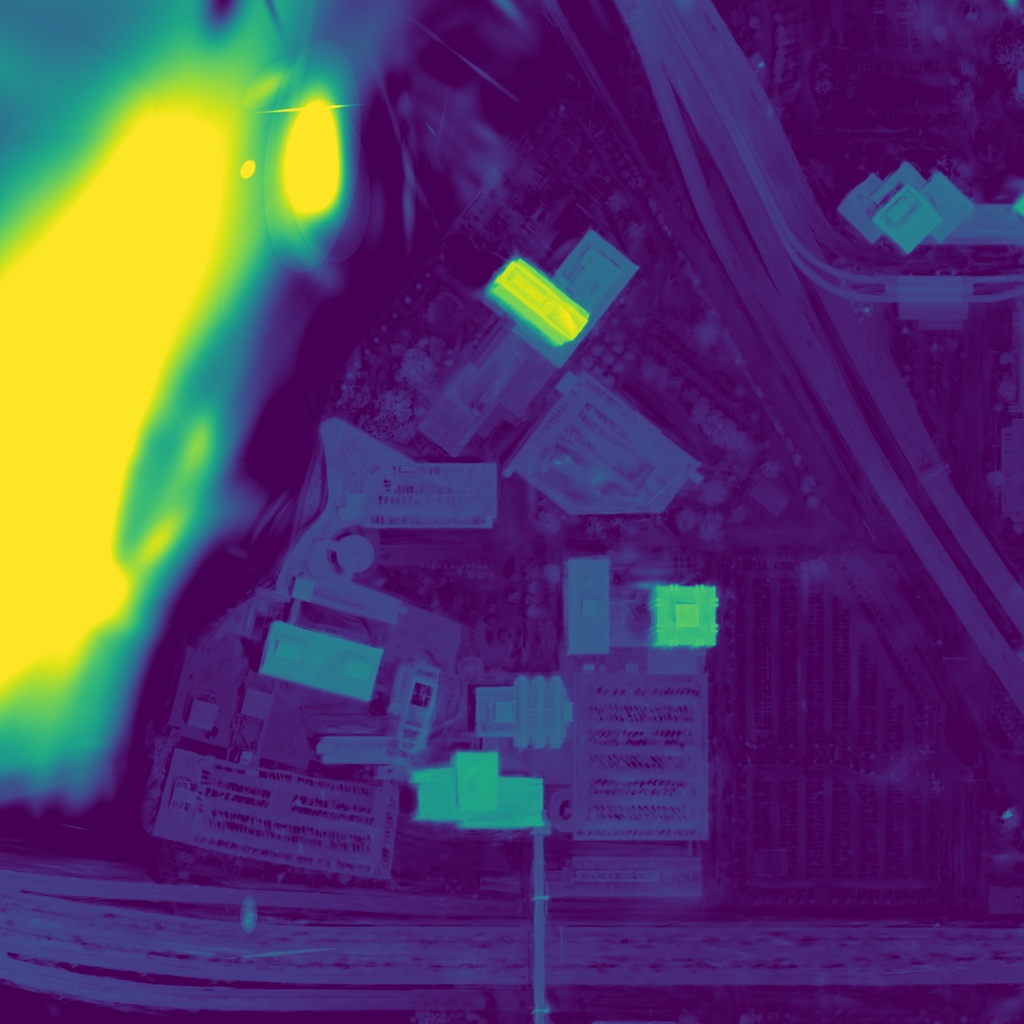}{fig:method_a_depth}{\detailCA}
		& \qpanel{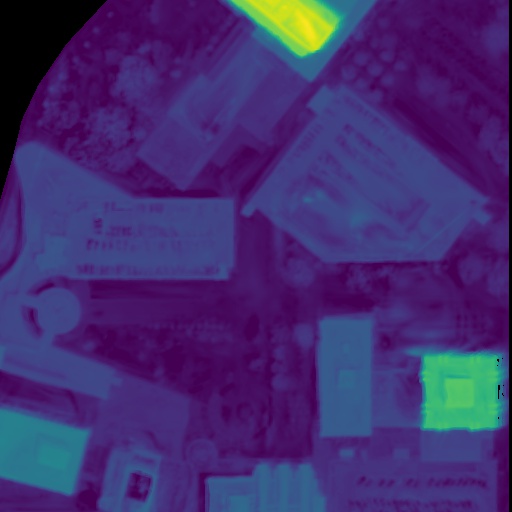}{fig:method_a_dsm}{\detailDSMCA}
		\\[0.2em]
		
		\rowlabel{Native RPC (Ours)}
		& \qpanel{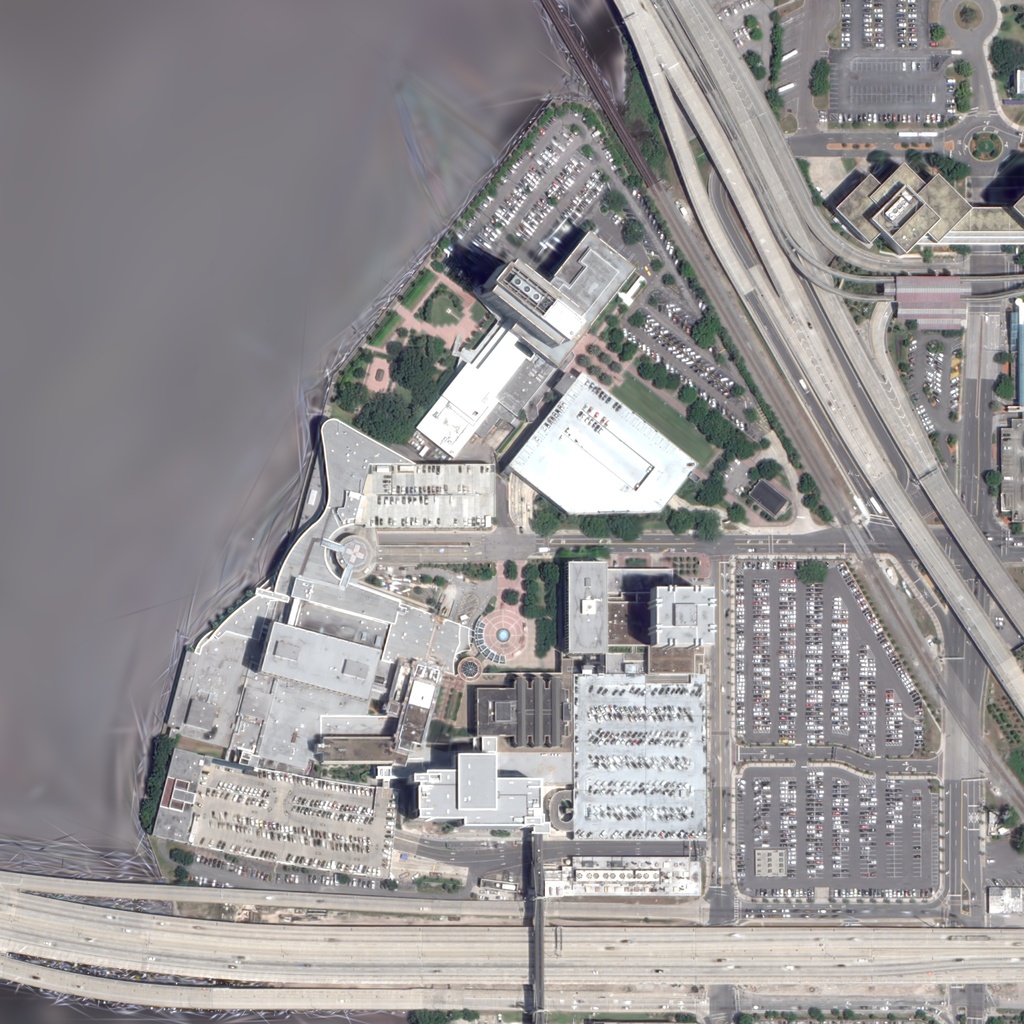}{fig:ours_rgb}{}
		& \qpanel{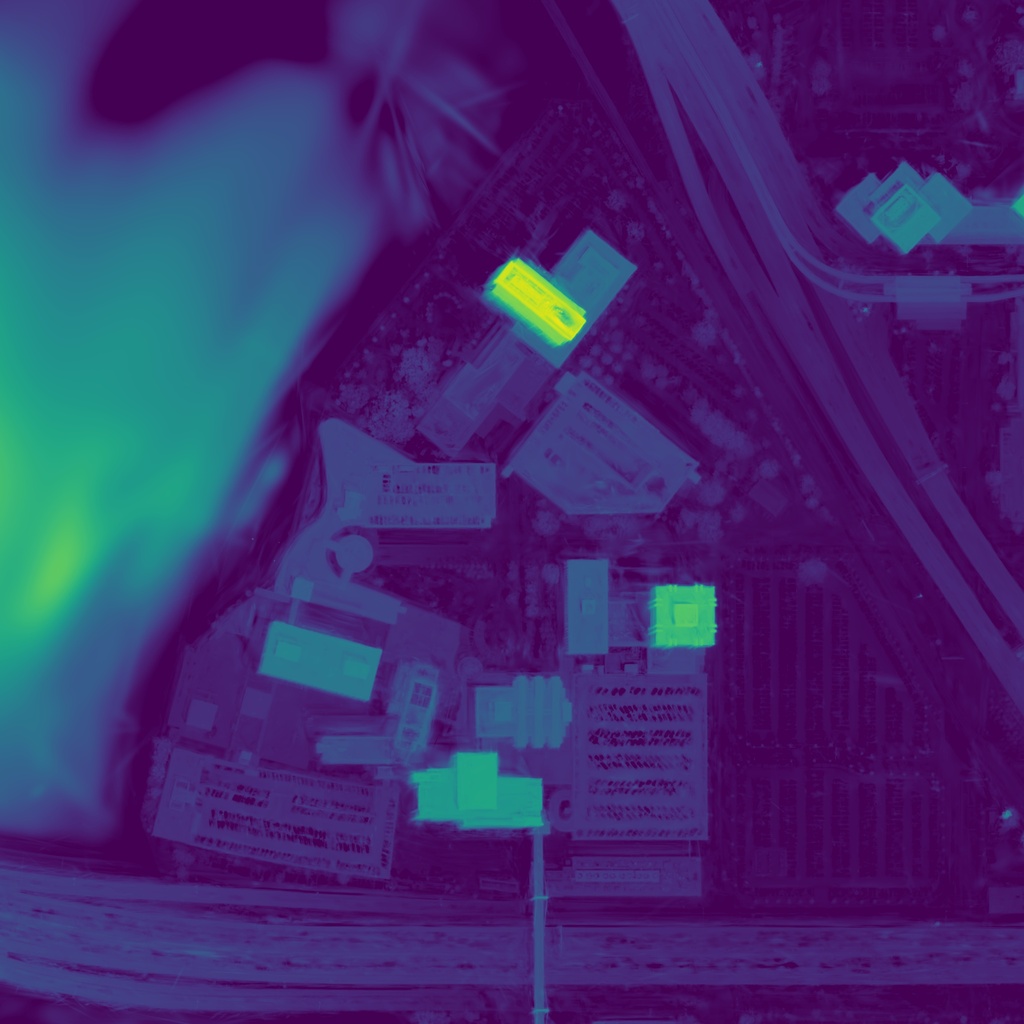}{fig:ours_depth}{\detailCA}
		& \qpanel{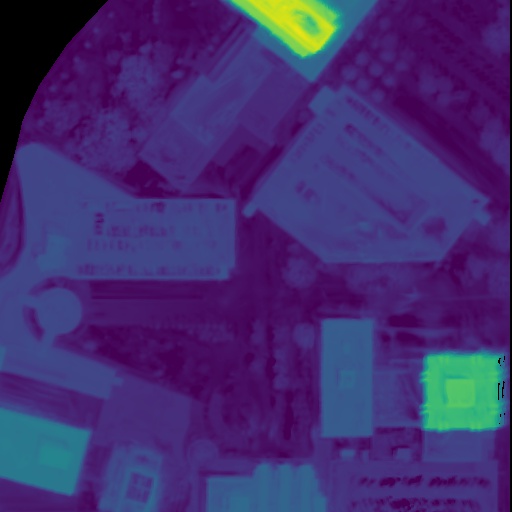}{fig:ours_dsm}{\detailDSMCA}
	\end{tabular}
	
	\caption{
		Qualitative comparison of scene JAX\_214 from the DFC2019 dataset~\cite{dfc2019}. 
		Depth is converted to altitude for cross-camera comparison; the evaluation DSM marks the LiDAR-covered region used for altitude error computation.
		Areas of interest are highlighted in yellow. Colors encode altitude, ranging from blue (low) to yellow (high).
	}
	\label{fig:app_214}
\end{figure*}

\newcommand{\detailDA}{%
	\detailbox{70}{55}{24}{30}%
}
\newcommand{\detailDB}{%
}
\newcommand{\detailDSMDA}{%
	\detailbox{27}{35}{19}{28}%
}

\begin{figure*}[ht]
	\centering
	\setlength{\tabcolsep}{4pt}
	\renewcommand{\arraystretch}{0.9}
	
	\begin{tabular}{
			@{}
			>{\centering\arraybackslash}m{1.1em}
			@{\hspace{2pt}}
			>{\centering\arraybackslash}m{\qualimg}
			@{\hspace{4pt}}
			>{\centering\arraybackslash}m{\qualimg}
			@{\hspace{4pt}}
			>{\centering\arraybackslash}m{\qualimg}
			@{}
		}
		& \textbf{RGB} & \textbf{Altitude} & \textbf{Evaluation DSM} \\[0.1em]
		
		\rowlabel{GT}
		& \qpanel{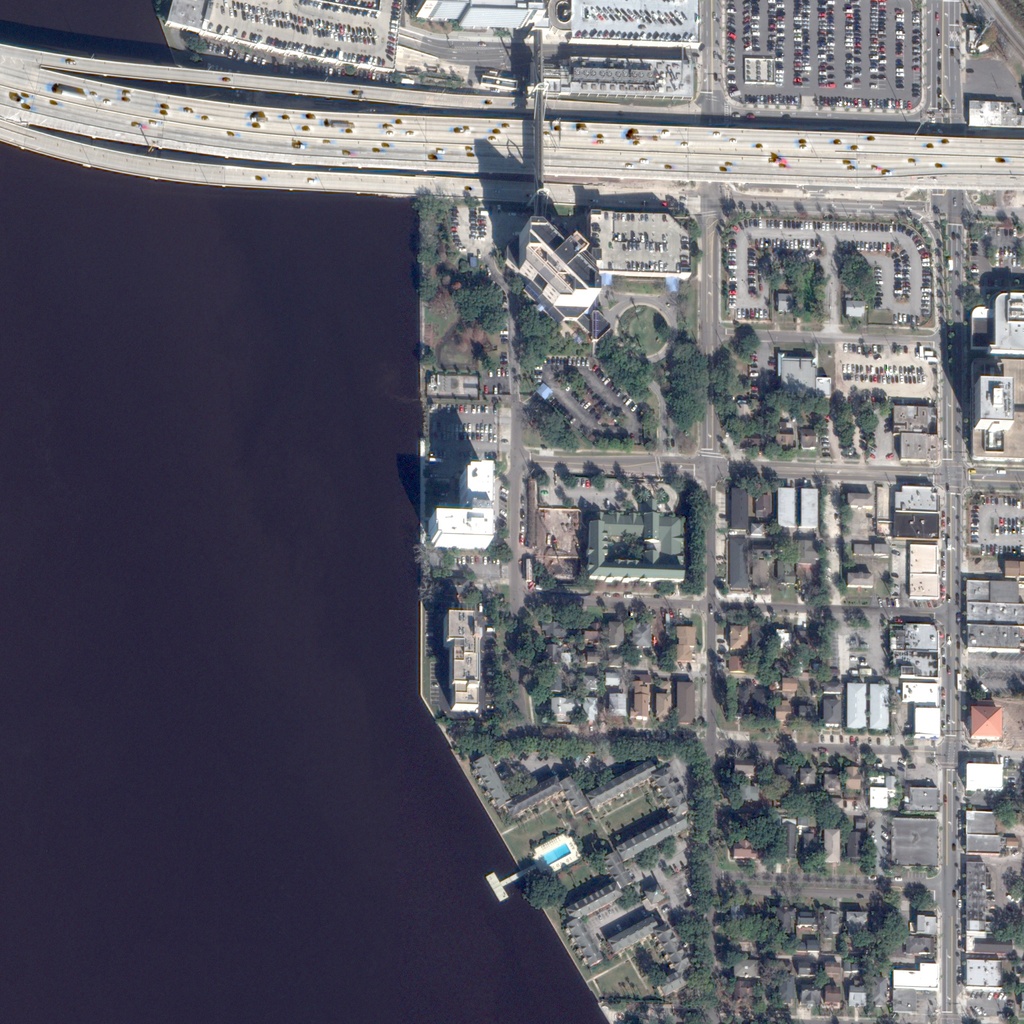}{fig:gt_rgb}{}
		& \qmissing{No view-specific GT available}{}
		& \qpanel{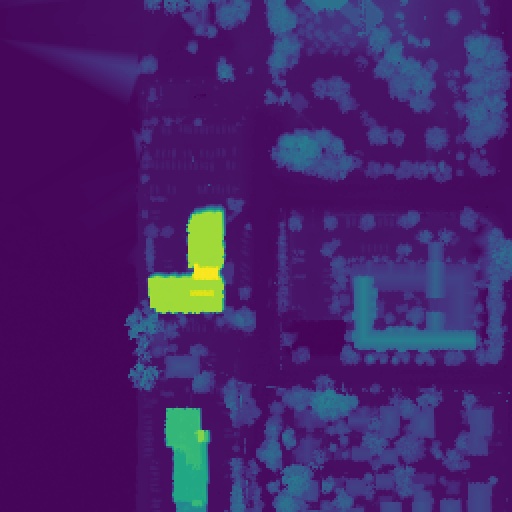}{fig:gt_dsm}{\detailDSMDA}
		\\[0.2em]

		\rowlabel{Affine Approx.}
		& \qpanel{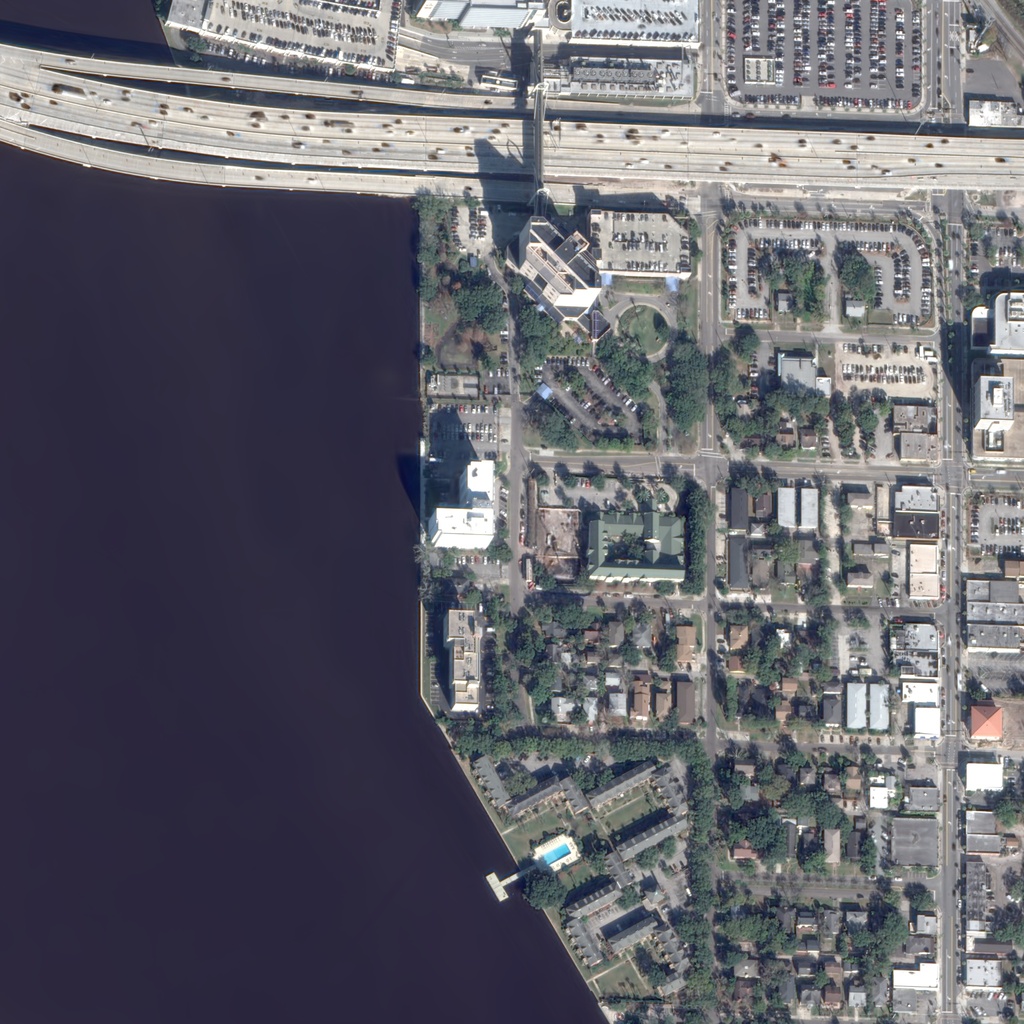}{fig:method_a_rgb}{}
		& \qpanel{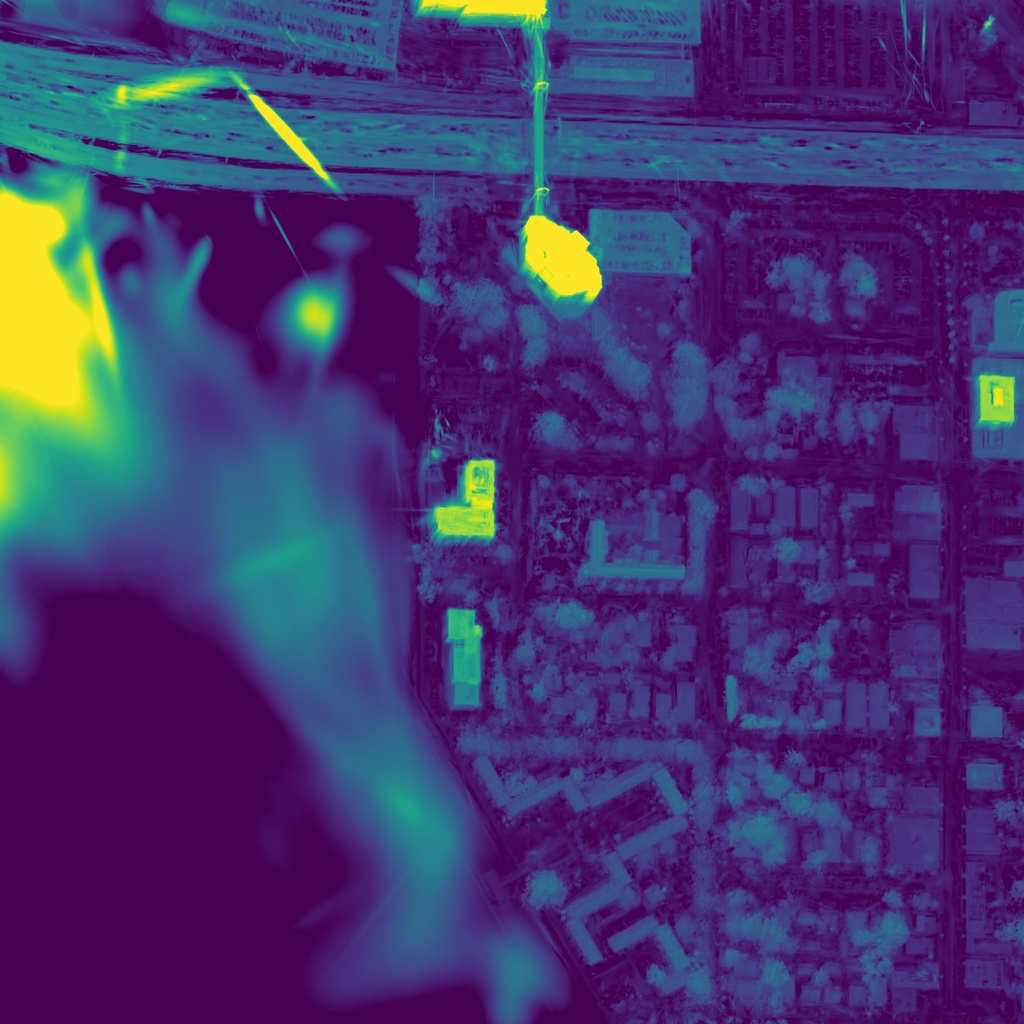}{fig:method_a_depth}{\detailDA\detailDB}
		& \qpanel{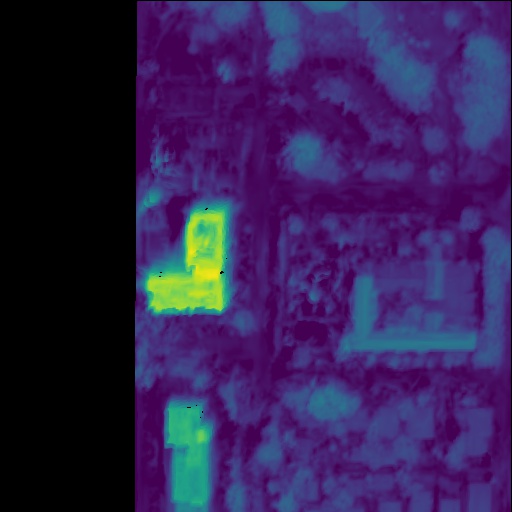}{fig:method_a_dsm}{\detailDSMDA}
		\\[0.2em]
		
		\rowlabel{Perspective Approx.}
		& \qpanel{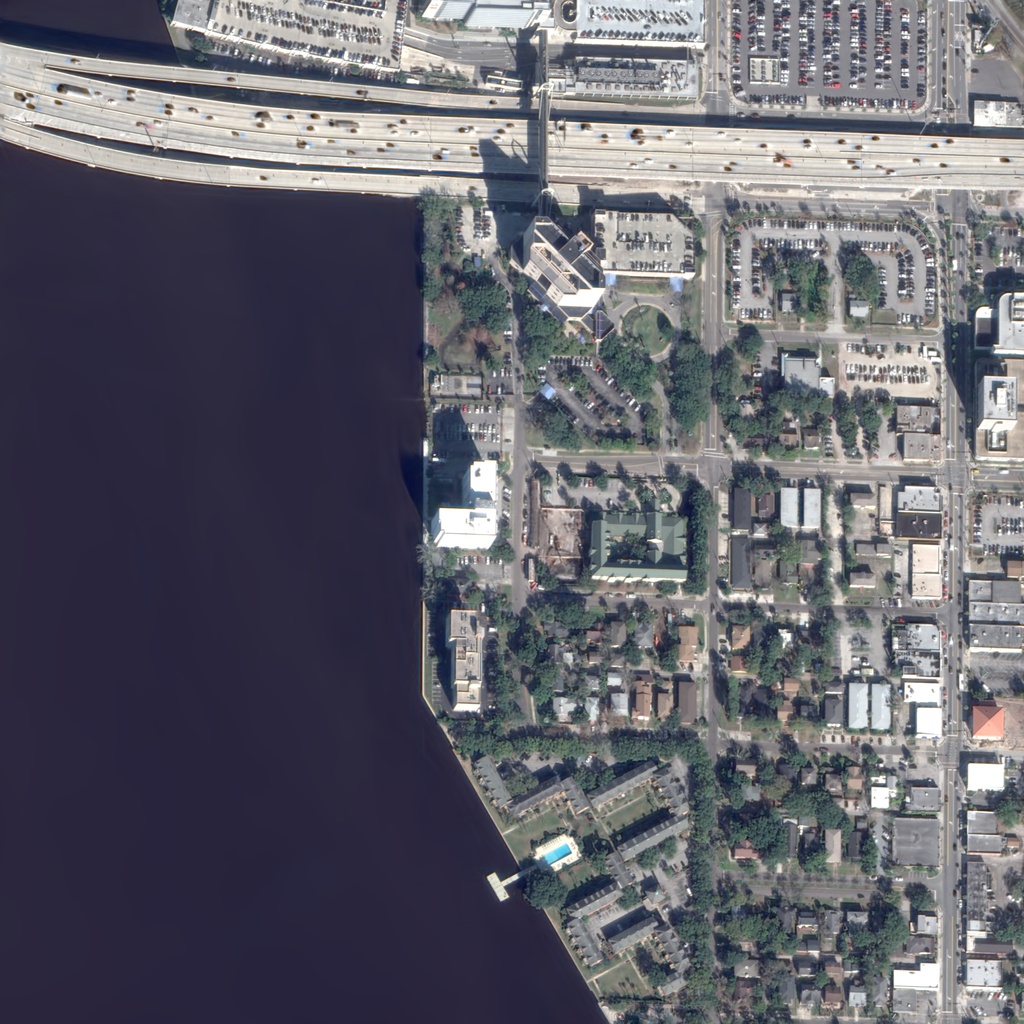}{fig:method_a_rgb}{}
		& \qpanel{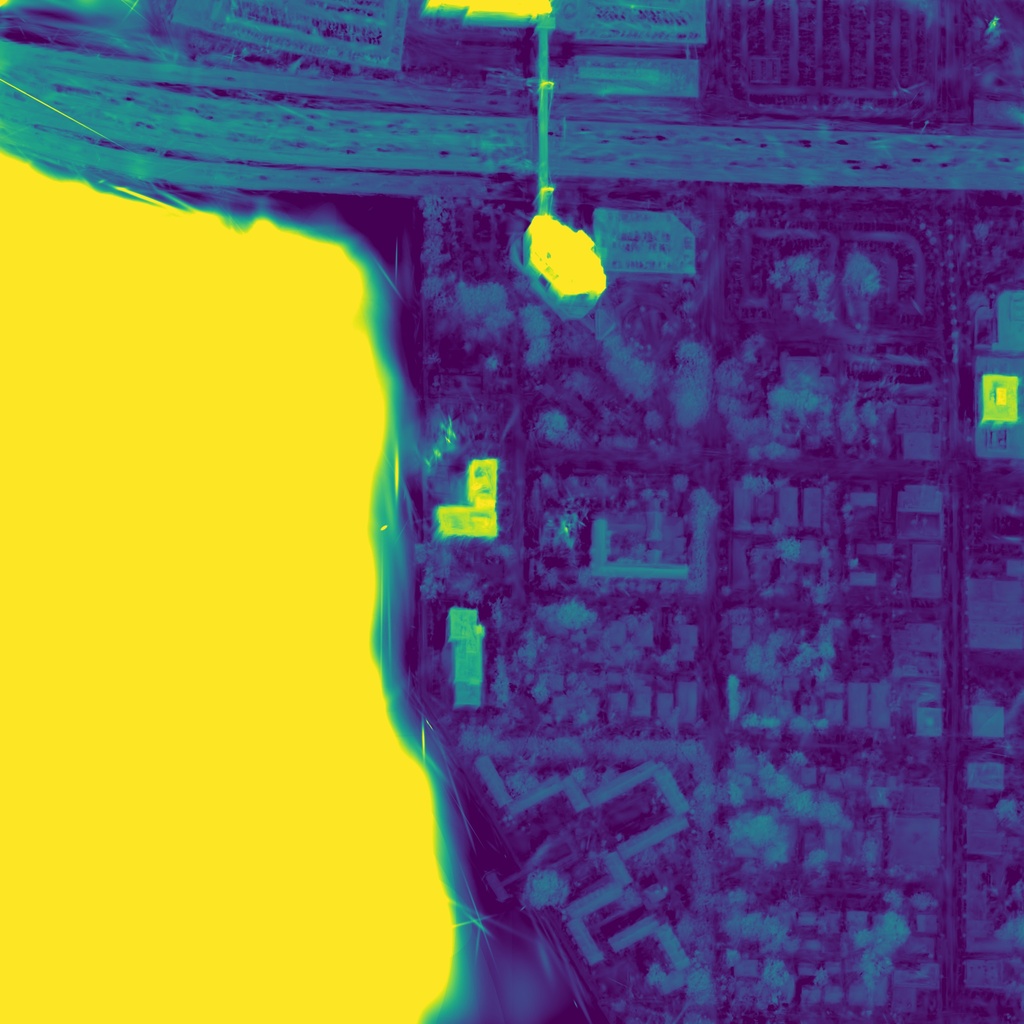}{fig:method_a_depth}{\detailDA\detailDB}
		& \qpanel{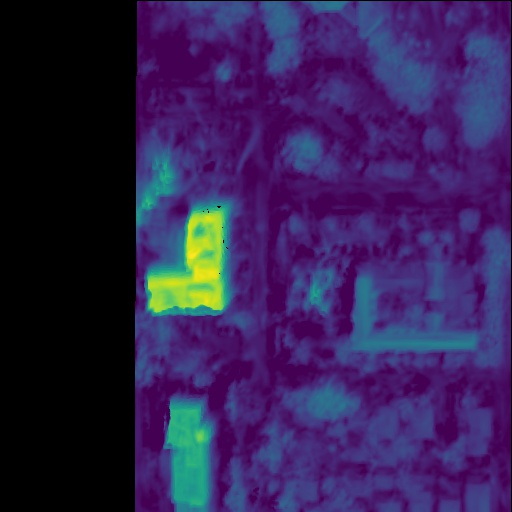}{fig:method_a_dsm}{\detailDSMDA}
		\\[0.2em]
		
		\rowlabel{Native RPC (Ours)}
		& \qpanel{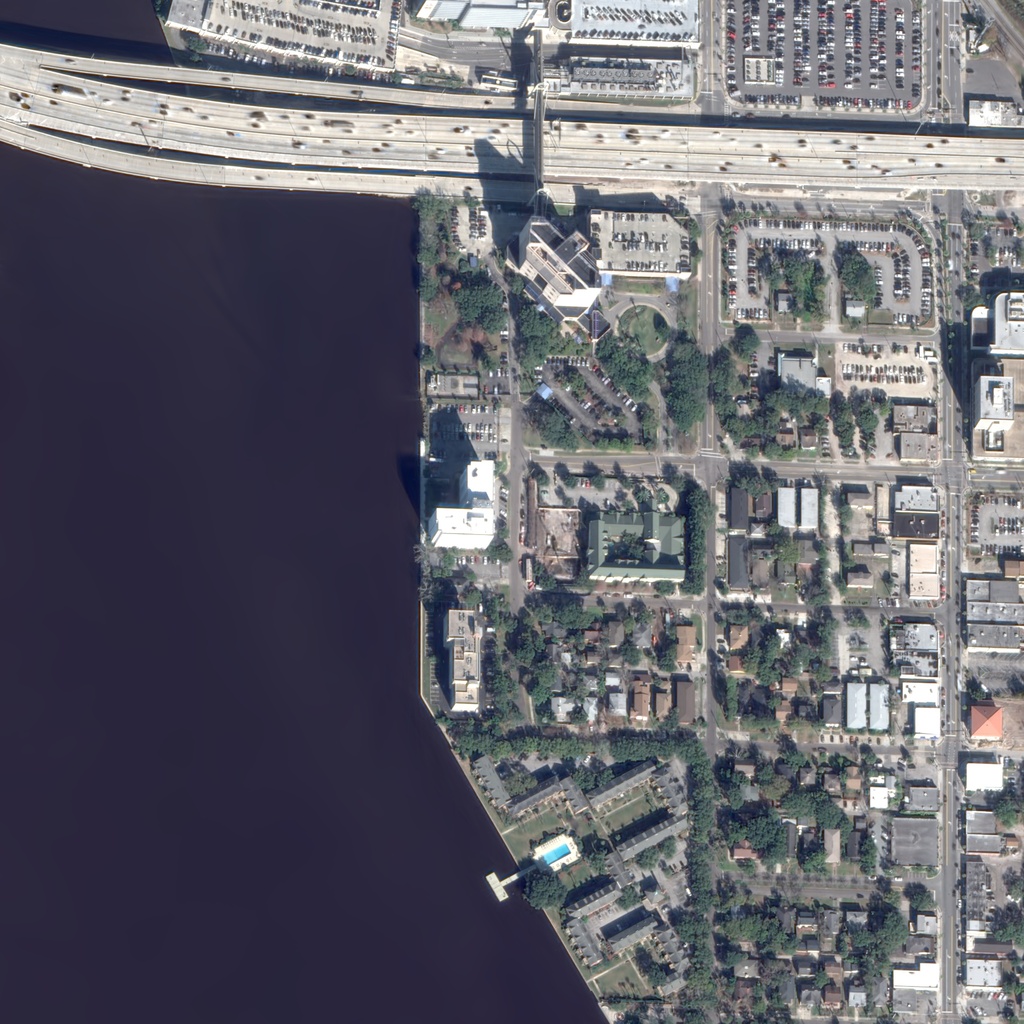}{fig:ours_rgb}{}
		& \qpanel{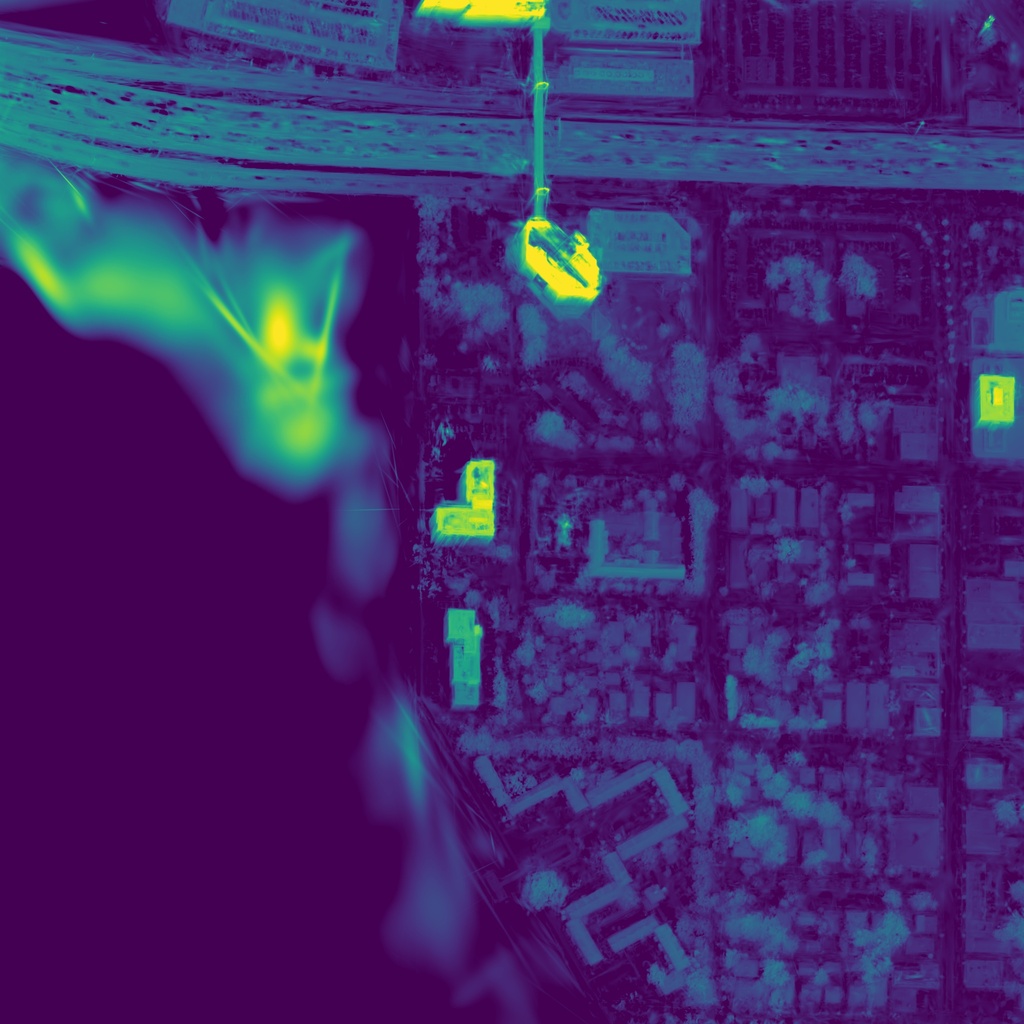}{fig:ours_depth}{\detailDA\detailDB}
		& \qpanel{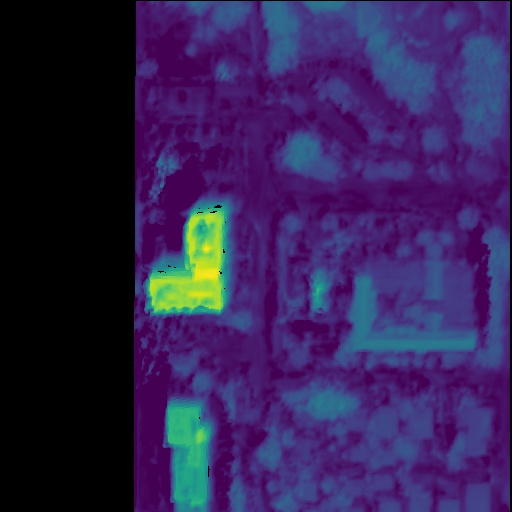}{fig:ours_dsm}{\detailDSMDA}
	\end{tabular}
	
	\caption{
		Qualitative comparison for scene JAX\_260 from the DFC2019 dataset~\cite{dfc2019}. 
		Depth is converted to altitude for cross-camera comparison; the evaluation DSM marks the LiDAR-covered region used for altitude error computation.
		Areas of interest are highlighted in yellow. Colors encode altitude, ranging from blue (low) to yellow (high).
	}
	\label{fig:app_260}
\end{figure*}

\newlength{\qcropw}
\newlength{\qcroprowgap}
\setlength{\qcropw}{0.32\textwidth}
\setlength{\qcroprowgap}{1.0em}

\newsavebox{\qcropbox}

\newcommand{\qcentercropsquare}[2]{%
	\sbox{\qcropbox}{\includegraphics{#2}}%
	\begin{tikzpicture}[inner sep=0pt, outer sep=0pt]
		\path[use as bounding box] (0,0) rectangle (#1,#1);
		\clip (0,0) rectangle (#1,#1);
		
		\ifdim\wd\qcropbox>\ht\qcropbox
		\node[inner sep=0pt, anchor=center]
		at (\dimexpr#1/2\relax,\dimexpr#1/2\relax)
		{\includegraphics[height=#1]{#2}};
		\else
		\node[inner sep=0pt, anchor=center]
		at (\dimexpr#1/2\relax,\dimexpr#1/2\relax)
		{\includegraphics[width=#1]{#2}};
		\fi
	\end{tikzpicture}%
}

\newcommand{\qpanelcrop}[3]{%
	\begin{subfigure}[c]{\qcropw}
		\centering
		\rotatebox{180}{%
			\begingroup
			\setlength{\unitlength}{\dimexpr\qcropw/100\relax}%
			\begin{picture}(100,100)
				\put(0,0){%
					\makebox(100,100)[lb]{%
						\qcentercropsquare{\qcropw}{#1}%
					}%
				}%
				#3%
			\end{picture}%
			\endgroup
		}%
		\phantomcaption\label{#2}%
	\end{subfigure}%
}

\newcommand{\qpanelcropnorotate}[3]{%
	\begin{subfigure}[c]{\qcropw}
		\centering
		\begingroup
		\setlength{\unitlength}{\dimexpr\qcropw/100\relax}%
		\begin{picture}(100,100)
			\put(0,0){%
				\makebox(100,100)[lb]{%
					\qcentercropsquare{\qcropw}{#1}%
				}%
			}%
			#3%
		\end{picture}%
		\endgroup
		\phantomcaption\label{#2}%
	\end{subfigure}%
}

\newcommand{\qmissingcrop}[1]{%
	\begin{minipage}[c][\qcropw][c]{\qcropw}
		\centering
		\fbox{%
			\begin{minipage}[c][0.92\qcropw][c]{0.92\qcropw}
				\centering
				\scriptsize #1
			\end{minipage}%
		}
	\end{minipage}%
}

\newcommand{\qrowlabelcrop}[1]{%
	\begin{minipage}[c][\qcropw][c]{1.1em}
		\centering
		\rotatebox[origin=c]{90}{\scriptsize\bfseries #1}%
	\end{minipage}%
}

\newcommand{\detailEA}{%
}
\newcommand{\detailDSMEA}{%
	\detailbox{6}{42}{25}{25}%
}

\begin{figure*}[ht]
	\centering
	\setlength{\tabcolsep}{3pt}
	\renewcommand{\arraystretch}{1.0}

	\begin{tabular}{
			@{}
			>{\centering\arraybackslash}m{1.1em}
			@{\hspace{3pt}}
			>{\centering\arraybackslash}m{\qcropw}
			@{\hspace{5pt}}
			>{\centering\arraybackslash}m{\qcropw}
			@{\hspace{5pt}}
			>{\centering\arraybackslash}m{\qcropw}
			@{}
		}
		& \textbf{RGB} & \textbf{Altitude} & \textbf{Evaluation DSM}
		\\
		\noalign{\vskip.4em}
		
		\qrowlabelcrop{GT}
		& \qpanelcrop{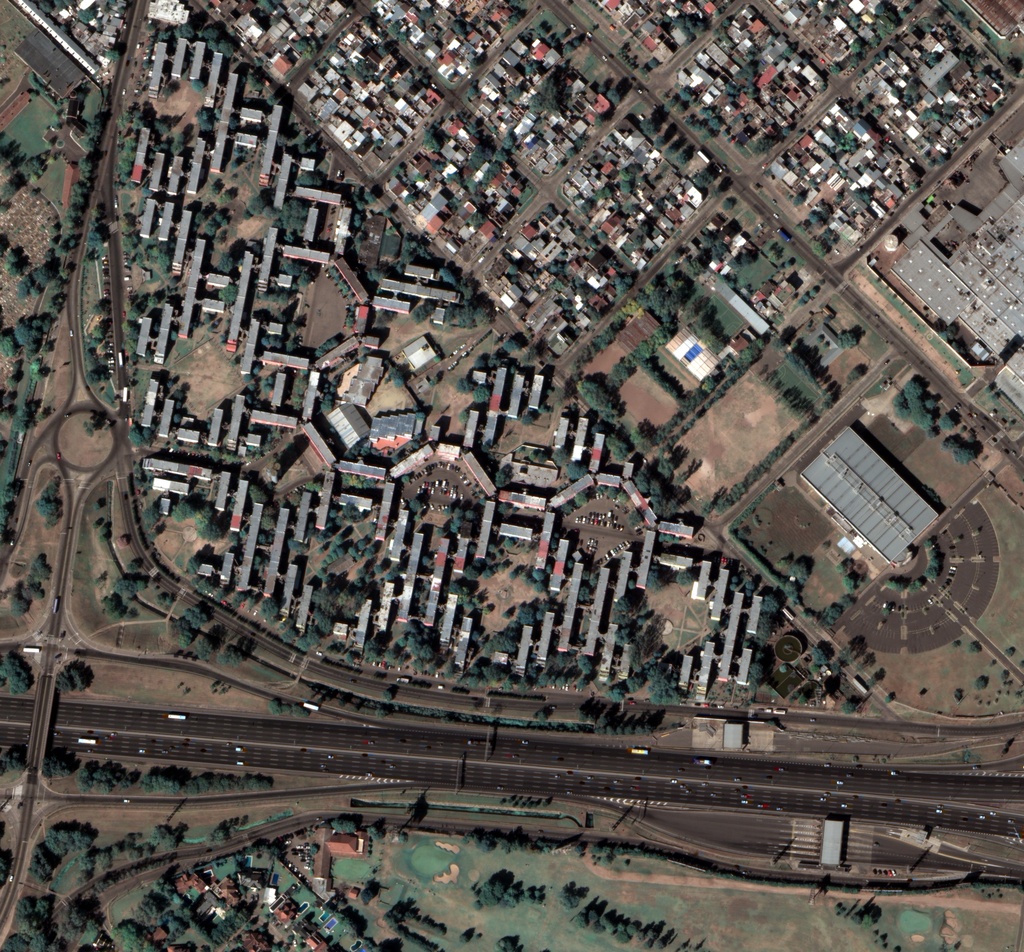}{fig:iarpa_gt_rgb}{}
		& \qmissingcrop{No view-specific GT available}
		& \qpanelcropnorotate{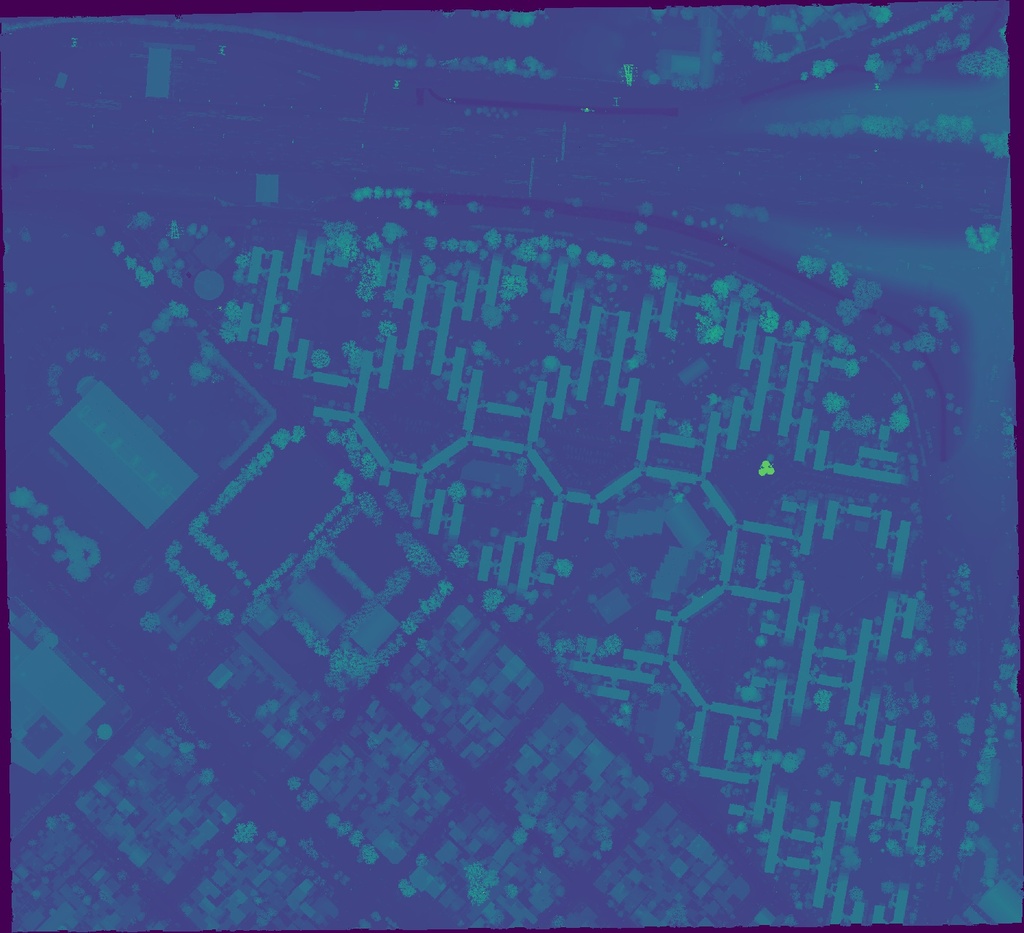}{fig:iarpa_gt_dsm}{\detailDSMEA}
		\\
		\noalign{\vskip\qcroprowgap}
		
		\qrowlabelcrop{Affine Approx.}
		& \qpanelcrop{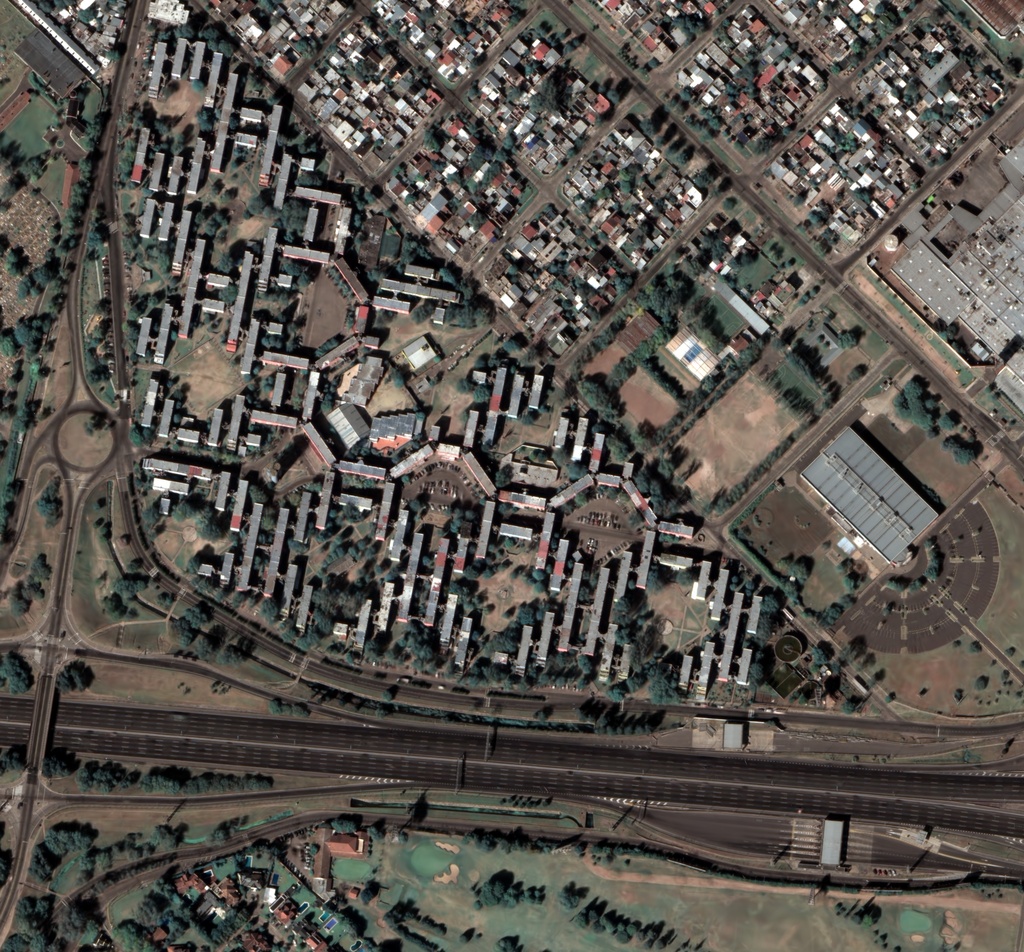}{fig:iarpa_affine_rgb}{}
		& \qpanelcrop{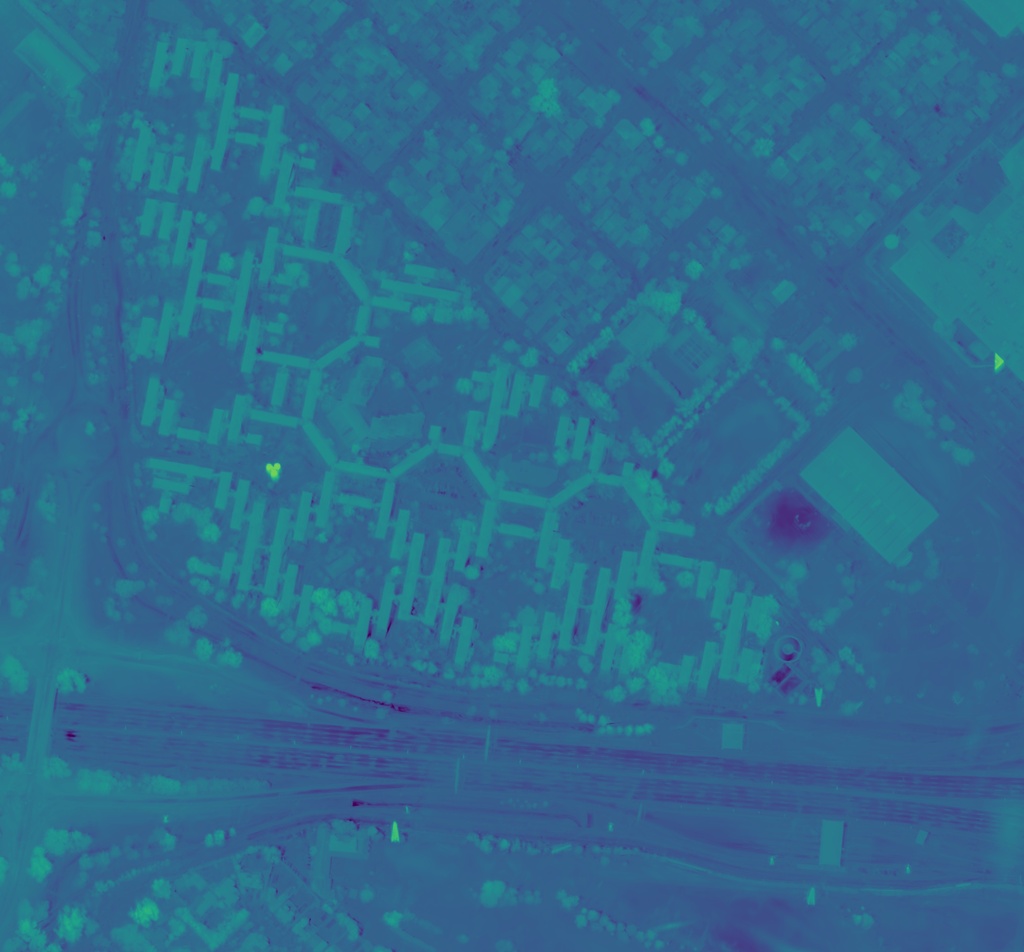}{fig:iarpa_affine_depth}{\detailEA}
		& \qpanelcropnorotate{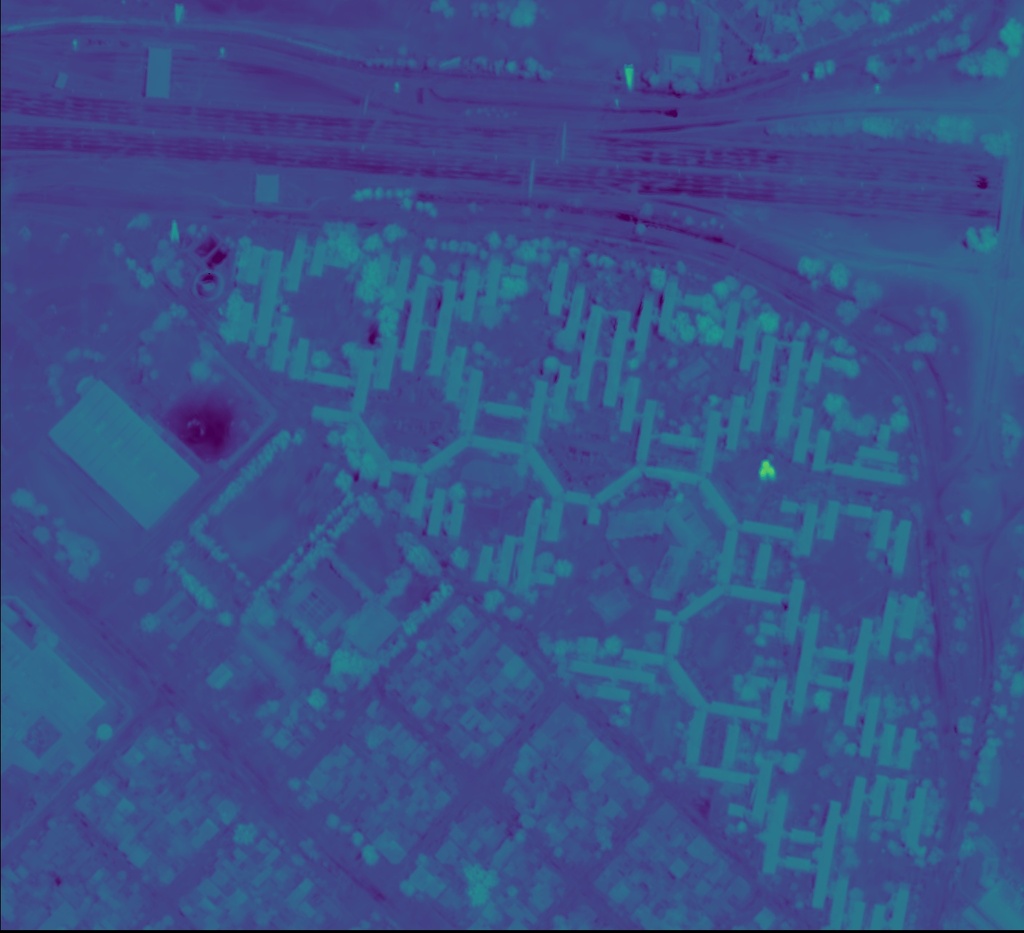}{fig:iarpa_affine_dsm}{\detailDSMEA}
		\\
		\noalign{\vskip\qcroprowgap}
		
		\qrowlabelcrop{Perspective Approx.}
		& \qpanelcrop{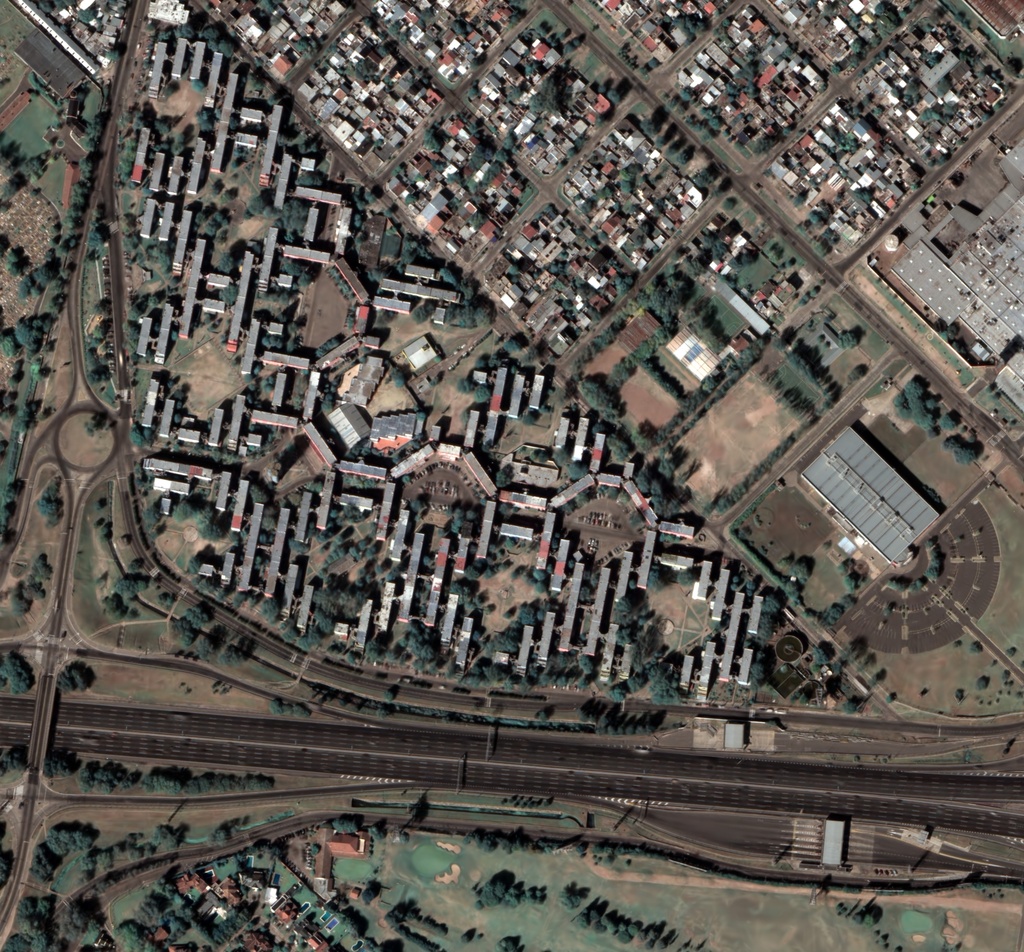}{fig:iarpa_perspective_rgb}{}
		& \qpanelcrop{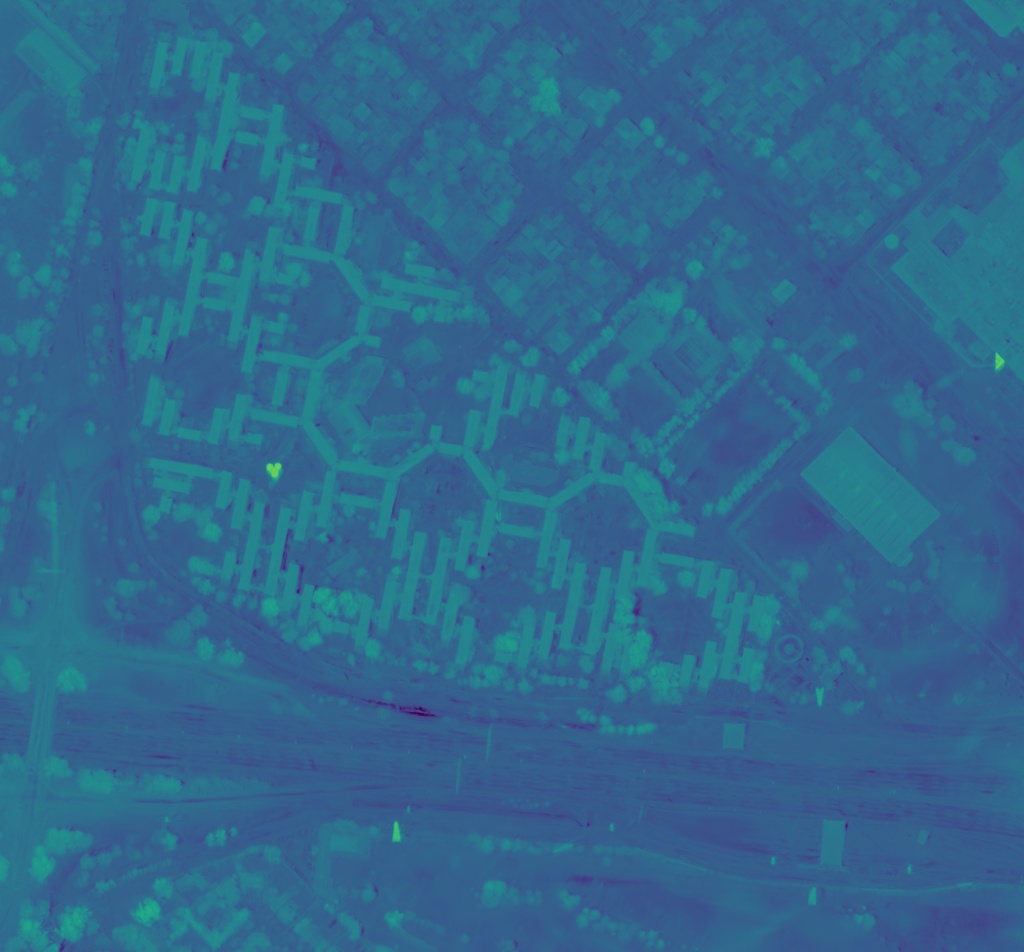}{fig:iarpa_perspective_depth}{\detailEA}
		& \qpanelcropnorotate{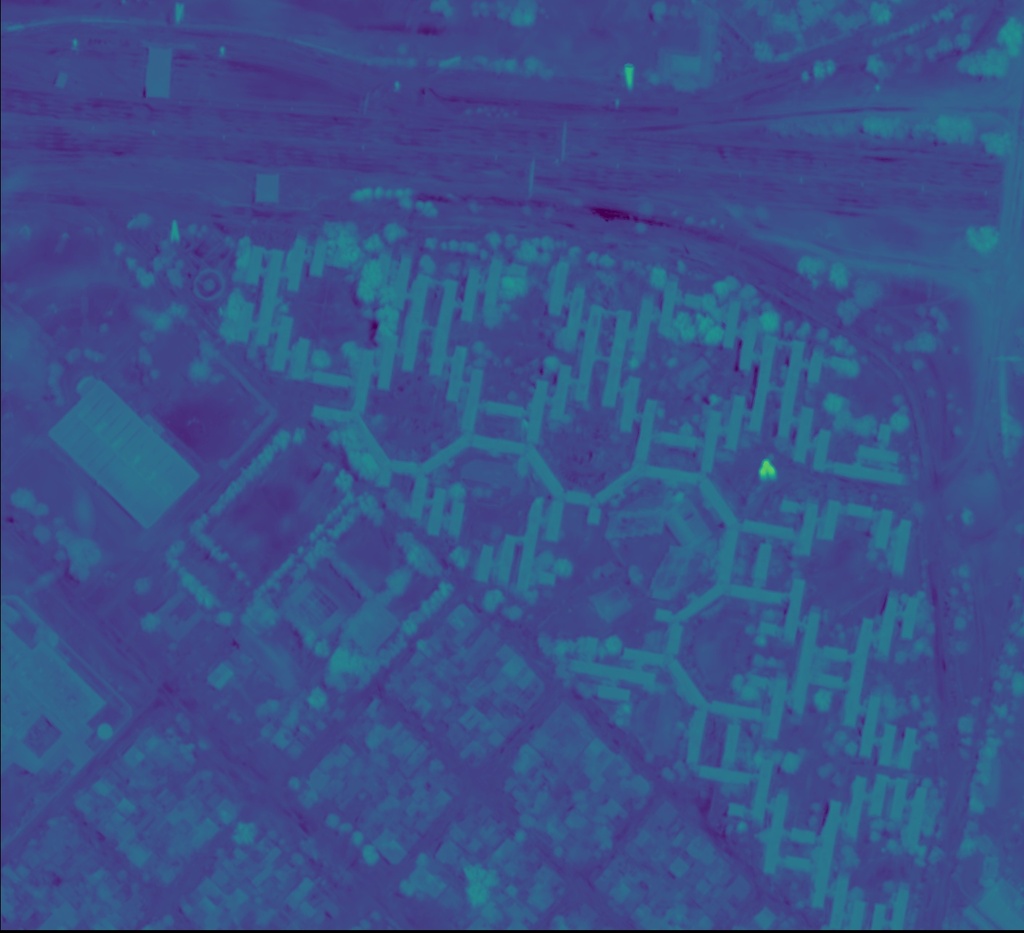}{fig:iarpa_perspective_dsm}{\detailDSMEA}
		\\
		\noalign{\vskip\qcroprowgap}
		
		\qrowlabelcrop{Native RPC (Ours)}
		& \qpanelcrop{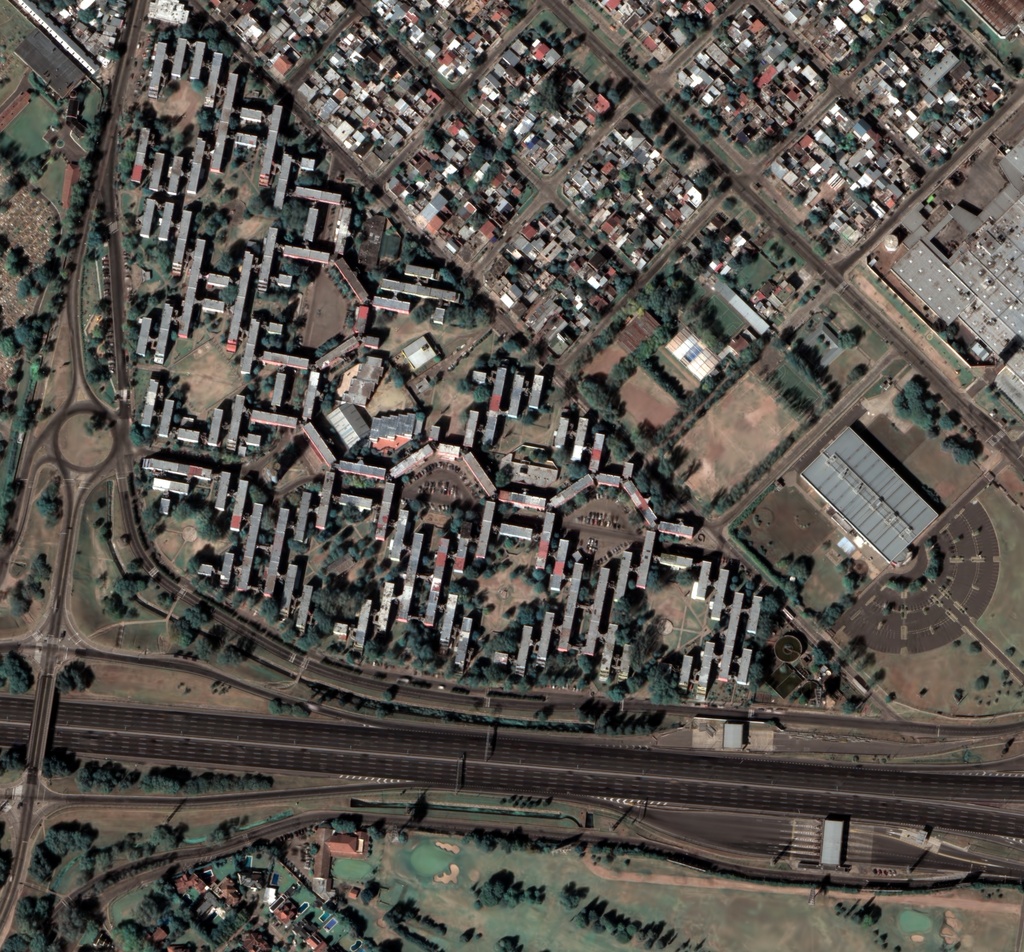}{fig:iarpa_ours_rgb}{}
		& \qpanelcrop{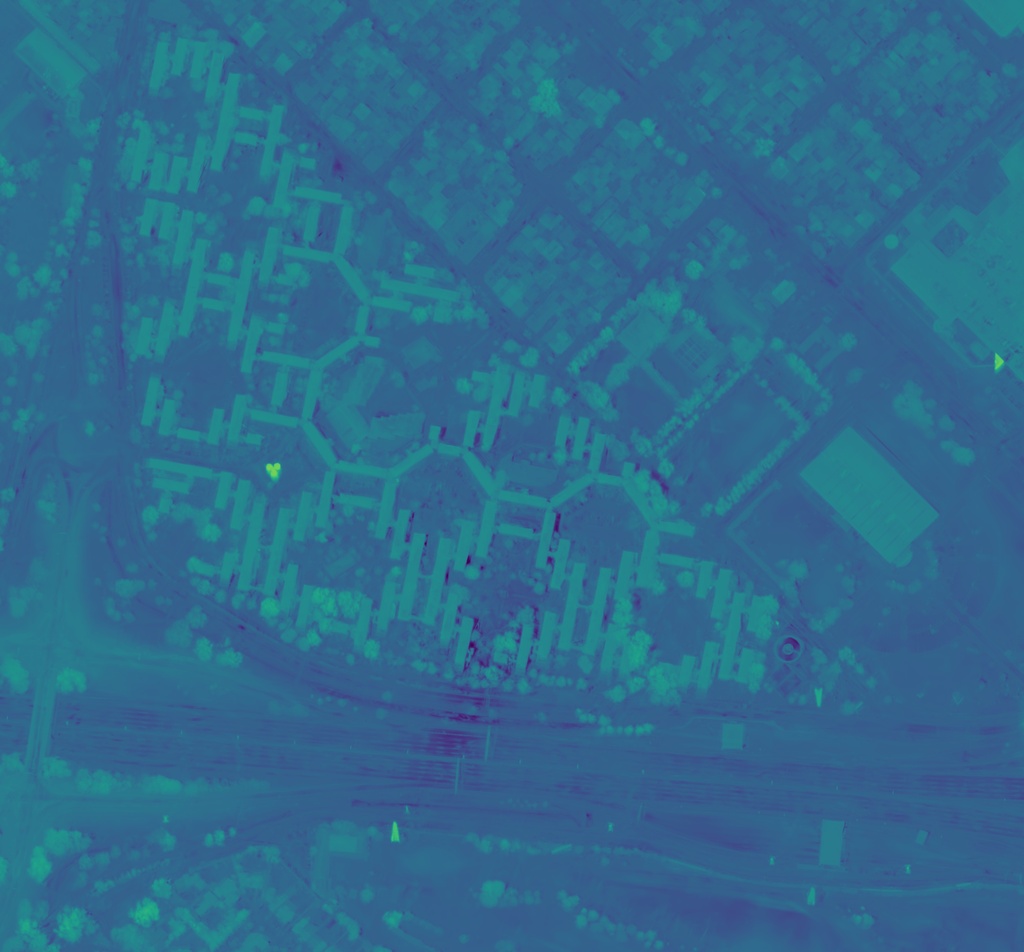}{fig:iarpa_ours_depth}{\detailEA}
		& \qpanelcropnorotate{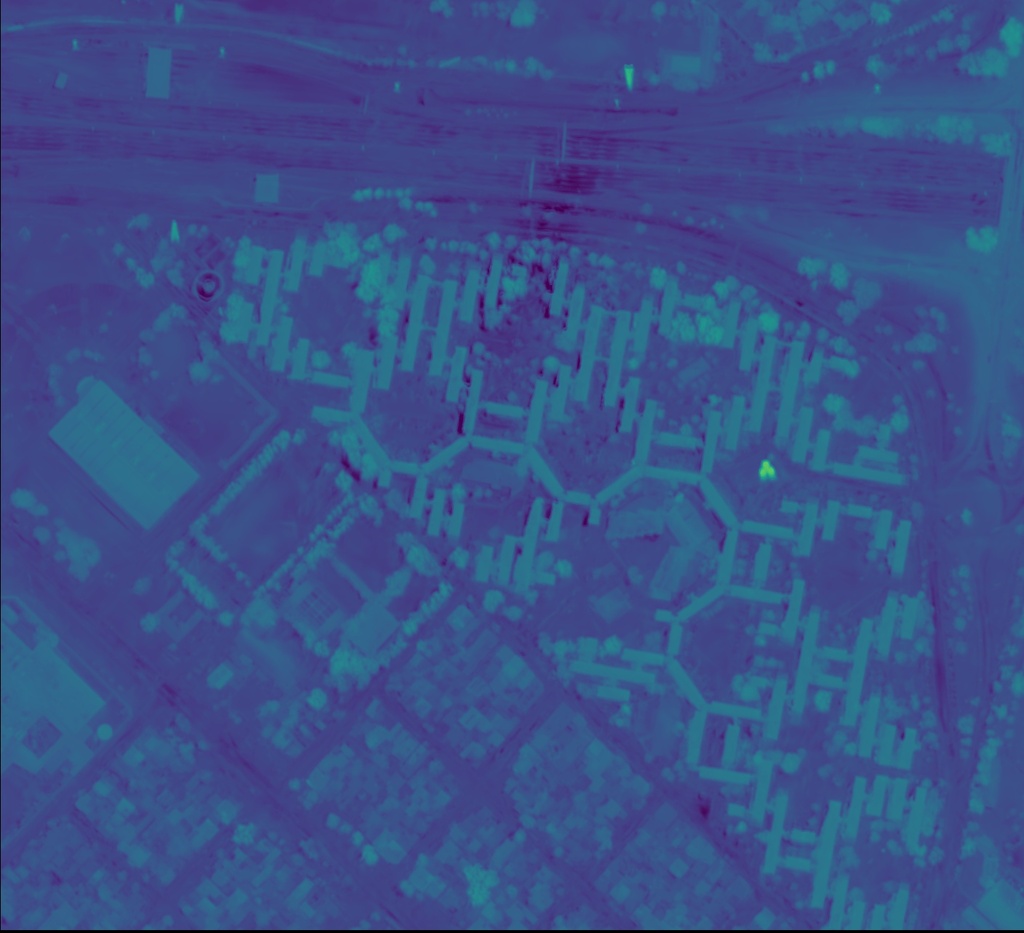}{fig:iarpa_ours_dsm}{\detailDSMEA}
	\end{tabular}
	
	\caption{
		Qualitative comparison for scene 001 from the IARPA2016 dataset~\cite{IARPA2016}. 
		Depth is converted to altitude for cross-camera comparison; the evaluation DSM marks the LiDAR-covered region used for altitude error computation.
		Areas of interest are highlighted in yellow. Colors encode altitude, ranging from blue (low) to yellow (high).
	}
	\label{fig:app_site1}
\end{figure*}	

\newcommand{\detailFA}{%
}
\newcommand{\detailDSMFA}{%
	\detailbox{28}{7}{45}{40}%
}

\begin{figure*}[ht]
	\centering
	\setlength{\tabcolsep}{3pt}
	\renewcommand{\arraystretch}{1.0}

	\begin{tabular}{
			@{}
			>{\centering\arraybackslash}m{1.1em}
			@{\hspace{3pt}}
			>{\centering\arraybackslash}m{\qcropw}
			@{\hspace{5pt}}
			>{\centering\arraybackslash}m{\qcropw}
			@{\hspace{5pt}}
			>{\centering\arraybackslash}m{\qcropw}
			@{}
		}
		& \textbf{RGB} & \textbf{Altitude} & \textbf{Evaluation DSM}
		\\
		\noalign{\vskip.4em}
		
		\qrowlabelcrop{GT}
		& \qpanelcrop{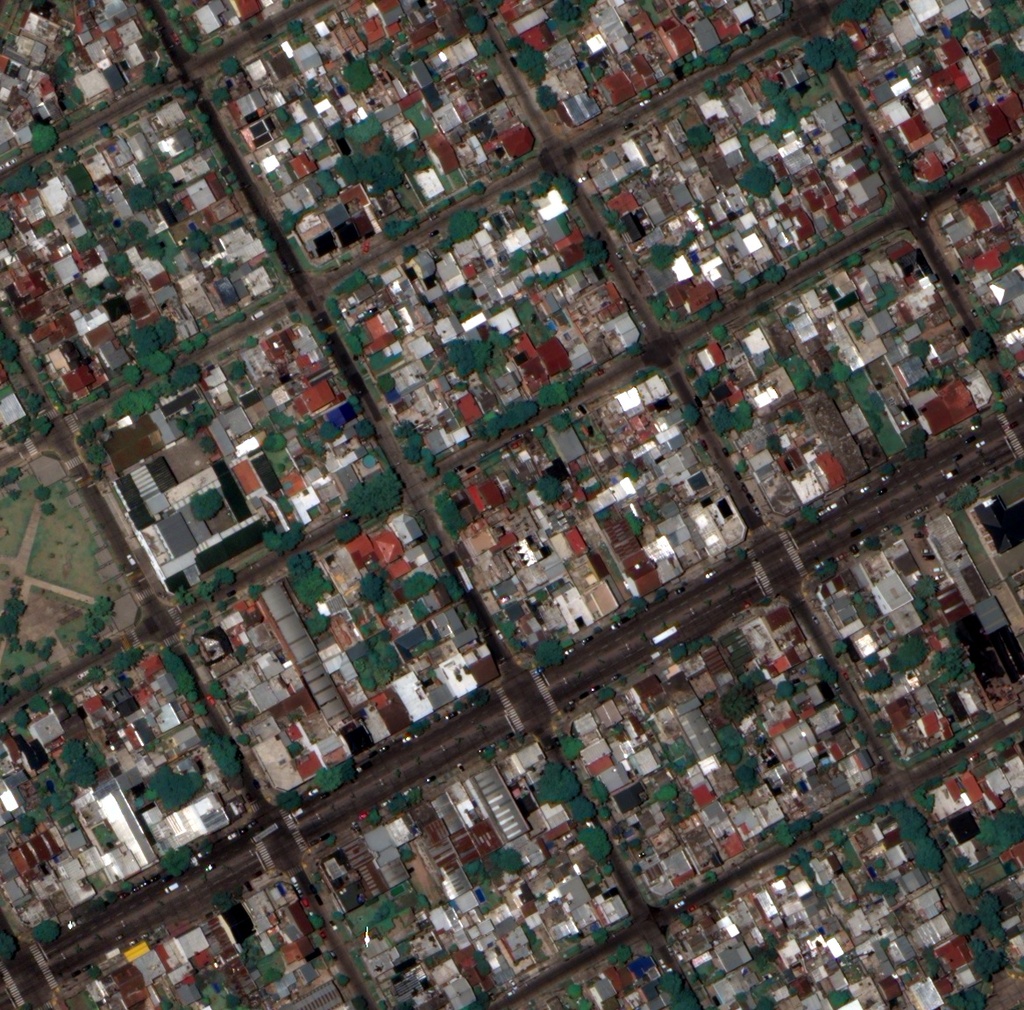}{fig:iarpa_gt_rgb}{}
		& \qmissingcrop{No view-specific GT available}
		& \qpanelcropnorotate{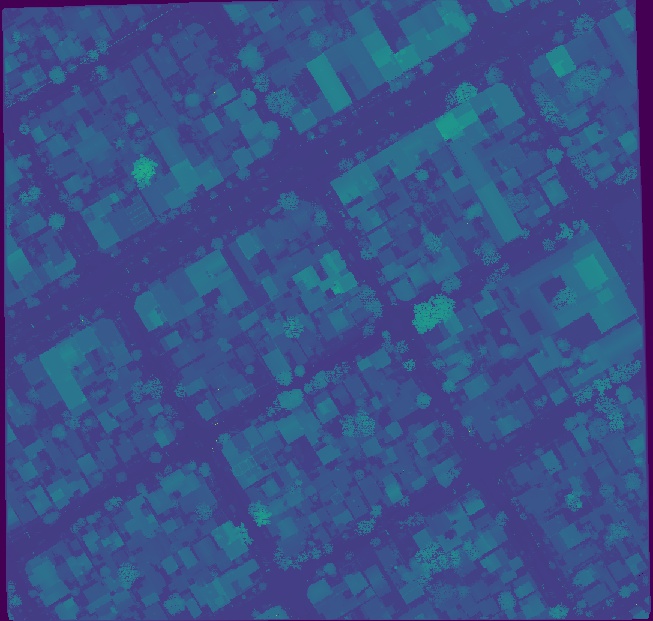}{fig:iarpa_gt_dsm}{\detailDSMFA}
		\\
		\noalign{\vskip\qcroprowgap}
		
		\qrowlabelcrop{Affine Approx.}
		& \qpanelcrop{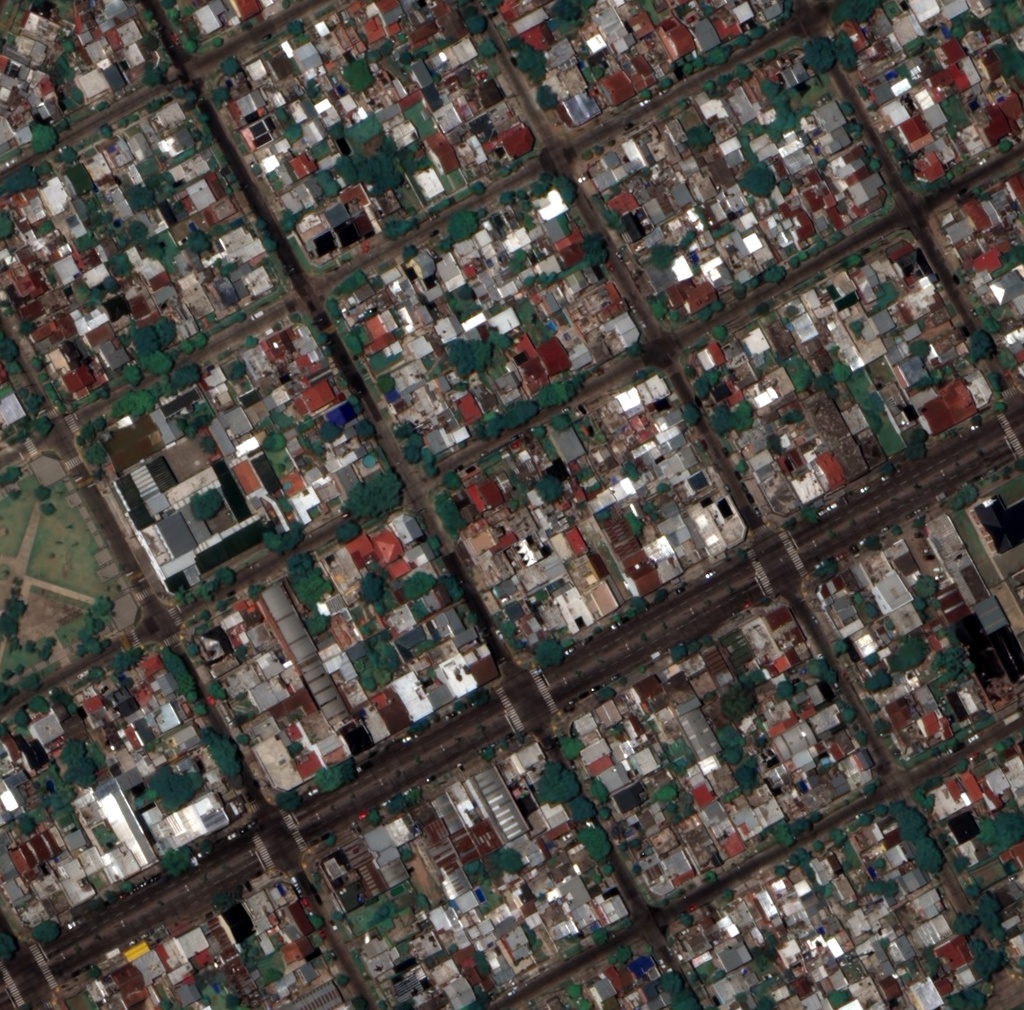}{fig:iarpa_affine_rgb}{}
		& \qpanelcrop{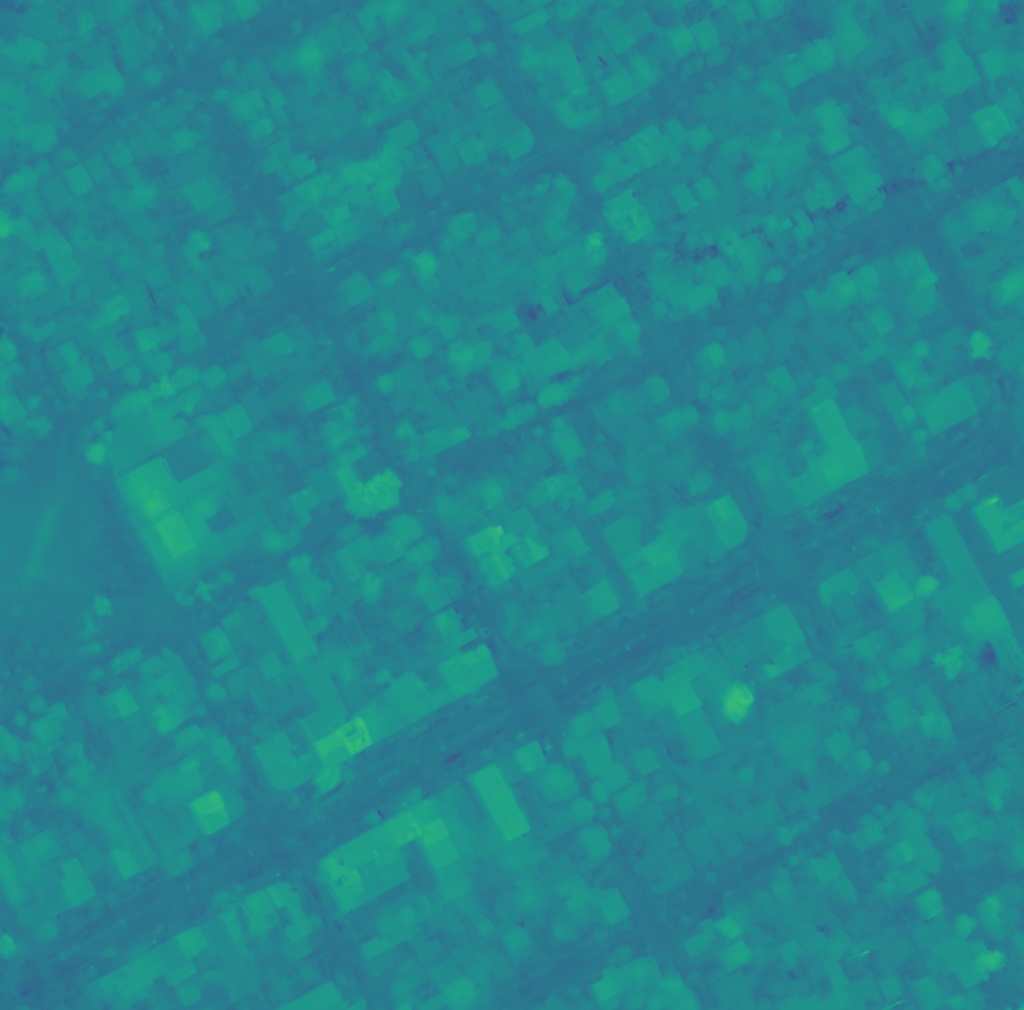}{fig:iarpa_affine_depth}{\detailFA}
		& \qpanelcropnorotate{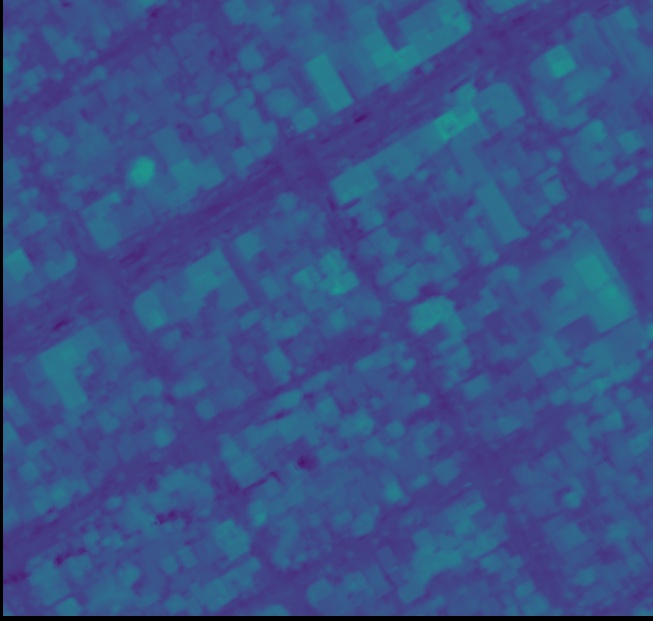}{fig:iarpa_affine_dsm}{\detailDSMFA}
		\\
		\noalign{\vskip\qcroprowgap}
		
		\qrowlabelcrop{Perspective Approx.}
		& \qpanelcrop{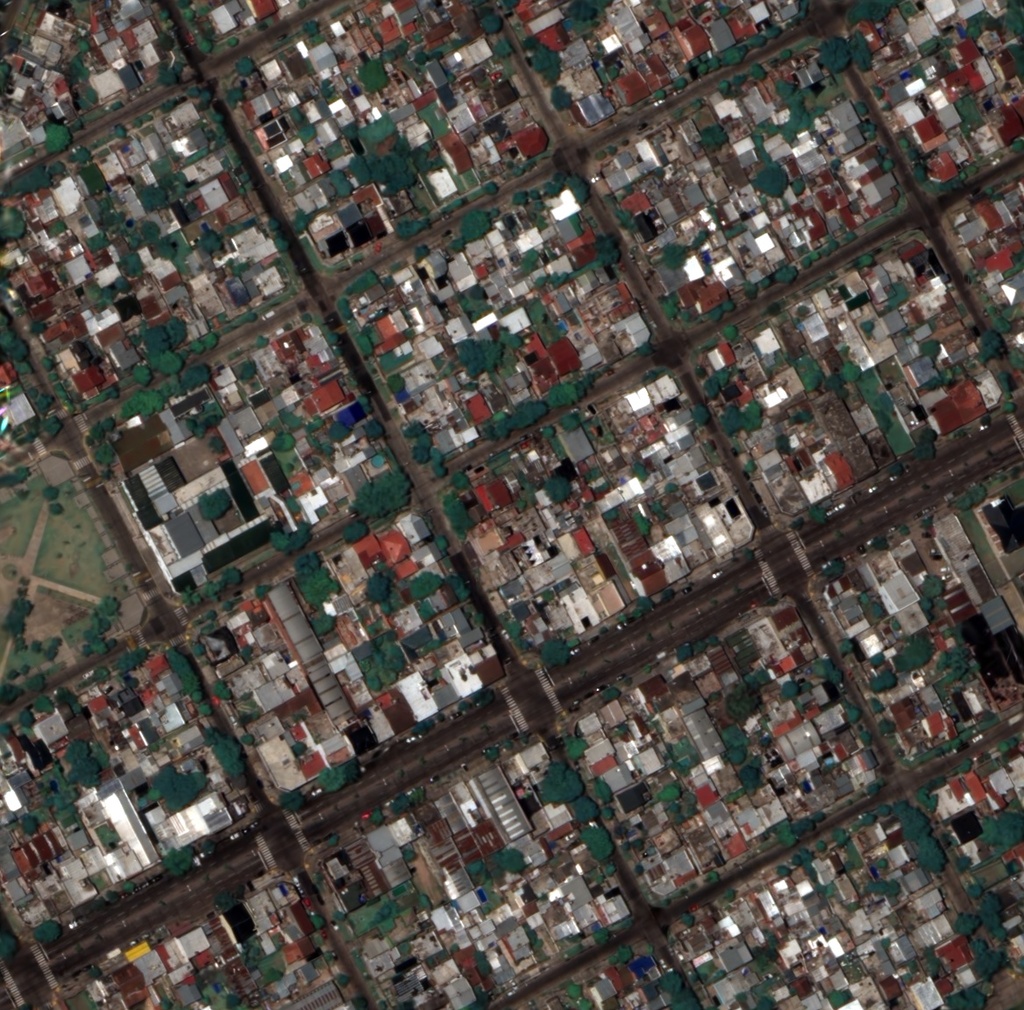}{fig:iarpa_perspective_rgb}{}
		& \qpanelcrop{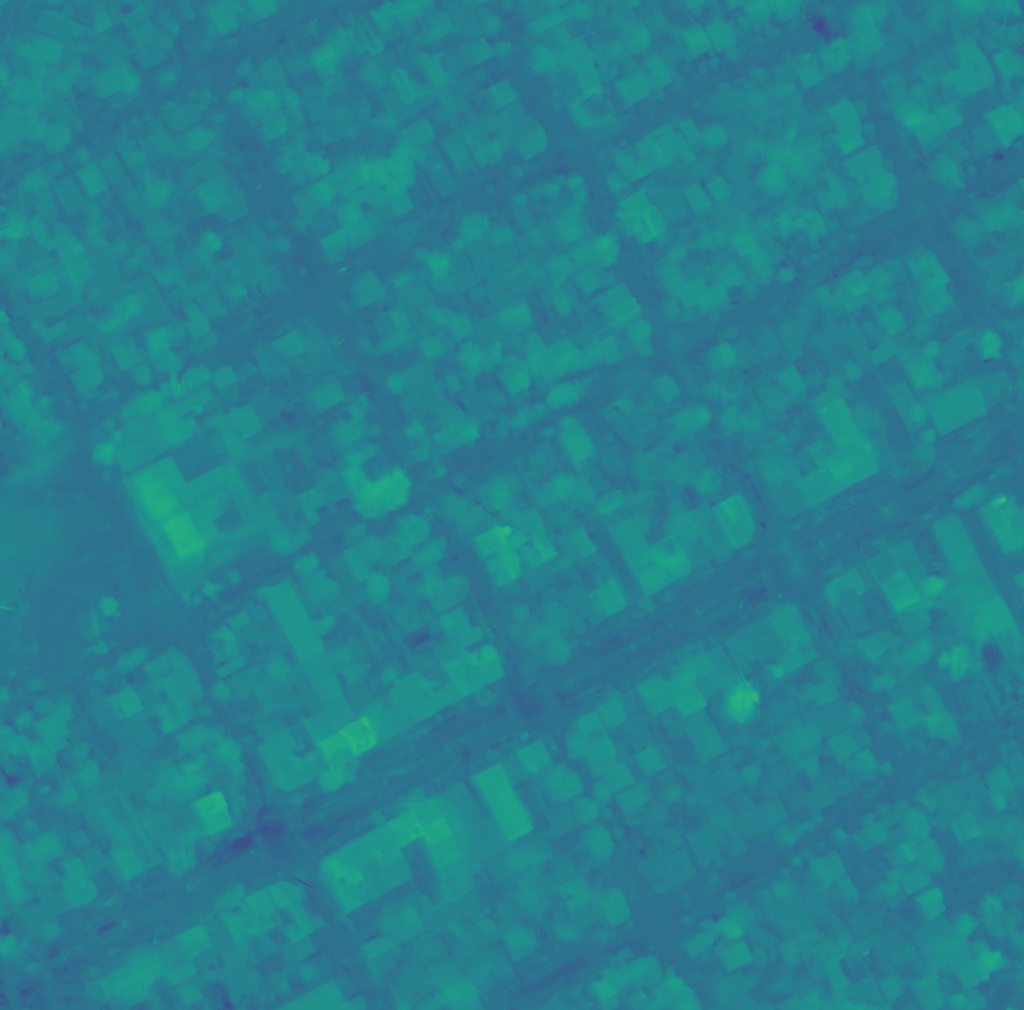}{fig:iarpa_perspective_depth}{\detailFA}
		& \qpanelcropnorotate{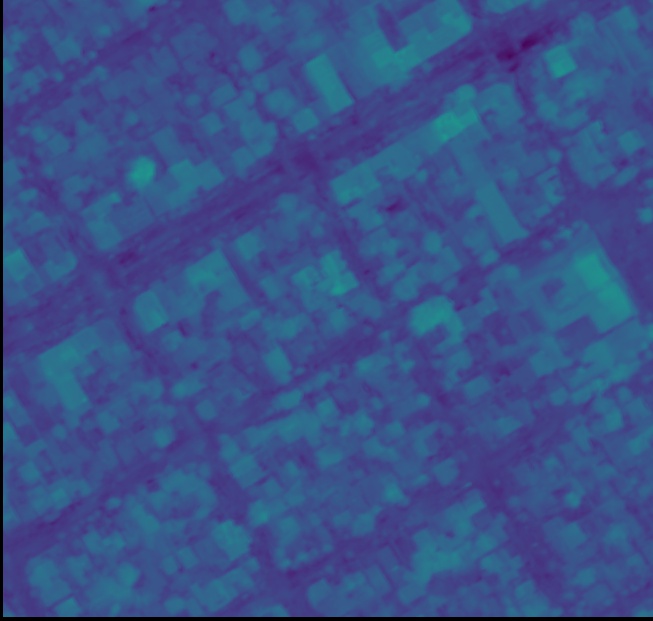}{fig:iarpa_perspective_dsm}{\detailDSMFA}
		\\
		\noalign{\vskip\qcroprowgap}
		
		\qrowlabelcrop{Native RPC (Ours)}
		& \qpanelcrop{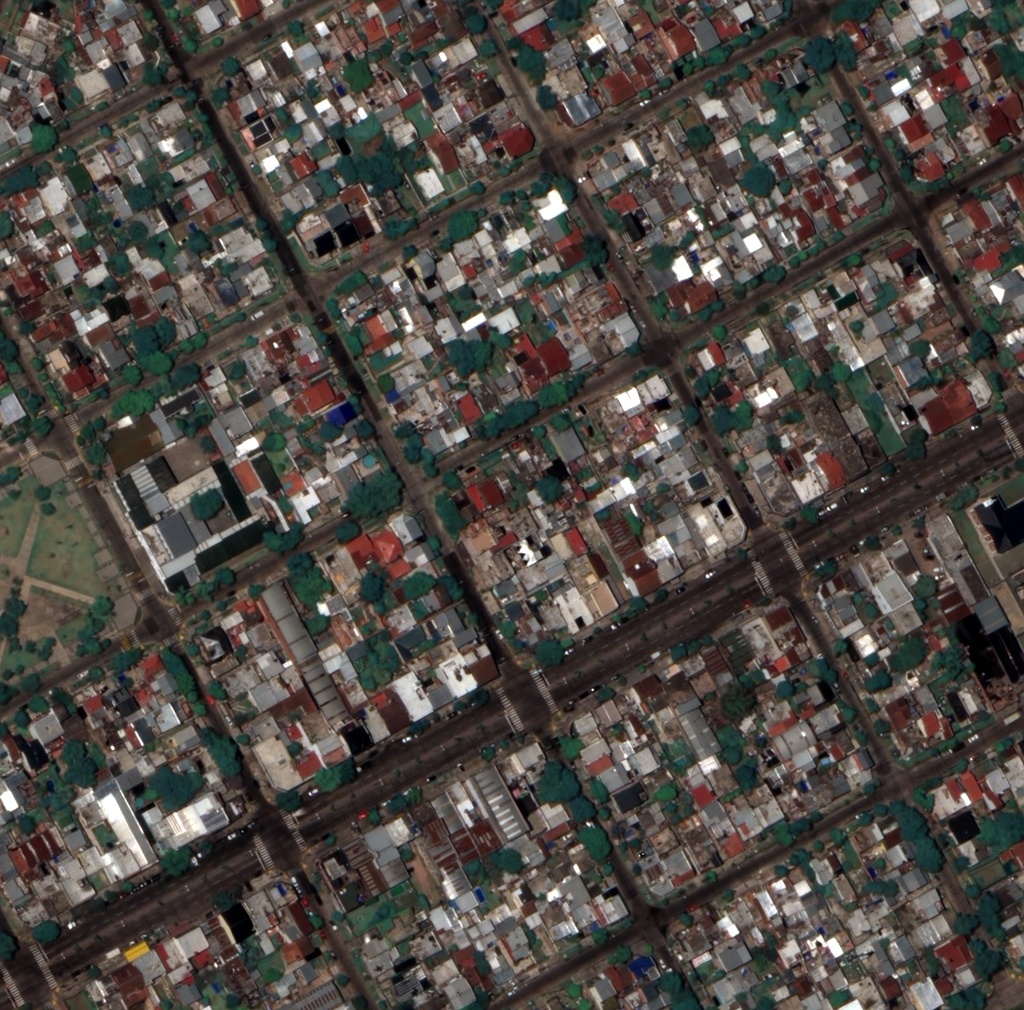}{fig:iarpa_ours_rgb}{}
		& \qpanelcrop{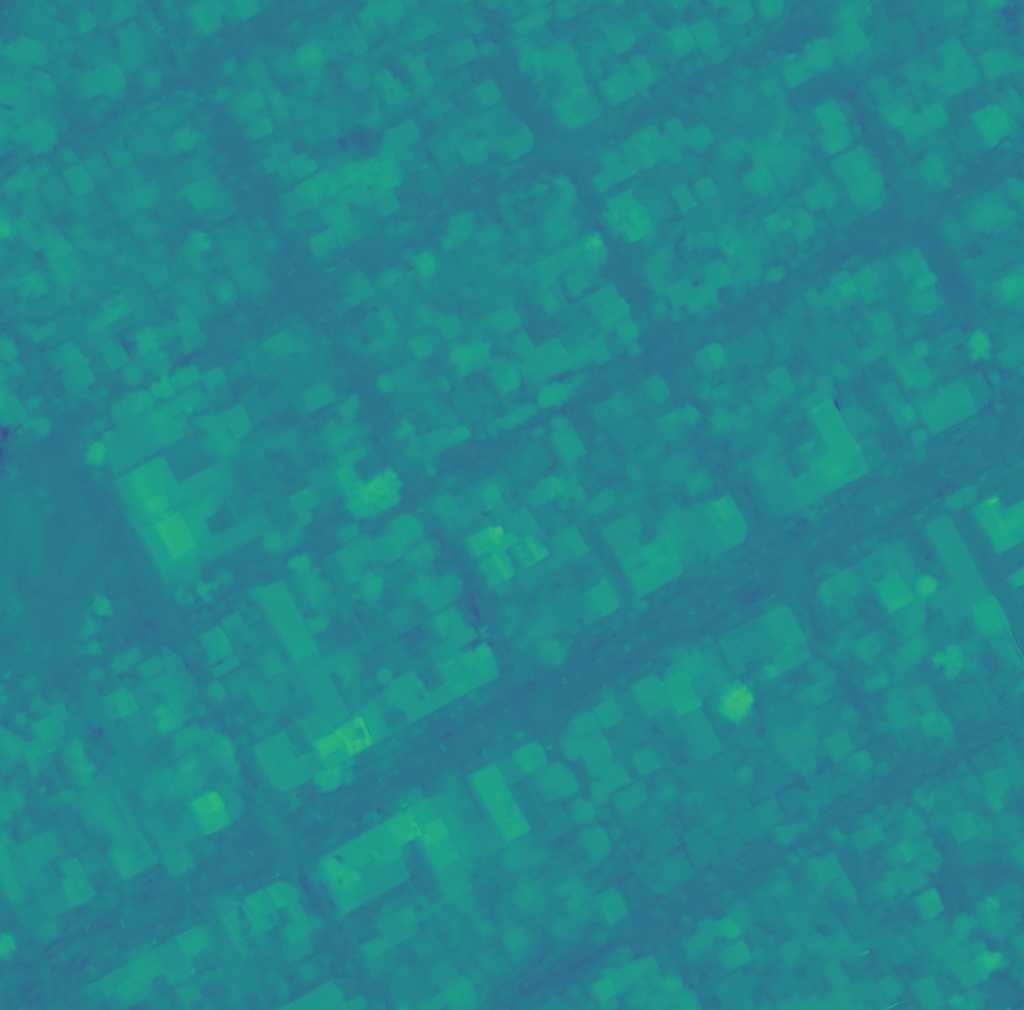}{fig:iarpa_ours_depth}{\detailFA}
		& \qpanelcropnorotate{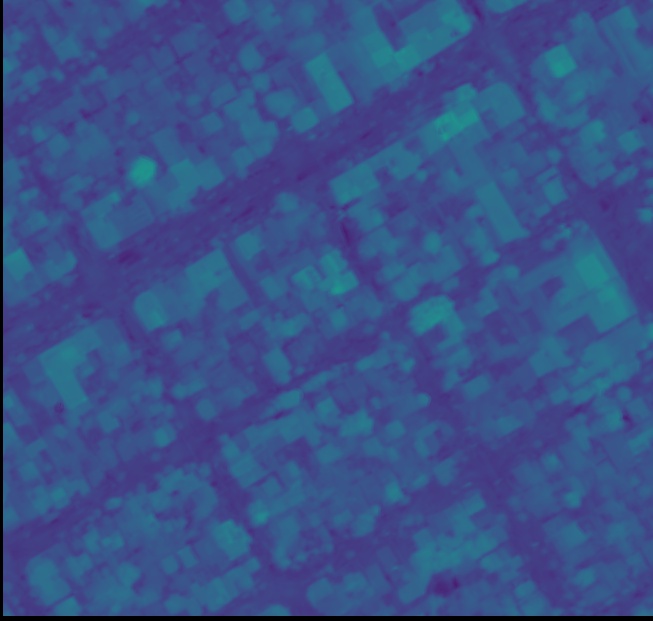}{fig:iarpa_ours_dsm}{\detailDSMFA}
	\end{tabular}
	
	\caption{
		Qualitative comparison for scene 002 from the IARPA2016 dataset~\cite{IARPA2016}. 
		Depth is converted to altitude for cross-camera comparison; the evaluation DSM marks the LiDAR-covered region used for altitude error computation.
		Areas of interest are highlighted in yellow. Colors encode altitude, ranging from blue (low) to yellow (high).
	}
	\label{fig:app_site2}
\end{figure*}	

\newcommand{\detailGA}{%
	\detailbox{23}{22}{30}{30}%
}
\newcommand{\detailDSMGA}{%
	\detailbox{50}{55}{30}{30}%
}
\begin{figure*}[ht]
	\centering
	\setlength{\tabcolsep}{3pt}
	\renewcommand{\arraystretch}{1.0}

	\begin{tabular}{
			@{}
			>{\centering\arraybackslash}m{1.1em}
			@{\hspace{3pt}}
			>{\centering\arraybackslash}m{\qcropw}
			@{\hspace{5pt}}
			>{\centering\arraybackslash}m{\qcropw}
			@{\hspace{5pt}}
			>{\centering\arraybackslash}m{\qcropw}
			@{}
		}
		& \textbf{RGB} & \textbf{Altitude} & \textbf{Evaluation DSM}
		\\
		\noalign{\vskip.4em}
		
		\qrowlabelcrop{GT}
		& \qpanelcrop{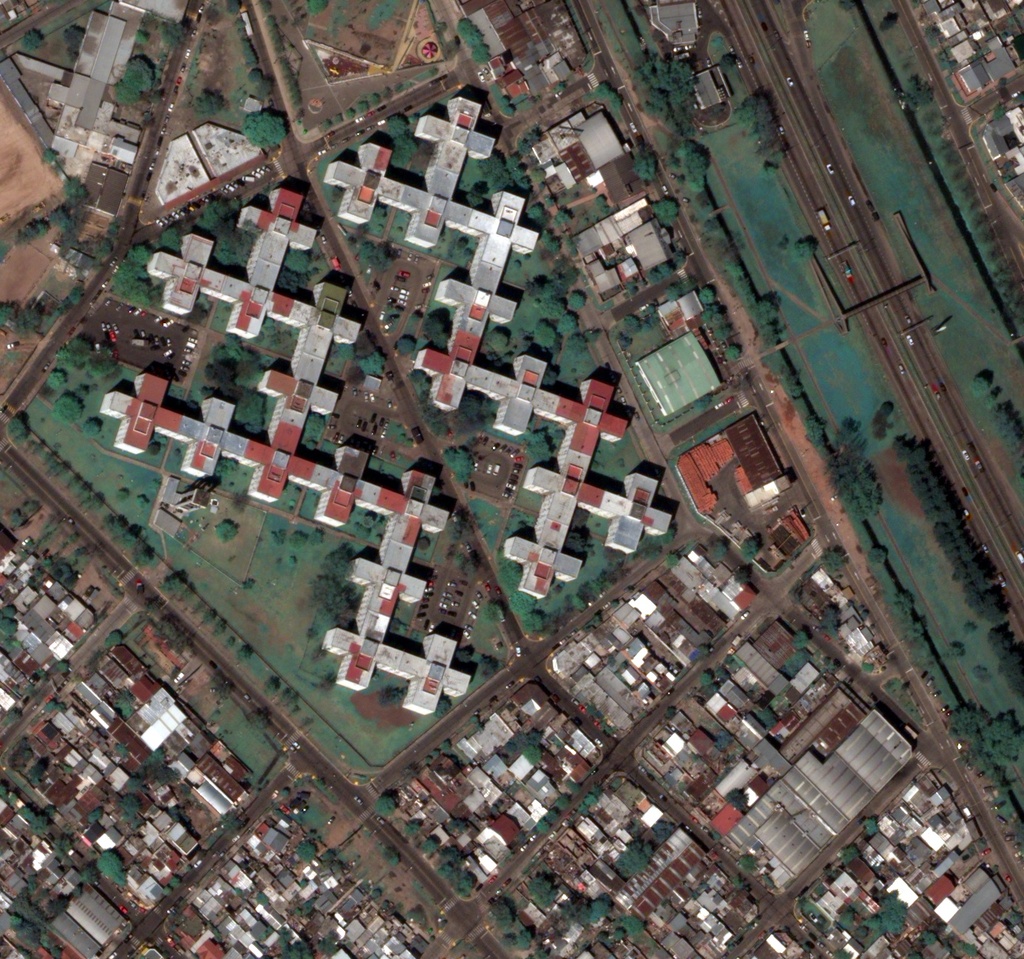}{fig:iarpa_gt_rgb}{}
		& \qmissingcrop{No view-specific GT available}
		& \qpanelcropnorotate{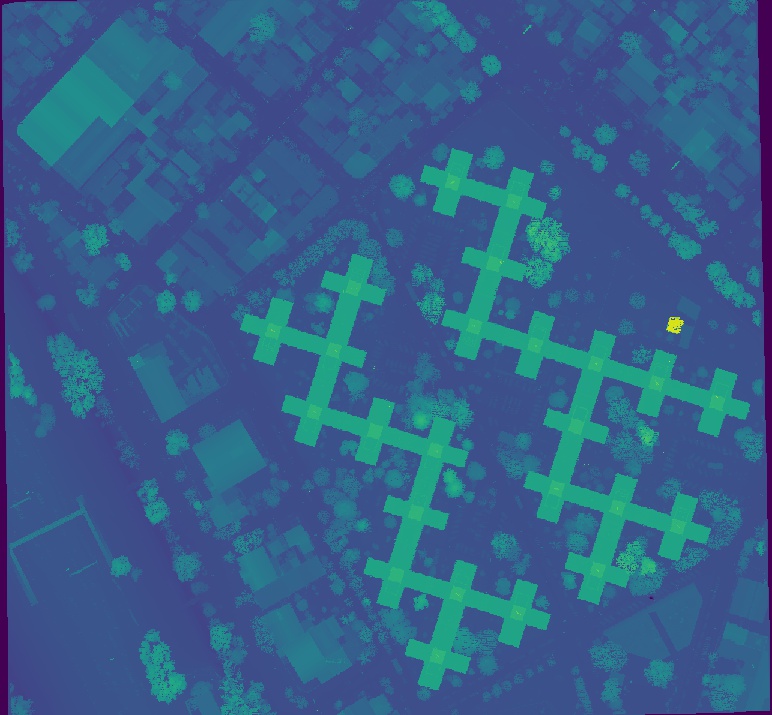}{fig:iarpa_gt_dsm}{\detailDSMGA}
		\\
		\noalign{\vskip\qcroprowgap}
		
		\qrowlabelcrop{Affine Approx.}
		& \qpanelcrop{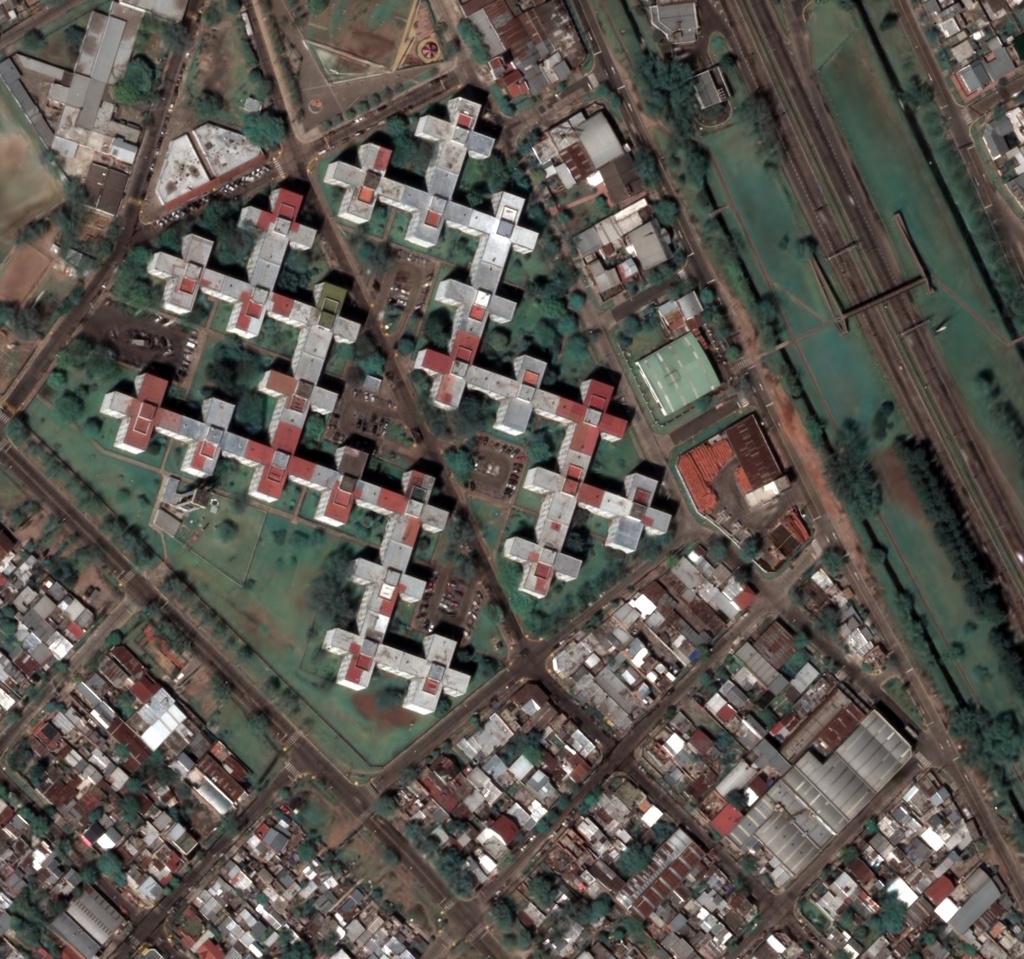}{fig:iarpa_affine_rgb}{}
		& \qpanelcrop{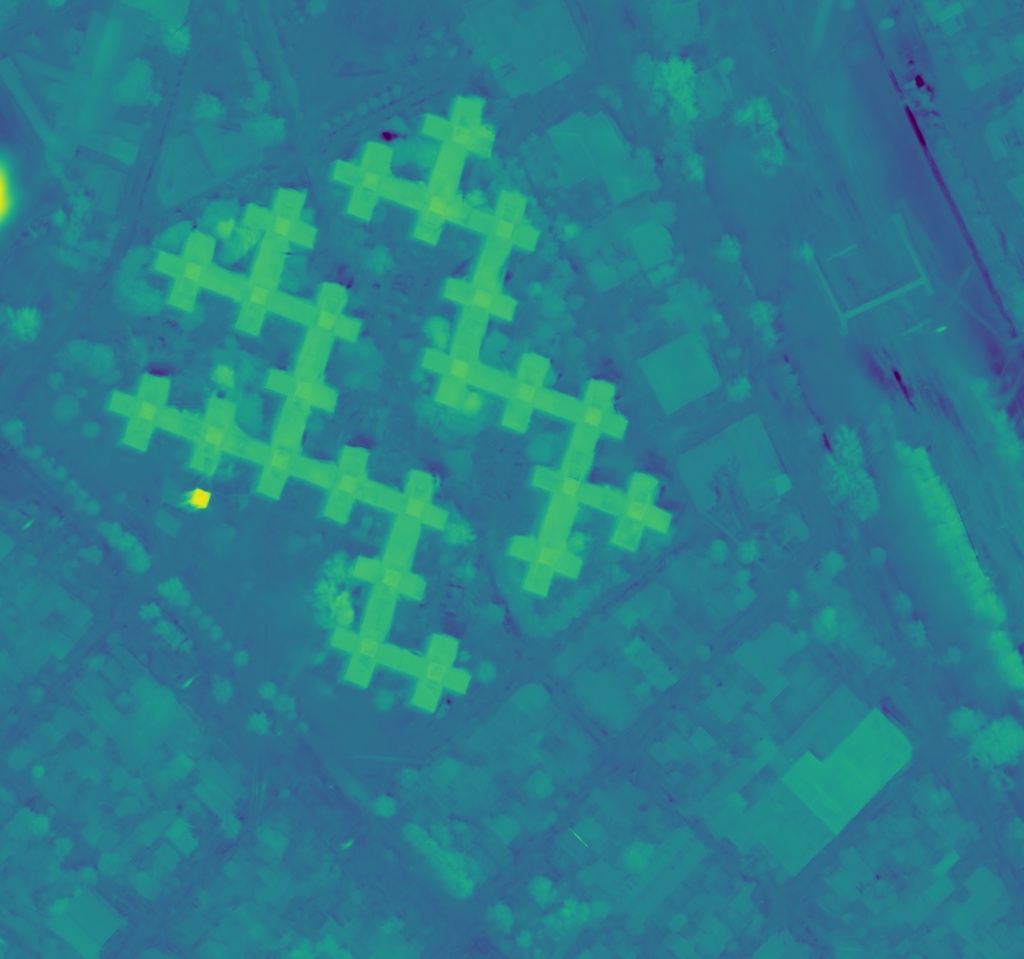}{fig:iarpa_affine_depth}{\detailGA}
		& \qpanelcropnorotate{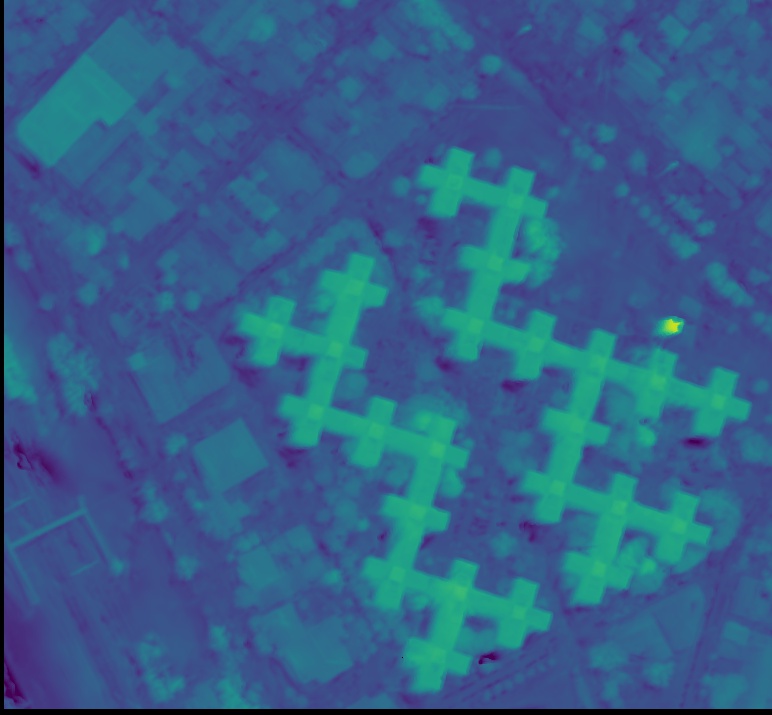}{fig:iarpa_affine_dsm}{\detailDSMGA}
		\\
		\noalign{\vskip\qcroprowgap}
		
		\qrowlabelcrop{Perspective Approx.}
		& \qpanelcrop{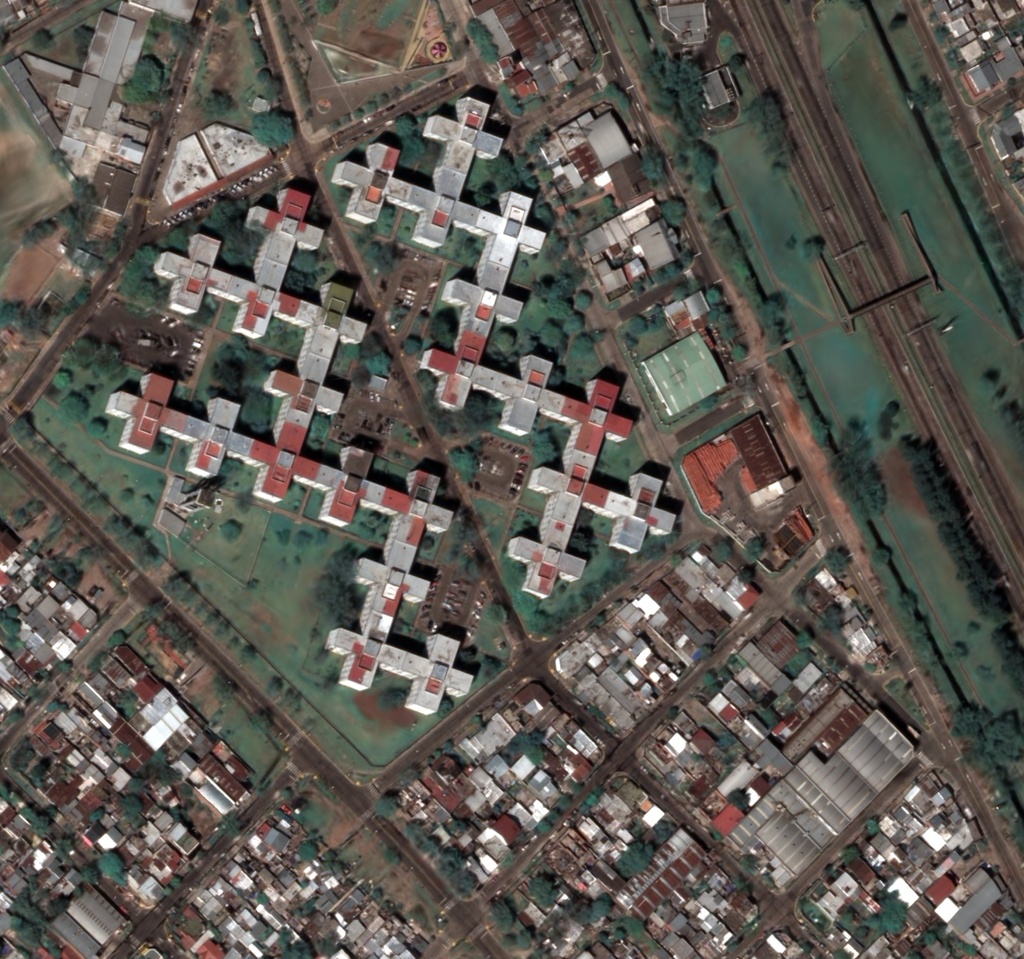}{fig:iarpa_perspective_rgb}{}
		& \qpanelcrop{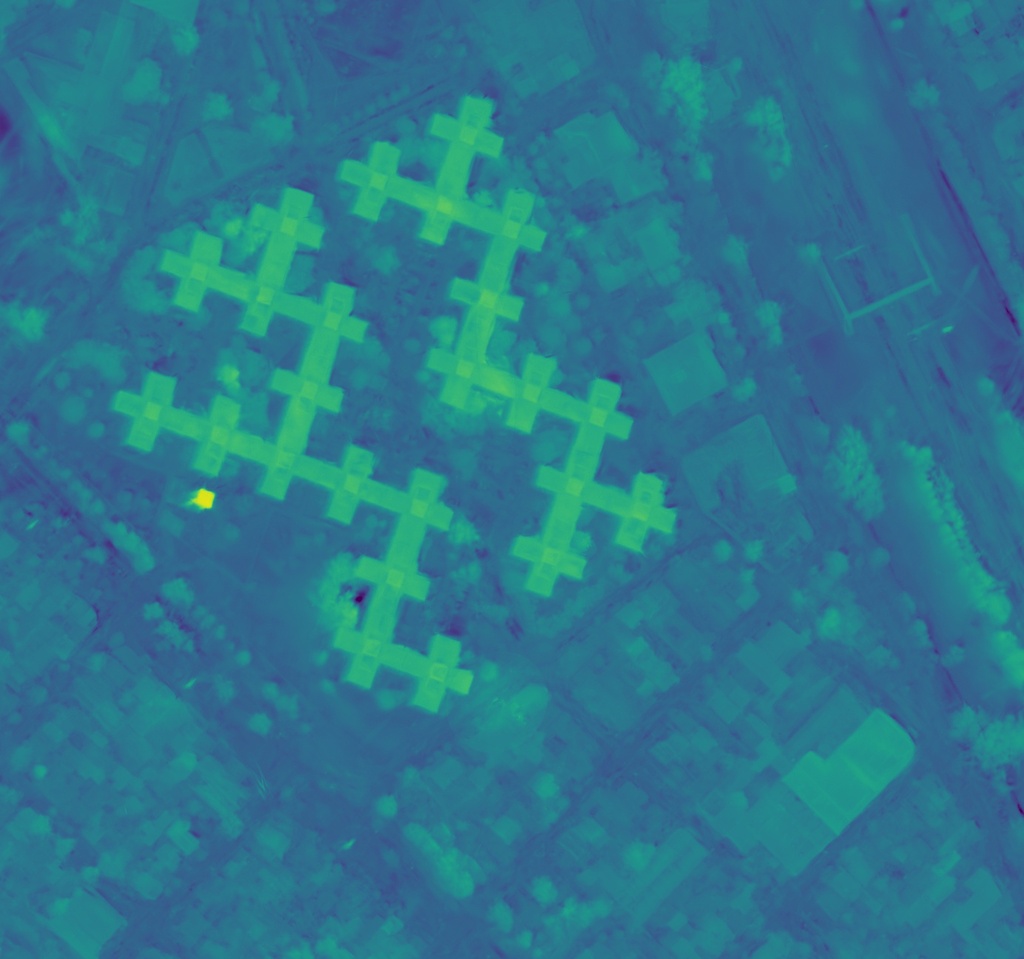}{fig:iarpa_perspective_depth}{\detailGA}
		& \qpanelcropnorotate{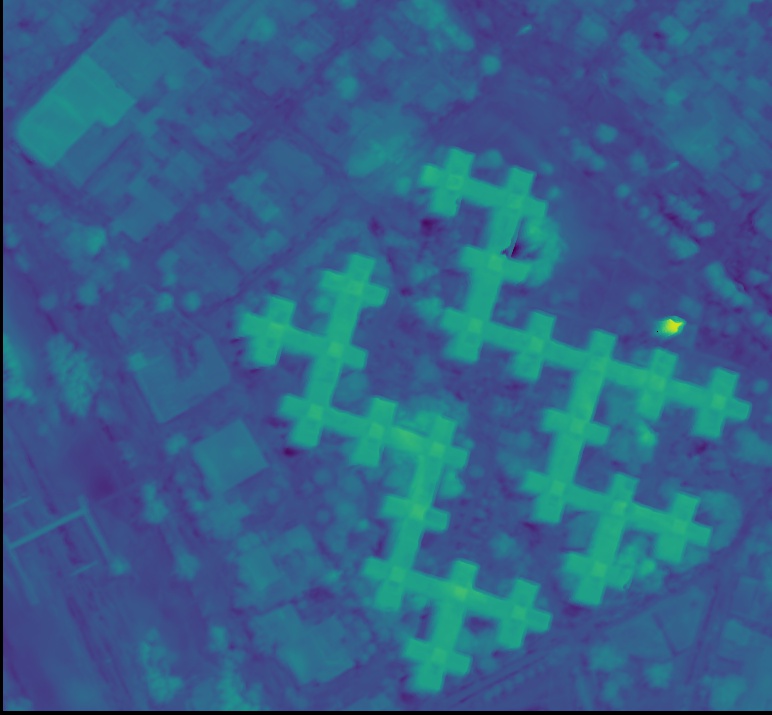}{fig:iarpa_perspective_dsm}{\detailDSMGA}
		\\
		\noalign{\vskip\qcroprowgap}
		
		\qrowlabelcrop{Native RPC (Ours)}
		& \qpanelcrop{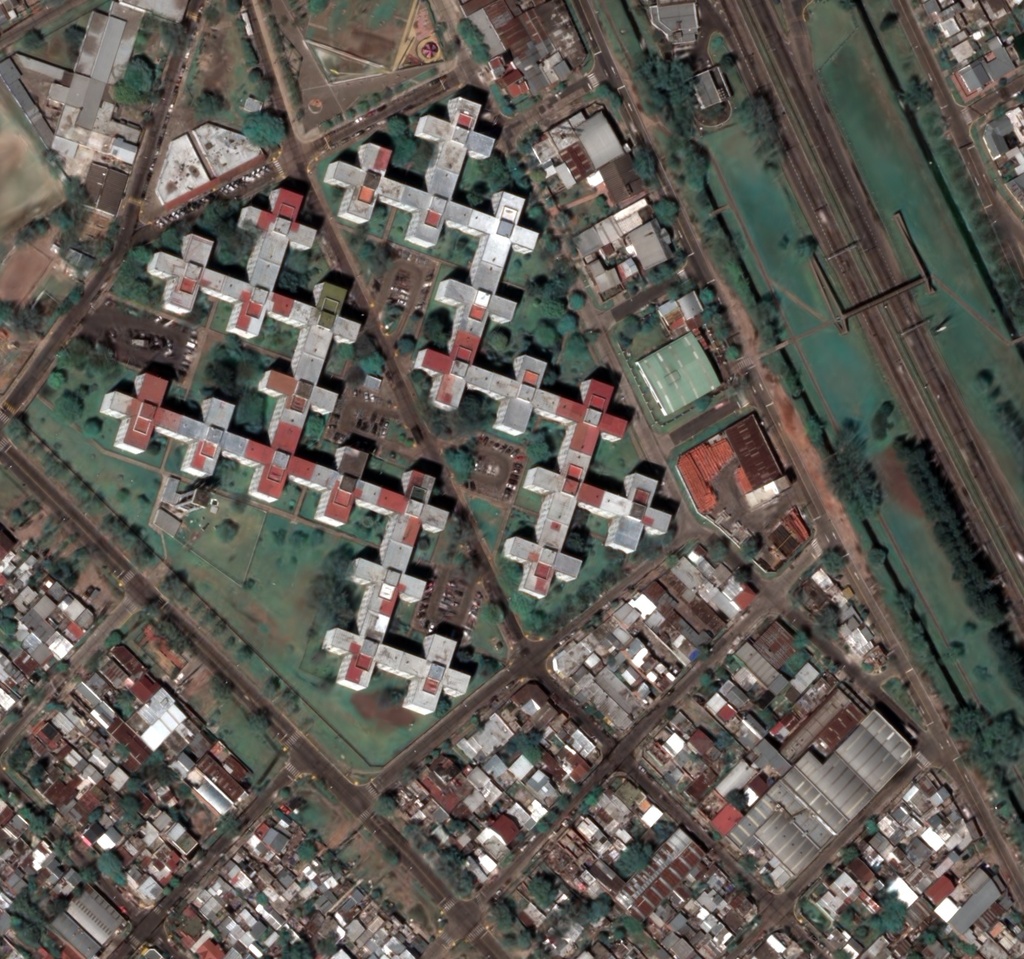}{fig:iarpa_ours_rgb}{}
		& \qpanelcrop{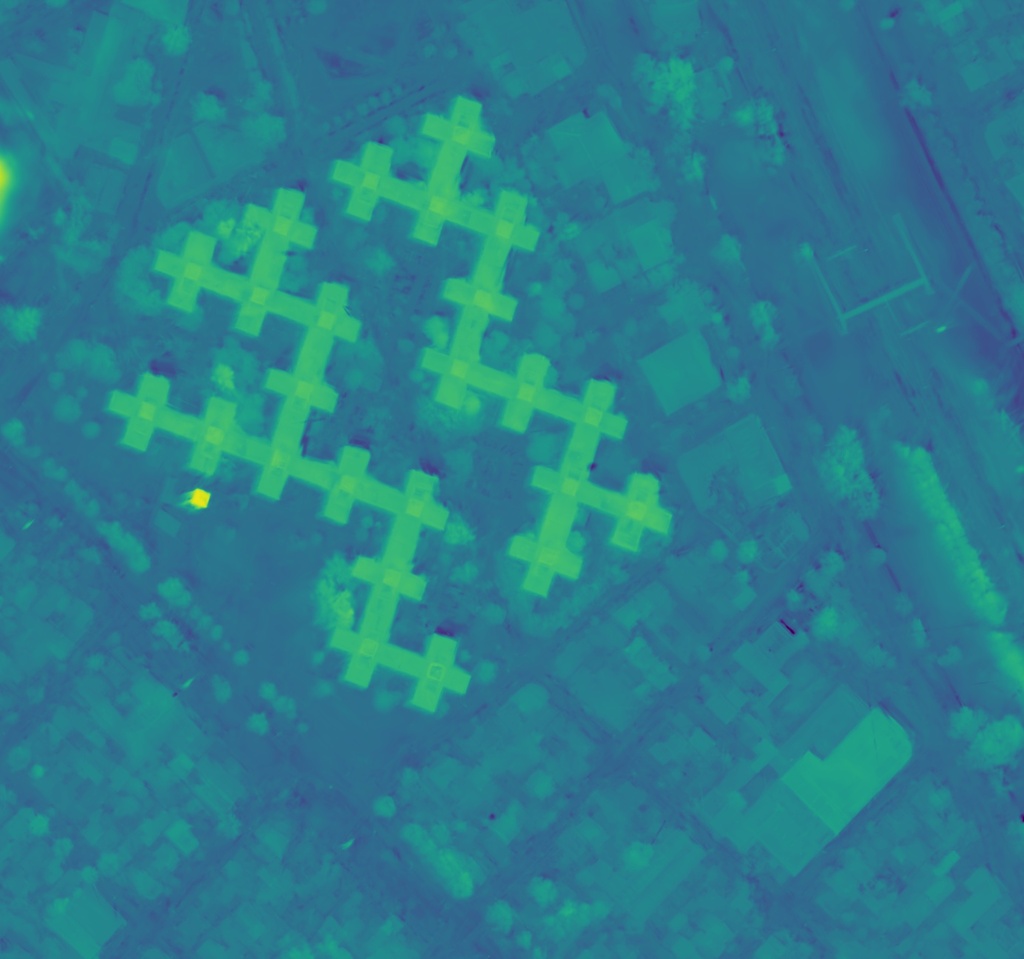}{fig:iarpa_ours_depth}{\detailGA}
		& \qpanelcropnorotate{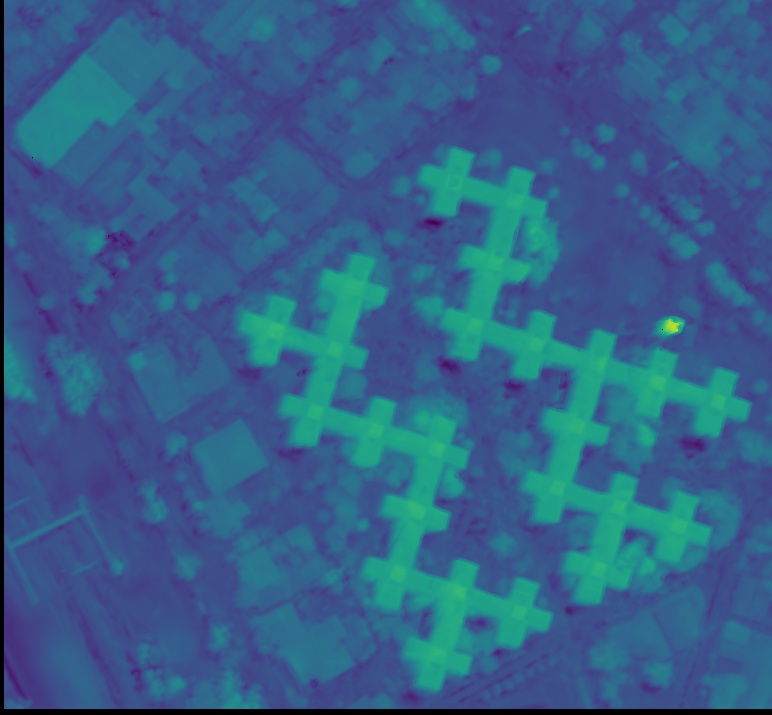}{fig:iarpa_ours_dsm}{\detailDSMGA}
	\end{tabular}
	
	\caption{
		Qualitative comparison for scene 003 from the IARPA2016 dataset~\cite{IARPA2016}. 
		Depth is converted to altitude for cross-camera comparison; the evaluation DSM marks the LiDAR-covered region used for altitude error computation.
		Areas of interest are highlighted in yellow. Colors encode altitude, ranging from blue (low) to yellow (high).
	}
	\label{fig:app_site3}
\end{figure*}	

\clearpage


\end{document}